%% file: main.tex
\documentclass{article}

\PassOptionsToPackage{numbers, compress}{natbib}

\usepackage[final]{neurips_2020}

\usepackage[utf8]{inputenc} 
\usepackage[T1]{fontenc}    
\usepackage{hyperref}       
\usepackage{url}            
\usepackage{booktabs}       
\usepackage{amsfonts}       
\usepackage{nicefrac}       
\usepackage{microtype}      

\input{technical/packages.tex}
\input{technical/newcom.tex}

\title{Certifiably Adversarially Robust Detection of Out-of-Distribution Data}

\author{%
  Julian Bitterwolf \\
  University of Tübingen\\
  \And
  Alexander Meinke \\
  University of Tübingen \\ 
  \And
  Matthias Hein \\
  University of Tübingen \\
}

\begin{document}

\maketitle

\input{sections/abstract}

\input{sections/intro.tex}
\input{figures/teaser}
\input{sections/foundations}

\input{sections/method.tex}
\input{sections/setup.tex}
\input{sections/results.tex}
\input{sections/impact.tex}
\input{sections/acknowledgements.tex}

\small

\bibliographystyle{abbrv}

\bibliography{main.bib}

\clearpage

\normalsize

\section*{APPENDIX}
\begin{appendices}
\input{sections/adversarial}

\input{sections/ood_conf_reduction}
\input{sections/auroc_definition}
\input{sections/experimental_details}

\input{sections/quantile_depiction}

\input{sections/distribution_of_confidences}
\input{sections/additional_datasets}
\input{sections/larger_radius}
\end{appendices}

\end{document}

%% file: technical/packages.tex
\usepackage{multicol}
\usepackage{multirow}
\usepackage{amsmath, mathtools, amssymb, amsthm}
\usepackage{graphbox}
\usepackage{array,booktabs,tabularx}
\usepackage{tikz}
\usepackage[ruled,vlined]{algorithm2e}
\usepackage{svg}
\usepackage{appendix}
\usepackage{color, colortbl}

%% file: technical/newcom.tex
\newcommand{\R}{\mathbb{R}}

\newcommand{\K}{\abs{K}}
\DeclareMathOperator{\Conf}{Conf}
\DeclareMathOperator{\AUROC}{AUC}
\DeclareMathOperator{\CAUROC}{cAUC}

\newcommand{\Lce}{\mathcal{L}_{\text{CE}}}
\newcommand{\Loe}{\mathcal{L}_{\text{OUT}}}
\newcommand{\LCUB}{\mathcal{L}_{\text{CUB}}}

\newcolumntype{C}{>{\centering\arraybackslash}X}

\DeclareMathOperator*{\argmax}{argmax}

\renewcommand{\cite}[1]{\citep{#1}}

\def\R{\mathbb{R}}

\def\K{K} 

\newcommand{\norm}[1]{\left\|#1\right\|}

\def\ones{\mathbf{1}}

\newcommand\irregularcircle[2]{
  \pgfextra {\pgfmathsetmacro\len{(#1)+rand*(#2)}}%
  +(0:\len pt)%
  \foreach \a in {10,20,...,350}{%
    \pgfextra {\pgfmathsetmacro\len{(#1)+rand*(#2)}}%
    -- +(\a:\len pt)%
  } -- cycle
}

\newcommand\Square[1]{+(-#1,-#1) rectangle +(#1,#1)}

\definecolor{lightgrey}{rgb}{0.9, 0.9, 0.9}

%% file: sections/abstract.tex
\begin{abstract}
Deep neural networks are known to be overconfident when applied to out-of-distribution (OOD) inputs which clearly do not belong to any class.
This is a problem in safety-critical applications since a reliable assessment of the uncertainty of a classifier is a key property, allowing the system to trigger human intervention or to transfer into a safe state.
In this paper, we aim for certifiable worst case guarantees for OOD detection by  enforcing not only low confidence at the OOD point but also in an $l_\infty$-ball around it. For this purpose, we use interval bound propagation (IBP) to upper bound the maximal confidence in the $l_\infty$-ball and minimize this upper bound during training time.
We show that non-trivial bounds on the confidence for OOD data generalizing beyond the OOD dataset seen at training time are possible.
Moreover, in contrast to certified adversarial robustness which typically comes with significant loss in prediction performance, certified guarantees for worst case OOD detection are possible without much loss in accuracy. 

\end{abstract}

%% file: sections/intro.tex
\section{Introduction}
\label{Sec:Introduction}

Deep neural networks are the state-of-the-art in many application areas. Nevertheless it is still a major concern to use deep learning in safety-critical systems, e.g. medical diagnosis or self-driving cars,
since it has been shown that deep learning classifiers suffer from a number of unexpected failure modes, such as low robustness to natural perturbations \citep{geirhos2018generalisation,hendrycks2019benchmarking}, overconfident predictions \citep{NguYosClu2015,GuoEtAl2017,HenGim2017,HeiAndBit2019} as well as adversarial vulnerabilities~\citep{SzeEtAl2014}. For safety critical applications, empirical checks are not sufficient in order to trust a deep learning system in a high-stakes decision. Thus provable guarantees on the behavior of a deep learning system are needed.

One property that one expects from a robust classifier is that it should \emph{not} make highly confident predictions on data that is very different from the training data. However, ReLU networks have been shown to be provably overconfident far away from the training data~\citep{HeiAndBit2019}. This is a big problem as (guaranteed) low confidence of a classifier when it operates out of its training domain can be used to trigger human intervention or to let the system try to achieve a safe state when it ``detects'' that it is applied outside of its specification.
Several approaches to the out-of-distribution (OOD) detection task have been studied \citep{HenGim2017,liang2017enhancing,LeeEtAl2018, lee2018simple,HeiAndBit2019}. The current state-of-the-art performance of OOD detection in image classification is achieved by enforcing low confidence on a large training set of natural images that is considered as out-distribution~\citep{HenMazDie2019,meinke2020towards}.

Deep neural networks are also notoriously susceptible to small adversarial perturbations in the  input~\citep{SzeEtAl2014,CarWag2016} which change the decision of a classifier. Research so far has concentrated on adversarial robustness around the in-distribution. Several empirical defenses have been proposed but many could be broken again \cite{CroHei2020,CarWag2017,AthCarWag2018}. Adversarial training and variations~\citep{MadEtAl2018,ZhaEtAl2019} perform well empirically, but typically no robustness guarantees can be given.  Certified adversarial robustness has been achieved by explicit computation of robustness certificates~\citep{HeiAnd2017, WonKol2018, RagSteLia2018, MirGehVec2018, gowal2018effectiveness} and randomized smoothing ~\citep{cohen2019certified}. 

Adversarial changes to generate high 
confidence predictions on the out-distribution have received much less attention although it has been shown early on that they can be used to fool a classifier ~\citep{NguYosClu2015,SchEtAl2018,sehwag2019better}. Thus, even if a classifier consistently manages to identify samples as not belonging to the in-distribution, it might still assign very high confidence to only marginally perturbed samples from the out-distribution, see Figure \ref{fig:teaser}. A first empirical defense using a type of adversarial training for OOD detection has been proposed in \cite{HeiAndBit2019}. However,
up to our knowledge in the area of certified out-of-distribution detection the only robustness guarantees for OOD were given in~\cite{meinke2020towards}, 
where a density estimator for in- and out-distribution is integrated into the predictive uncertainty of the neural network, which allows them to guarantee that far away from the training data the confidence of the neural network becomes uniform over the classes.
Moreover, they provide guarantees on the maximal confidence attained on $l_2$-type balls around uniform noise.
However, this technique is not able to provide meaningful guarantees around points which are similar or even close to the in-distribution data and, as we will show, provide only weak guarantees against $l_\infty$-adversaries.

In this work we aim to provide worst-case OOD guarantees not only for noise but also for images from related but different image classification tasks. For this purpose we use the techniques from interval bound propagation (IBP)~\citep{gowal2018effectiveness} to derive a provable upper bound on the maximal confidence of the classifier in an $l_\infty$-ball of radius $\epsilon$ around a given point. By minimizing this bound on the out-distribution using our training scheme GOOD (Guaranteed Out-Of-distribution Detection) we arrive at the first models which have guaranteed low confidence even on image classification tasks related to the original one; e.g., we get state-of-the-art results on separating letters from EMNIST from digits in MNIST even though the digit classifier has never seen any
images of letters at training time. In particular, the guarantees for the training out-distribution generalize to other out-distribution datasets. In contrast to classifiers which have certified adversarial robustness on the in-distribution, GOOD has the desirable property to achieve provable guarantees for OOD detection with almost no loss in accuracy on the in-distribution task even on datasets like CIFAR-10.

%% file: figures/teaser.tex
\begin{figure}
\centering
\pgfmathsetseed{5}
\resizebox{1.0\textwidth}{!}{
\noindent
\begin{tikzpicture}
  \coordinate (in) at (0.5,0.2);
  \coordinate (sin) at (-0.3,0.4);
  \coordinate (out) at (4.5,.5);
  \coordinate (sadv) at (3.7,0.8);
  \coordinate (pin) at (-3.5,0.8);
  \coordinate (pout) at (7.5,.8);
  \coordinate (padv) at (10.3,.8);
  \coordinate (padvtop) at (10.3,1.6);
  \coordinate (outtxt) at (2.7,2.0);
  \draw[black!40!green, rounded corners=1mm, fill=black!10!green, line width=0.8mm,] (in) \irregularcircle{1.8cm}{6mm};
  \draw[black!40!blue, fill=blue!30, line width=0.8mm] (out) \Square{1.1cm};
  \draw (in) node {\huge \textsc{in}};
  \draw (outtxt) node {\huge \textsc{out}};
  \draw (sin) node {\huge $\times$};
  \draw (out) node {\huge $\times$};
  \draw (sadv) node {\huge $\times$};
  \draw[black, fill=black, line width=0.5mm] (pin) \Square{0.95cm};
  \draw[line width=0.35mm] (sin) -- (pin);
  \node[inner sep=0pt] at (pin) {\includegraphics[width=2cm]{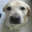}};
  \draw[black, fill=black, line width=0.5mm] (pout) \Square{0.95cm};
  \draw[line width=0.35mm] (out) -- (pout);
  \node[inner sep=0pt] at (pout) {\includegraphics[width=2cm]{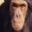}};
  \draw[black, fill=black, line width=0.5mm] (padv) \Square{0.95cm};
  \draw[line width=0.35mm] (sadv) to[out=40,in=160] (padvtop);
  \node[inner sep=0pt] at (padv) {\includegraphics[width=2cm]{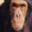}};
  \draw[anchor=north east, text width=2.9cm, align=right] (pin) + (1.4,-1) node {Plain:\,100\%\,\textit{dog}   OE\citep{HenMazDie2019}:\,98\%\,\textit{dog} CCU\cite{meinke2020towards}:\,100\%\,\textit{dog}   ACET\cite{HeiAndBit2019}:\,99\%\,\textit{dog} GOOD\textsubscript{80}:\,100\%\,\textit{dog}};
  \draw[anchor=north east, text width=2.7cm, align=right] (pout) + (0.5,-1) node {Plain:\,100\%  OE:\,15\%  CCU:\,18\%  ACET:\,12\% GOOD\textsubscript{80}:\,12\%};
  \draw[anchor=north west, text width=0.8cm, align=left] (pout) + (0.35,-1) node {\textit{dog} \textit{dog} \textit{dog} \textit{dog} \textit{dog}};
  \draw[anchor=north east, text width=2.7cm, align=right] (padv) + (0.6,-1) node {Plain:\,100\%   OE:\,100\%   CCU:\,99\%  ACET:\,13\% GOOD\textsubscript{80}:\,15\%};
  \draw[anchor=north west, text width=0.8cm, align=left] (padv) + (0.45,-1) node {\textit{dog}  \textit{dog} \textit{dog} \textit{dog} \textit{dog}};  
  \draw[black!40!blue, anchor=north, text width=2.5cm, align=center] (out) + (0,-1.15) node {\large GOOD\textsubscript{80} {guarantees} Conf < 22.7\% };
\end{tikzpicture}
}
\caption{\label{fig:teaser}\textbf{Overconfident predictions on out-distribution inputs.} \textbf{Left:} On the in-distribution CIFAR-10 all methods have similar high confidence on the image of a \textit{dog}. \textbf{Middle:} For the out-distribution image of a \textit{chimpanzee} from CIFAR-100 the plain model is overconfident while an out-distribution aware method like Outlier Exposure (OE) \cite{HenMazDie2019} produces low confidence. \textbf{Right:} When maximizing the confidence inside the $l_\infty$-ball of radius $0.01$ around the \textit{chimpanzee} image (for the OE model), OE as well as CCU become overconfident (right image). ACET and our GOOD\textsubscript{80} perform well in having empirical low confidence, but only GOOD\textsubscript{80} guarantees that the confidence in the $l_\infty$-ball of radius $0.01$ around the \textit{chimpanzee} image (middle image) is less than 22.7\% for any class (note that $10\%$ corresponds to maximal uncertainty as CIFAR-10 has $10$ classes).
}
\end{figure}
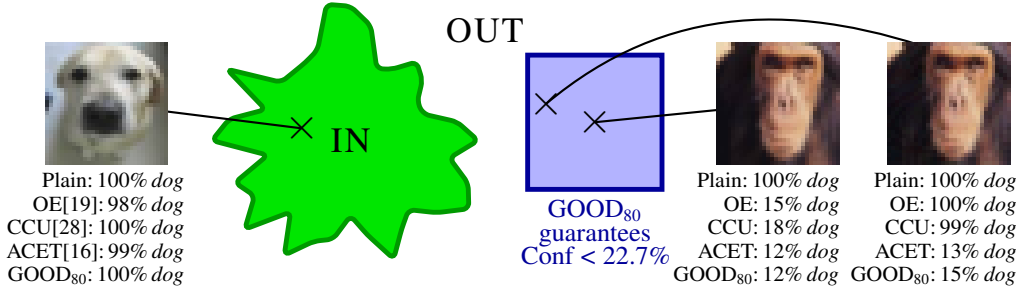

%% file: sections/foundations.tex
\section{Out-of-distribution detection: setup and baselines}
Let $f:\R^d \rightarrow \R^K$ be a feedforward neural network (DNN) with a last linear layer where $d$ is the input dimension and $K$ the number of classes. In all experiments below we use the ReLU activation function. The logits of $f(x)$ for $x \in \R^d$ are transformed via the softmax function into a probability distribution $p(x)$ over the classes with:
\begin{align}
    p_k(x) := \frac{e^{f_k(x)}}{\sum_{l=1}^K e^{f_l(x)}} \; \text{ for } k=1,\ldots,K .\label{eq:defprob}
\end{align}
By $\Conf_{\!f}(x)=\max_{k=1,\ldots, K} p_k(x)$ we define the confidence of the classifier $f$ in the prediction $\argmax_{k=1,\ldots,K} p_k(x)$ at $x$.

The general goal of OOD detection is to construct a feature that can reliably separate the in-distribution from all inputs which clearly do not belong to the in-distribution task, especially inputs from regions which have zero probability under the in-distribution.
One typical criterion to measure OOD detection performance is to use $\Conf_{\!f}(x)$
as a feature and compute the AUC of in- versus out-distribution (how well are confidences of in- and out-distribution separated). We discuss a proper conservative measurement of the AUC in case of indistinguishable confidence values, e.g. due to numerical precision,
in Appendix \ref{section:auroc}.

As baselines and motivation for our provable approach we
use the OOD detection methods Outlier Exposure (OE)~\cite{HenMazDie2019} and Confidence Enhancing Data Augmentation (CEDA)~\cite{HeiAndBit2019}, which use as objective for training 
\begin{align} \label{eq:objective_OE}
    \frac{1}{N}\sum_{i = 1}^{N} \Lce(x_i^{\text{IN}},y_i^{\text{IN}}) + \frac{\kappa}{M}\sum_{j=1}^{M} \Loe(x_j^{\text{OUT}})
    \ ,
\end{align}
where $\left\{(x_i^{\text{IN}},y_i^{\text{IN}}) \mid 1 \leq i \leq N \right\}$ is the in-distribution training set, $\left\{x_j^{\text{OUT}} \mid 1 \leq j \leq M \right\}$ the out-distribution training set, and $\Lce$ the cross-entropy loss. The hyper-parameter $\kappa$ determines the relative magnitude of the two loss terms and is usually chosen to be one~\citep{HenMazDie2019,meinke2020towards,HeiAndBit2019}. OE and CEDA differ in the choice of the loss $\Loe$ for the out-distribution where OE
uses the cross-entropy loss between $p(x_j^\text{OUT})$ and the uniform distribution and CEDA uses $\log\Conf_{\!f}(x_j^{\text{OUT}})$. 
Note that both the CEDA and OE loss attain their global minimum when $p(x)$ is the uniform distribution. Their difference is typically minor in practice. An important question is the choice of the out-distribution. For general image classification, it makes sense to use an out-distribution which encompasses basically any possible image one could ever see at test time and thus the set of all natural images is a good out-distribution; following
\cite{HenMazDie2019} we use the 80 Million Tiny Images dataset \cite{torralba200880} as a proxy for that.

While OE and CEDA yield state-of-the-art OOD detection performance for image classification tasks when used together with the 80M Tiny Images dataset as out-distribution, they are, similarly to normal classifiers, vulnerable to adversarial manipulation of the out-distribution images where the attack is trying to maximize the confidence in this scenario \cite{meinke2020towards}. Thus \cite{HeiAndBit2019} proposed Adversarial Confidence Enhanced Training (ACET) which replaces the CEDA loss with $\max_{\norm{\hat{x}-x_j^{\text{OUT}}}_\infty \leq \epsilon} \log\Conf_{\!f}(\hat{x})$ and can be seen as adversarial training on the out-distribution for an $l_\infty$-threat model. However, as for to adversarial training on the in-distribution \cite{MadEtAl2018} this does not yield any guarantees for out-distribution detection. In the next section we discuss how to use interval-bound-propagation (IBP) to get guaranteed OOD detection performance in an
$l_\infty$-neighborhood of every out-distribution input.

%% file: sections/method.tex
\section{Provable guarantees for out-of-distribution detection}

Our goal is to minimize the confidence of the classifier not only on
the out-distribution images themselves
but in a whole neighborhood around them. For this purpose, we first derive bounds on the maximal confidence on some $l_\infty$-ball around a given point.
In certified adversarial robustness, IBP \cite{gowal2018effectiveness} currently leads to the best guarantees for deterministic classifiers under the $l_\infty$-threat model. While other methods for deriving guarantees yield tighter bounds \cite{WonKol2018, MirGehVec2018}, they are not easily scalable and,  when optimized, the bounds given by IBP have been shown to be very tight \cite{gowal2018effectiveness}.

\paragraph{IBP.} Interval bound propagation \citep{gowal2018effectiveness} provides entrywise lower and upper bounds $\underline{z_k}_ \epsilon$ resp. $\overline{z_k}^\epsilon$ for the output $z_k$ of the $k$-th layer of a neural network given that the input $x$ is varied in the $l_\infty$-ball of radius $\epsilon$.
Let $\sigma:\R \rightarrow \R$ be a monotonically increasing activation function e.g. we use the ReLU function $\sigma(x)=\max\{0,x\}$ in the paper. We set $z_0=x$ and $\underline{z_0}_\epsilon=x-\epsilon \cdot \ones$ and $\overline{z_0}^\epsilon=x+\epsilon \cdot \ones$ ($\ones$ is the vector of all ones).
If the $k$-th layer is linear (fully connected, convolutional, residual etc.) with weight matrix $W_k$, one gets upper and lower bounds of the next layer via forward propagation:
\begin{align}
     \overline{z_k}^\epsilon &= \sigma\big(\max(W_k, 0) \cdot  \overline{z_{k-1}}^\epsilon + \min(W_k, 0)  \cdot \underline{z_{k-1}}_\epsilon + b_k\big) \nonumber\\
    \underline{z_k}_\epsilon &= \sigma\big(\min(W_k, 0) \cdot \overline{z_{k-1}}^\epsilon + \max(W_k, 0) \cdot \underline{z_{k-1}}_\epsilon +b_k\big)\ ,
\end{align}
where the $\min$/$\max$ expressions are taken componentwise and the activation function $\sigma$ is applied componentwise as well. Note that the derivation in \citep{gowal2018effectiveness} is slightly different, but the bounds are the same.
The forward propagation of the bounds is of similar nature as a standard forward pass and back-propagation w.r.t. the weights is relatively straightforward.

\paragraph{Upper bound on the confidence in terms of the logits.} 
The log confidence of the model at  $x$ can be written as 
\begin{align}
\begin{split}
    \log \Conf(x)   
    = \max_{k=1,\ldots,K} \log \frac{e^{f_k(x)}}{\sum_{l=1}^{ K}e^{f_l(x)}}
    = \max_{k=1,\ldots,K} {-\log}\sum_{l=1}^K e^{f_l(x)-f_k(x)}.\label{eq:softmax}
\end{split}\end{align}

We assume that the last layer is affine: $f(x) = W_L \cdot z_{L-1}(x) + b_L$, where $L$ is the number of layers of the network. We calculate the upper bounds of all $\K^2$ logit differences as: 
\begin{align}
\begin{split}
    \max_{\norm{\hat{x}-x}_\infty \leq \epsilon} f_k(\hat{x}) - f_l(\hat{x}) 
    &= \max_{\norm{\hat{x}-x}_\infty \leq \epsilon} W_{L,k} \cdot z_{L-1}(\hat{x}) + b_{L,k} - W_{L,l} \cdot z_{L-1}(\hat{x}) - b_{L,l}\\
    &= \max_{\norm{\hat{x}-x}_\infty \leq \epsilon} (W_{L,k} - W_{L,l}) \cdot z_{L-1}(\hat{x}) + b_{L,k} - b_{L,l}\\
    &\leq \phantom{a} \max(W_{L,k} - W_{L,l}, 0) \cdot \overline{z_{L-1}(x)}^\epsilon \label{eq:logitud} \\
   &\quad+ \min(W_{L,k} - W_{L,l},0) \cdot \underline{z_{L-1}(x)}_\epsilon + b_{L,k} - b_{L,l} \\
    &=: \overline{f_k(x) - f_l(x)}^\epsilon,
\end{split}
\end{align}
where $W_{L,k}$ denotes the $k$-th row of $W_L$ and $b_{L,k}$ is the $k$-th component of $b_L$.
Note that this upper bound of the logit difference can be negative and is zero for $l = k$. 
Using this upper bound on the logit difference in Equation \eqref{eq:softmax}, we obtain an upper bound on the log confidence:
\begin{align}\label{eq:softmaxub}
    \max_{\norm{\hat{x}-x}_\infty \leq \epsilon} \log \Conf(\hat{x})
    \leq \max_{k=1,\ldots,K} -\log\sum_{l=1}^K e^{-(\overline{f_k(x) - f_l(x)}^\epsilon)} 
\end{align}
We use the bound in \eqref{eq:softmaxub} 
to evaluate the
guarantees on the confidences for given out-distribution datasets.
However, minimizing it directly during training leads to numerical problems, especially at the beginning of training, when the upper bounds $\overline{f_k(x) - f_l(x)}^\epsilon$ are very large for $l \neq k$ 
, which makes training numerically infeasible.
Instead, we rather upper bound the log confidence again by bounding the sum inside the negative log from below with $\K$ times its lowest term:
\begin{align}
\begin{split}
    \max_{k=1,\ldots,K} -\log\sum_{l=1}^K e^{-(\overline{f_k(x) - f_l(x)}^\epsilon)} 
    &\leq \max_{k=1,\ldots,K} -\log \left(K \cdot \min_{l=1,\ldots,K} e^{-(\overline{f_k(x) - f_l(x)}^\epsilon)} \right) \\
    &= \max_{k,l=1,\ldots,K} \overline{f_k(x) - f_l(x)}^\epsilon -\log K
\end{split} \label{eq:bound2}
\end{align}
While this bound can considerably differ from the potentially tighter bound of Equation \eqref{eq:softmaxub}, it is often quite close as one term in the sum dominates the others. Moreover, both bounds have the same global minimum when all logits are equal over the $l_\infty$-ball. We omit the constant $\log K$ in the following as it does not matter for training. 

The direct minimization of the upper bound in \eqref{eq:bound2} is still difficult, in particular for more challenging in-distribution datasets like SVHN and CIFAR-10, as the bound
 $\max_{k,l=1,\ldots,K} \overline{f_k(x) - f_l(x)}^\epsilon$
can be several orders of magnitude larger than the in-distribution loss. Therefore, we use the logarithm of this quantity. However, we also want to have a more fine-grained optimization when the upper bound becomes small in the later stage of the training. 
Thus we define the Confidence Upper Bound loss $\LCUB$ for an OOD input as 
\begin{align}
\label{eq:cub_loss}
    \LCUB(x; \epsilon) := \log \left( \frac{\Big(\max\limits_{k,l=1,\ldots,K} \overline{f_k(x) - f_l(x)}^\epsilon\Big)^2}{2} + 1 \right).
\end{align}
Note that $\log(\frac{a^2}{2}+1)\approx \frac{a^2}{2}$ for small $a$
and thus we achieve the more fine-grained optimization with an $l_2$-type of loss in the later stages of training which tries to get all upper bounds small.
The overall objective of \textbf{fully applied Guaranteed OOD Detection training (GOOD\textsubscript{100})} is the minimization of
\begin{align} \label{eq:objective_100}
    \frac{1}{N} \sum_{i = 1}^{N}  \Lce(x_i^{\text{IN}},y_i^{\text{IN}}) + \frac{\kappa}{M}\sum_{j=1}^M \LCUB(x_{j}^{\text{OUT}}; \epsilon)
    \ ,
\end{align}
where $\left\{(x_i^{\text{IN}},y_i^{\text{IN}}) \mid 1 \leq i \leq N \right\}$ is the in-distribution training set and $\left\{x_j^{\text{OUT}} \mid 1 \leq j \leq M \right\}$ the out-distribution. 
The hyper-parameter $\kappa$ determines the relative magnitude of the two loss terms. During training we slowly increase this value and $\epsilon$ in order to further stabilize the training with GOOD.

\paragraph{Quantile-GOOD: trade-off between clean and guaranteed AUC.}
Training models by minimizing \eqref{eq:objective_100} means that the classifier gets severely punished if \emph{any} training OOD input receives a high confidence upper bound.
If OOD inputs exist to which the classifier already assigns high confidence without even considering the worst case, e.g. as these inputs share features with the in-distribution, it makes little sense to enforce low confidence guarantees.
Later in the experiments we show that for difficult tasks like CIFAR-10 this can happen. In such cases the normal AUC for OOD detection gets worse as the high loss of the out-distribution part effectively leads to low confidence on a significant part of the in-distribution which is clearly undesirable.

Hence, for OOD inputs $x$ which are not clearly distinguishable from the in-distribution, it is preferable to just have the ``normal'' loss  $\LCUB(x_{j}^{\text{OUT}}; 0)$ without considering the worst case.
We realize this by enforcing the loss with the guaranteed upper bounds on the confidence just on some quantile of the easier OOD inputs, namely the ones with the lowest guaranteed out-distribution loss $\LCUB(x;\epsilon)$.
We first order the OOD training set by the potential loss $\LCUB(x; \epsilon)$ of each sample in ascending order $\pi$, that is $\LCUB(x_{\pi_1}^\text{OUT})\leq \LCUB(x_{\pi_2}^\text{OUT}) \leq \ldots \leq \LCUB(x_{\pi_M}^\text{OUT})$. We then apply the loss $\LCUB(x; \epsilon)$ to the lower quantile $q$ of the points (the ones with the smallest loss $\LCUB(x;\epsilon)$) and take $\LCUB(x; 0)$ for the remaining samples, which means no worst-case guarantees on the confidence are enforced:
\begin{align} \label{eq:objective_quantile}
    \frac{1}{N} \sum_{i = 1}^{N} \Lce(x_i^{\text{IN}},y_i^{\text{IN}}) 
    +  \frac{\kappa}{M} \sum_{j = 1}^{\lfloor q \cdot M \rfloor} \LCUB(x_{\pi_j}^{\text{OUT}}; \epsilon)
    +  \frac{\kappa}{M} \sum_{\mathclap{j = \lfloor q \cdot M \rfloor + 1}}^{M} \LCUB(x_{\pi_j}^{\text{OUT}}; 0)
    \ .
\end{align}
During training we do this ordering on the part of each batch consisting of out-distribution images. On CIFAR-10, where the out-distribution dataset 80M Tiny Images is closer to the in-distribution, the quantile GOOD-loss allows us  to choose the trade-off between clean and guaranteed AUC for OOD detection, similar to the trade-off between clean and robust accuracy in adversarial robustness.

%% file: sections/setup.tex
\section{Experiments}
We provide experimental results for image recognition tasks with MNIST \cite{mnist}, SVHN \cite{SVHN} and CIFAR-10 \cite{krizhevsky2009learning} as in-distribution datasets. We first discuss the training details, hyperparameters and evaluation before we present the results of GOOD
and competing methods. Code is available under \url{https://gitlab.com/Bitterwolf/GOOD}.

\subsection{Model architectures, training procedure and evaluation}

\textbf{Model architectures and data augmentation.} For all experiments, we use deep convolutional neural networks consisting of convolutional, affine and ReLU layers. 
For MNIST, we use the large architecture from~\citep{gowal2018effectiveness}, and for SVHN and CIFAR-10 a similar but deeper and wider model. The layer structure is laid out in Table \ref{table:architectures} in the appendix.
Data augmentation is applied to both in- and out-distribution images during training.
For MNIST we use random crops to size $28\!\times\!28$ with padding 4 and for SVHN and CIFAR-10 random crops with padding 4 as well as the quite aggressive augmentation AutoAugment \cite{cubuk2019autoaugment}. Additionally, we apply random horizontal flips for CIFAR-10.

\textbf{GOOD training procedure.}  As it is the case with IBP training \citep{gowal2018effectiveness} for certified adversarial robustness, we have observed that the inclusion of IBP bounds can make the training unstable or cause it to fail completely.
This can happen for our GOOD training despite the logarithmic damping in the $\LCUB$ loss in \eqref{eq:cub_loss}.
Thus, in order to further stabilize the training similar to \cite{gowal2018effectiveness}, we use linear ramp up schedules for $\epsilon$ and $\kappa$, which are detailed in Appendix \ref{section:experimental_details}.
As radii for the $l_\infty$-perturbation model on the out-distribution we use
$\epsilon = 0.3$ for MNIST, $\epsilon=0.03$ for SVHN and $\epsilon=0.01$ for CIFAR-10 (note that $0.01>\frac{2}{255}\approx 0.0078$). 
The chosen $\epsilon=0.01$ for CIFAR-10 is so small that the changes are hardly visible (see Figure \ref{fig:teaser}). As parameter $\kappa$ for the trade-off between cross-entropy loss and
the GOOD regularizer in \eqref{eq:objective_100} and \eqref{eq:objective_quantile}, we set $\kappa = 0.3$ for MNIST and $\kappa = 1$ for SVHN and CIFAR-10.

In order to explore the potential trade-off between the separation
of in- and out-distribution for clean and perturbed out-distribution
inputs (clean AUCs vs guaranteed AUCs - see below), we train GOOD
models for different quantiles $q \in [0,1]$ in \eqref{eq:objective_quantile}
which we denote as GOOD\textsubscript{$Q$} in the following.
Here, $Q = 100q$ is the percentage of out-distribution training samples for which we minimize the guaranteed upper bounds on the confidence of the neural network in the $l_\infty$-ball of radius $\epsilon$ around the out-distribution point during training.
Note that GOOD\textsubscript{100} corresponds to \eqref{eq:objective_100} where we minimize the guaranteed upper bound
on the worst-case confidence for all out-distribution samples, whereas GOOD\textsubscript{0} can be seen as a variant of OE or CEDA.
A training batch consists of 128 in- and 128 out-distribution samples. Examples of OOD training batches with the employed augmentation and their quantile splits for a  GOOD\textsubscript{60} model are shown in Table \ref{table:quantile_images} in the appendix.

For the training out-distribution, we use 80 Million Tiny Images (80M) \cite{torralba200880}, which is a large collection of natural images associated to nouns in wordnet \cite{fellbaum2012wordnet}.
All methods get the same out-distribution for training and 
we are \emph{neither} training \emph{nor} adapting hyperparameters for each
OOD dataset separately as in some previous work.
Since CIFAR-10 and CIFAR-100 are subsets of 80M, we follow \cite{HenMazDie2019} and filter them out. 
As can be seen in the example batches in Table \ref{table:quantile_images}, even this reduced dataset still contains images from CIFAR-10 classes, which explains why our quantile-based loss is essential for good performance on CIFAR-10.
We take a subset of 50 million images as OOD training set.
Since the size of the training set of the in-distribution datasets (MNIST: 60,000; SVHN: 73,257; CIFAR-10: 50000) 
is small compared to 50 million, typically an OOD image appears only once during training.

\textbf{Evaluation.} For each method, we compute the test accuracy
on the in-distribution task, and for various out-distribution datasets (not seen during training) we report the area under the receiver operating characteristic curve (AUC) as a measure for the separation of in- from out-distribution samples based on the predicted confidences on the test sets.
As OOD evaluation sets we use FashionMNIST~\citep{XiaoEtAl2017}, the Letters of EMNIST~\citep{CohEtAl2017}, grayscale CIFAR-10, and Uniform Noise for MNIST, and
CIFAR-100~\cite{krizhevsky2009learning}, CIFAR-10/SVHN, LSUN Classroom~\citep{lsun}, and Uniform Noise for SVHN/CIFAR-10.
Further evaluation on other OOD datasets can be found in Appendix \ref{section:additional_datasets}.

We are particularly interested in the worst case OOD detection performance of all methods under the $l_\infty$-perturbation model for the out-distribution. For this purpose, we compute the \textbf{adversarial AUC (AAUC)} and the \textbf{guaranteed AUC (GAUC)}. 
These AUCs are based on the maximal confidence in the $l_\infty$-ball of radius $\epsilon$ around each out-distribution image. 
For the adversarial AUC, we compute a lower bound on the maximal confidence in the $l_\infty$-ball by using Auto-PGD
\cite{CroHei2020}
for maximizing the confidence of the classifier inside the intersection of the $l_\infty$- ball and the image domain $[0,1]^d$. Auto-PGD uses an automatic stepsize selection scheme and has been shown to outperform PGD. We use an adaptation to our setting (described in detail in Appendix \ref{section:adversarial}) with 500 steps and 5 restarts on 1000 points from each test set. Gradient masking poses a significant challenge, so we also perform a transfer attack on all models and on MNIST, we even use an additional attack (see Appendix \ref{section:adversarial}). We report the per-sample worst-case across attacks. Note that attacking these models on different out-distributions poses somewhat different challenges than classical adversarial attacks. Around the in-distribution models with good prediction performance are unlikely to be completely flat (and thus have zero gradient) in the whole region defined by an $l_\infty$-threat model. On the out-distribution, however, it is quite possible that all neurons in some layer return negative pre-activations which causes all gradients to be zero. Therefore the choice of initialization together with several restarts matters a lot as otherwise non-robust OOD detection models can easily appear to be robust. Moreover, the transfer attacks were necessary for some methods as otherwise the true robustness would have been significantly overestimated. Indeed even though we invested quite some effort into adaptive attacks which are specific for our robust OOD detection scenario, it might still be that the AAUC of some methods is overestimated. This again shows how important it is to get provable guarantees.

For the guaranteed AUC, we compute an upper bound on the confidence in the intersection
of the $l_\infty$- ball with the image domain $[0,1]^d$ via IBP
using \eqref{eq:softmaxub} for the full test set.
These worst case/guaranteed confidences for the out-distributions are then used for the AUC computation.

\textbf{Competitors.} We compare a normally trained model (Plain), the state-of-the-art OOD detection
method Outlier Exposure (OE) \cite{HenMazDie2019}, 
CEDA~\cite{HeiAndBit2019}  and
Adversarial Confidence Enhanced Training (ACET)~\cite{HeiAndBit2019},
which we adjusted to the given task as described in the appendix. As CEDA performs very similar to OE, we omit it in the figures for better readability.
The $\epsilon$-radii for the
$l_\infty$-balls are the same for ACET and GOOD.
So far the only method which could provide robustness guarantees for OOD detection is Certified Certain Uncertainty (CCU) with a data-dependent Mahalanobis-type $l_2$ threat model. We use their publicly available code to train a CCU model with our architecture and we evaluate their guarantees for our $l_\infty$ threat model. In Appendix \ref{section:OOD_conf}, we provide details and explain why their guarantees turn out to be vacuous in our setting.

%% file: sections/results.tex
\subsection{Results}

In Table \ref{table:all_aucs} we present the results on all datasets.
\input{tables/table_all_aucs}

\textbf{GOOD is provably better than OE/CEDA with regard to worst
case OOD detection.}
We note that for almost all OOD datasets GOOD achieves non-trivial GAUCs. Thus the guarantees generalize from the training out-distribution 80M to the test OOD datasets. For the easier in-distributions MNIST and SVHN, which are more clearly separated from the out-distribution, the overall best results are achieved for GOOD\textsubscript{100}. 
For CIFAR-10, the clean AUCs of GOOD\textsubscript{100} are low even when compared to plain training.
Arguably the best trade-off for CIFAR-10 is achieved by GOOD\textsubscript{80}. Note that the guaranteed AUC (GAUC) of these models is always better than the adversarial AUC (AAUC) of OE/CEDA (except for EMNIST). Thus it is fair to say that the worst-case OOD detection performance of GOOD is provably better than that of OE/CEDA.
As expected,
ACET yields good AAUCs but has no guarantees. The failure of CCU regarding guarantees is discussed in Appendix \ref{section:OOD_conf}.
It is notable that GOOD\textsubscript{100} has close to perfect guaranteed OOD detection performance for MNIST on CIFAR-10/uniform noise and for SVHN on \textbf{all} out-distribution datasets. In Appendix \ref{section:bigger_bounds} we show that the guarantees of GOOD generalize surprisingly well to larger radii than seen during training.

\textbf{GOOD achieves certified OOD performance with almost no loss in accuracy.} While there is a small drop in clean accuracy for MNIST, on SVHN, with $96.3\%$ GOOD\textsubscript{100} has a better clean accuracy than all competing methods. On CIFAR-10, GOOD\textsubscript{80} achieves an accuracy of $90.1\%$ which is better than ACET and only slightly worse than CCU and OE. 
This is remarkable as we are not aware of any model with certified \emph{adversarial robustness on the in-distribution} which gets even close to this range; e.g. IBP \cite{gowal2018effectiveness} achieves an accuracy of 85.2\% on SVHN with $\epsilon=0.01$ (we have 96.3\%), on CIFAR-10 with $\epsilon=\frac{2}{255}$ they get 71.2\% (we have 90.1\%). Previous certified methods had even worse clean accuracy. Since a significant loss in prediction performance is usually not acceptable, certified methods have not yet had much practical impact. Thus we think it is an encouraging and interesting observation that properties different from adversarial robustness like worst-case out-of-distribution detection can be certified without suffering much in accuracy. In particular, it is quite surprising that certified methods can be trained effectively with aggressive data augmentation like AutoAugment.

\textbf{Trade-off between clean and guaranteed AUC via Quantile-GOOD.}
As discussed above, for the CIFAR-10 experiments, our training out-distribution contains images from in-distribution classes.
This seems to be the reason why GOOD\textsubscript{100} suffers from a significant drop in clean AUC, as the only way to ensure small loss $\LCUB$, if in- and out-distribution can partially not be distinguished, is to reduce also the confidence on the in-distribution.
This conflict is resolved via GOOD\textsubscript{80} and GOOD\textsubscript{90} which both have better clean AUCs. It is an interesting open question if similar trade-offs can also be useful for certified adversarial robustness.

\textbf{EMNIST: distinguishing letters from digits without ever having seen letters.}
GOOD\textsubscript{100} achieves an excellent AUC of 99.0\% for the letters of EMNIST which is, up to our knowledge, state-of-the-art.
Indeed, an AUC of 100\% should not be expected as even for humans some letters like i and l are indistinguishable from digits.
This result is quite remarkable as GOOD\textsubscript{100} has never seen letters during training.
Moreover, as the AUC just distinguishes the separation of in- and out-distribution based on the confidence, we provide the mean confidence on all datasets in the Appendix in Table~\ref{tab:MMC} and in Figure~\ref{Fig:Samples} (see also Figure~\ref{Fig:Samples_cont} in the Appendix) we show some samples from EMNIST together with their prediction/confidences for all models. GOOD\textsubscript{100} has a mean confidence of $98.4\%$ on MNIST but only $27.1\%$ on EMNIST in contrast to ACET with
$75.0\%$, OE $87.9\%$ and Plain $91.5\%$. This shows that while the AUC's of ACET and OE are good for EMNIST, these methods are still highly overconfident on EMNIST. Only GOOD\textsubscript{100} produces meaningful higher confidences on EMNIST, when the letter has clear features of the corresponding digit.

\input{figures/samples.tex}

\section{Conclusion}
We propose GOOD, a novel training method to achieve guaranteed OOD detection in a worst-case setting. GOOD provably outperforms OE,
the state-of-the-art in OOD detection, in worst case OOD detection and has state-of-the-art performance on EMNIST which is a particularly challenging out-distribution dataset. As the test accuracy of GOOD is comparable to the one of normal training, this shows that certified methods have the potential to be useful in practice even for more complex tasks. In future work it will be interesting to explore how close certified methods
can get to state-of-the-art test performance.

%% file: tables/table_all_aucs.tex
\begin{table*}[!htbp]
\caption{Accuracies as well as AUC, adversarial AUC (AAUC) and guaranteed AUC (GAUC) values for the MNIST, SVHN and CIFAR-10 in-distributions with respect to several unseen out-distributions. 
The GAUC of GOOD\textsubscript{100} on MNIST/SVHN resp. GOOD\textsubscript{80} on CIFAR-10 is better than the corresponding AAUC of OE and CEDA on almost all OOD datsets (except EMNIST). Thus GOOD is provably better than OE and CEDA w.r.t. worst-case OOD detection. GOOD achieves this without significant loss in accuracy. Especially on SVHN, GOOD\textsubscript{100} has very good accuracy and almost perfect provably worst-case OOD detection performance.
}
\label{table:all_aucs}
\setlength\tabcolsep{.5pt} 
\vskip 0.15in
\begin{center}
\begin{small}
\begin{sc}
\makebox[\textwidth][c]{
\noindent\begin{tabularx}{1.0\textwidth}{lc|CCC|CCC|CCC|CCC}
\toprule
\multicolumn{14}{c}{in: MNIST \hspace{.8cm} $\epsilon = 0.3$} \\
\midrule
\multicolumn{1}{l}{\multirow{2}{*}{Method \ }}  & \multicolumn{1}{c|}{\multirow{2}{*}{\ Acc. \ }}
 & \multicolumn{3}{c|}{FashionMNIST}  & \multicolumn{3}{c|}{EMNIST Letters}  & \multicolumn{3}{c|}{CIFAR-10} & \multicolumn{3}{c}{Uniform Noise}  \\
& & auc & aauc & gauc & auc & aauc & gauc & auc & aauc & gauc & auc & aauc & gauc \\ \midrule
Plain                 & \noindent\phantom{0}99.4 & \noindent\phantom{0}98.0  & \noindent\phantom{0}34.2 & \noindent\phantom{00}0.0 & \noindent\phantom{0}88.0 & \noindent\phantom{0}31.4 & \noindent\phantom{00}0.0 & \noindent\phantom{0}98.8 & \noindent\phantom{0}36.6 & \noindent\phantom{00}0.0 & \noindent\phantom{0}99.1 & \noindent\phantom{0}36.5 & \noindent\phantom{00}0.0 \\
CEDA                  & \noindent\phantom{0}99.4 & \noindent\phantom{0}99.9  & \noindent\phantom{0}82.1 & \noindent\phantom{00}0.0 & \noindent\phantom{0}92.6 & \noindent\phantom{0}52.8 & \noindent\phantom{00}0.0 & \textbf{100.0}             & \noindent\phantom{0}95.1 & \noindent\phantom{00}0.0 & \textbf{100.0}             & \textbf{100.0}             & \noindent\phantom{00}0.0 \\
OE                    & \noindent\phantom{0}99.4 & \noindent\phantom{0}99.9  & \noindent\phantom{0}76.8 & \noindent\phantom{00}0.0 & \noindent\phantom{0}92.7 & \noindent\phantom{0}50.9 & \noindent\phantom{00}0.0 & \textbf{100.0}             & \noindent\phantom{0}92.4 & \noindent\phantom{00}0.0 & \textbf{100.0}             & \textbf{100.0}             & \noindent\phantom{00}0.0 \\
ACET                  & \noindent\phantom{0}99.4 & \textbf{100.0}            & \noindent\phantom{0}\textbf{98.4} & \noindent\phantom{00}0.0 & \noindent\phantom{0}95.9 & \noindent\phantom{0}61.5 & \noindent\phantom{00}0.0 & \textbf{100.0}             & \noindent\phantom{0}99.3 & \noindent\phantom{00}0.0 & \textbf{100.0}             & \textbf{100.0}             & \noindent\phantom{00}0.0 \\
CCU                   &  \noindent\phantom{0}\textbf{99.5} & \textbf{100.0}  & \noindent\phantom{0}76.6 & \noindent\phantom{00}0.0 & \noindent\phantom{0}92.9 & \noindent\phantom{00}3.1 & \noindent\phantom{00}0.0 & \textbf{100.0}             & \noindent\phantom{0}98.9 & \noindent\phantom{00}0.0 & \textbf{100.0}             & \textbf{100.0}             & \noindent\phantom{00}0.0 \\
GOOD\textsubscript{0} & \noindent\phantom{0}99.5 & \noindent\phantom{0}99.9  & \noindent\phantom{0}82.3 & \noindent\phantom{00}0.0 & \noindent\phantom{0}92.9 & \noindent\phantom{0}55.0 & \noindent\phantom{00}0.0 & \textbf{100.0}             & \noindent\phantom{0}94.7 & \noindent\phantom{00}0.0 & \textbf{100.0}             & \textbf{100.0}             & \noindent\phantom{00}0.0 \\
GOOD\textsubscript{20} & \noindent\phantom{0}99.0 & \noindent\phantom{0}99.8 & \noindent\phantom{0}88.2 & \noindent\phantom{00}9.7 & \noindent\phantom{0}95.3 & \noindent\phantom{0}54.3 & \noindent\phantom{00}0.0 & \textbf{100.0}             & \noindent\phantom{0}97.6 & \noindent\phantom{0}28.3 & \textbf{100.0}             &\textbf{ 100.0}             & \textbf{100.0}             \\
GOOD\textsubscript{40} & \noindent\phantom{0}99.0 & \noindent\phantom{0}99.8 & \noindent\phantom{0}88.0 & \noindent\phantom{0}29.1 & \noindent\phantom{0}95.7 & \noindent\phantom{0}56.6 & \noindent\phantom{00}0.0 & \textbf{100.0}             & \noindent\phantom{0}97.7 & \noindent\phantom{0}65.2 & \textbf{100.0}             & \textbf{100.0}             & \textbf{100.0}             \\
GOOD\textsubscript{60} & \noindent\phantom{0}99.0 & \noindent\phantom{0}99.9 & \noindent\phantom{0}88.8 & \noindent\phantom{0}42.0 & \noindent\phantom{0}96.6 & \noindent\phantom{0}57.9 & \noindent\phantom{00}0.1 & \textbf{100.0}             & \noindent\phantom{0}97.9 & \noindent\phantom{0}85.3 & \textbf{100.0}             & \textbf{100.0}             & \textbf{100.0}             \\
GOOD\textsubscript{80} & \noindent\phantom{0}99.1 & \noindent\phantom{0}99.8 & \noindent\phantom{0}90.3 & \noindent\phantom{0}55.5 & \noindent\phantom{0}97.9 & \noindent\phantom{0}\textbf{63.1} & \noindent\phantom{00}3.4 & \textbf{100.0}             & \noindent\phantom{0}98.4 & \noindent\phantom{0}94.7 & \textbf{100.0}             & \textbf{100.0}             & \textbf{100.0}    \\
GOOD\textsubscript{90} & \noindent\phantom{0}98.8 & \noindent\phantom{0}99.9 & \noindent\phantom{0}91.4 & \noindent\phantom{0}66.9 & \noindent\phantom{0}98.0 & \noindent\phantom{0}59.4 & \noindent\phantom{00}5.1 & \textbf{100.0}             & \noindent\phantom{0}99.0 & \noindent\phantom{0}97.8 & \textbf{100.0}             & \textbf{100.0}             & \textbf{100.0}             \\
GOOD\textsubscript{95} & \noindent\phantom{0}98.8 & \noindent\phantom{0}99.9 & \noindent\phantom{0}93.1 & \noindent\phantom{0}73.9 & \noindent\phantom{0}98.7 & \noindent\phantom{0}59.2 & \noindent\phantom{00}\textbf{5.6} & \textbf{100.0}             & \noindent\phantom{0}99.4 & \noindent\phantom{0}98.8 & \textbf{100.0}             & \textbf{100.0}             & \textbf{100.0}    \\
\rowcolor{lightgrey} GOOD\textsubscript{100} & \noindent\phantom{0}98.7 & \textbf{100.0}          & \noindent\phantom{0}96.5 & \noindent\phantom{0}\textbf{78.0} & \noindent\phantom{0}\textbf{99.0} & \noindent\phantom{0}53.8 & \noindent\phantom{00}3.3 & \textbf{100.0} & \noindent\phantom{0}\textbf{99.9} & \noindent\phantom{0}\textbf{99.4} & \textbf{100.0}  & \textbf{100.0}  & \textbf{100.0}  \\
\bottomrule
\toprule
\multicolumn{14}{c}{in: SVHN \hspace{1cm} $\epsilon = 0.03$} \\
\midrule
\multicolumn{1}{l}{\multirow{2}{*}{Method \ }}  & \multicolumn{1}{c|}{\multirow{2}{*}{\ Acc. \ }}
 & \multicolumn{3}{c|}{CIFAR-100}  & \multicolumn{3}{c|}{CIFAR-10}  & \multicolumn{3}{c|}{LSUN Classroom} & \multicolumn{3}{c}{Uniform Noise}  \\
& & auc & aauc & gauc & auc & aauc & gauc & auc & aauc & gauc & auc & aauc & gauc \\ \midrule
Plain                  & \noindent\phantom{0}95.5 & \noindent\phantom{0}94.9 & \noindent\phantom{0}11.3 & \noindent\phantom{00}0.0 & \noindent\phantom{0}95.2 & \noindent\phantom{0}11.1 & \noindent\phantom{00}0.0 & \noindent\phantom{0}95.7 & \noindent\phantom{0}14.1 & \noindent\phantom{00}0.0 & \noindent\phantom{0}99.4 & \noindent\phantom{0}57.9 & \noindent\phantom{00}0.0 \\
CEDA                   & \noindent\phantom{0}95.3 & \noindent\phantom{0}99.9 & \noindent\phantom{0}63.9 & \noindent\phantom{00}0.0 & \noindent\phantom{0}99.9 & \noindent\phantom{0}68.7 & \noindent\phantom{00}0.0 & \noindent\phantom{0}99.9 & \noindent\phantom{0}80.7 & \noindent\phantom{00}0.0 & \noindent\phantom{0}99.9 & \noindent\phantom{0}99.3 & \noindent\phantom{00}0.0 \\
OE                     & \noindent\phantom{0}95.5 & \textbf{100.0}           & \noindent\phantom{0}60.2 & \noindent\phantom{00}0.0 & \textbf{100.0}             & \noindent\phantom{0}62.5        & \noindent\phantom{00}0.0 & \textbf{100.0}       & \noindent\phantom{0}77.3 & \noindent\phantom{00}0.0 & \textbf{100.0}  & \noindent\phantom{0}98.2 & \noindent\phantom{00}0.0 \\
ACET                   & \noindent\phantom{0}96.0 & \textbf{100.0}           & \noindent\phantom{0}\textbf{99.4} & \noindent\phantom{00}0.0 & \textbf{100.0}             & \noindent\phantom{0}\textbf{99.5}        & \noindent\phantom{00}0.0 & \textbf{100.0} & \noindent\phantom{0}\textbf{99.8} & \noindent\phantom{00}0.0 & \noindent\phantom{0}99.9 & \noindent\phantom{0}96.3 & \noindent\phantom{00}0.0 \\
CCU                    & \noindent\phantom{0}95.7 & \textbf{100.0}           & \noindent\phantom{0}52.5 & \noindent\phantom{00}0.0 & \textbf{100.0}             & \noindent\phantom{0}56.8        & \noindent\phantom{00}0.0  & \textbf{100.0}   & \noindent\phantom{0}72.1  & \noindent\phantom{00}0.0   & \textbf{100.0}   &\textbf{100.0}         & \noindent\phantom{00}0.0   \\
GOOD\textsubscript{0}  & \noindent\phantom{0}\textbf{97.0} & \textbf{100.0}           & \noindent\phantom{0}61.0 & \noindent\phantom{00}0.0 & \textbf{100.0}             & \noindent\phantom{0}60.0        & \noindent\phantom{00}0.0 & \textbf{100.0}       & \noindent\phantom{0}60.8 & \noindent\phantom{00}0.0 & \textbf{100.0}             & \noindent\phantom{0}82.5 & \noindent\phantom{00}0.0 \\
GOOD\textsubscript{20} & \noindent\phantom{0}95.9 & \noindent\phantom{0}99.8 & \noindent\phantom{0}78.2 & \noindent\phantom{0}24.4 & \noindent\phantom{0}99.9 & \noindent\phantom{0}81.8 & \noindent\phantom{0}20.3 & \noindent\phantom{0}99.9 & \noindent\phantom{0}91.2 & \noindent\phantom{0}21.6 & \noindent\phantom{0}99.7 & \noindent\phantom{0}99.5 & \noindent\phantom{0}99.5 \\
GOOD\textsubscript{40} & \noindent\phantom{0}96.3 & \noindent\phantom{0}99.5 & \noindent\phantom{0}81.6 & \noindent\phantom{0}46.0 & \noindent\phantom{0}99.5 & \noindent\phantom{0}85.0 & \noindent\phantom{0}50.6 & \noindent\phantom{0}99.5 & \noindent\phantom{0}95.1 & \noindent\phantom{0}55.7 & \noindent\phantom{0}99.5 & \noindent\phantom{0}99.5 & \noindent\phantom{0}99.4 \\
GOOD\textsubscript{60} & \noindent\phantom{0}96.1 & \noindent\phantom{0}99.4 & \noindent\phantom{0}83.9 & \noindent\phantom{0}67.4 & \noindent\phantom{0}99.4 & \noindent\phantom{0}87.4 & \noindent\phantom{0}72.9 & \noindent\phantom{0}99.4 & \noindent\phantom{0}96.5 & \noindent\phantom{0}82.3 & \noindent\phantom{0}99.4 & \noindent\phantom{0}99.4 & \noindent\phantom{0}99.4 \\
GOOD\textsubscript{80} & \noindent\phantom{0}96.3 & \textbf{100.0}           & \noindent\phantom{0}93.5 & \noindent\phantom{0}87.7 & \textbf{100.0}             & \noindent\phantom{0}95.3 & \noindent\phantom{0}91.3 & \textbf{100.0}        & \noindent\phantom{0}98.8 & \noindent\phantom{0}96.7 & \textbf{100.0}             &\textbf{100.0}             & \noindent\phantom{0}99.7 \\
GOOD\textsubscript{90} & \noindent\phantom{0}96.2 & \noindent\phantom{0}99.8 & \noindent\phantom{0}96.0 & \noindent\phantom{0}93.9 & \noindent\phantom{0}99.8 & \noindent\phantom{0}97.3 & \noindent\phantom{0}96.1 & \noindent\phantom{0}99.8 & \noindent\phantom{0}98.9 & \noindent\phantom{0}98.3 & \noindent\phantom{0}99.8 & \noindent\phantom{0}99.8 & \noindent\phantom{0}\textbf{99.8} \\
GOOD\textsubscript{95} & \noindent\phantom{0}96.4 & \noindent\phantom{0}99.8 & \noindent\phantom{0}97.2 & \noindent\phantom{0}96.1 & \noindent\phantom{0}99.8 & \noindent\phantom{0}98.0 & \noindent\phantom{0}97.3 & \noindent\phantom{0}99.8 & \noindent\phantom{0}99.3 & \noindent\phantom{0}\textbf{98.9} & \noindent\phantom{0}99.9 & \noindent\phantom{0}99.9 & \noindent\phantom{0}\textbf{99.8} \\
\rowcolor{lightgrey} GOOD\textsubscript{100} & \noindent\phantom{0}96.3 & \noindent\phantom{0}99.6 & \noindent\phantom{0}97.7 & \noindent\phantom{0}\textbf{97.3} & \noindent\phantom{0}99.7 & \noindent\phantom{0}98.4 & \noindent\phantom{0}\textbf{98.1} & \noindent\phantom{0}99.9 & \noindent\phantom{0}99.2 & \noindent\phantom{0}\textbf{98.9} & \textbf{100.0}   & \noindent\phantom{0}99.9 & \noindent\phantom{0}\textbf{99.8} \\
\bottomrule
\toprule
\multicolumn{14}{c}{in: CIFAR-10 \hspace{.5cm} $\epsilon = 0.01$} \\
\midrule
\multicolumn{1}{l}{\multirow{2}{*}{Method \ }}  & \multicolumn{1}{c|}{\multirow{2}{*}{\ Acc. \ }}
 & \multicolumn{3}{c|}{CIFAR-100}  & \multicolumn{3}{c|}{SVHN}  & \multicolumn{3}{c|}{LSUN Classroom} & \multicolumn{3}{c}{Uniform Noise}  \\
& & auc & aauc & gauc & auc & aauc & gauc & auc & aauc & gauc & auc & aauc & gauc \\ \midrule
Plain & \noindent\phantom{0}90.1 & \noindent\phantom{0}84.3 & \noindent\phantom{0}13.0 & \noindent\phantom{00}0.0 & \noindent\phantom{0}87.7 & \noindent\phantom{0}10.6 & \noindent\phantom{00}0.0 & \noindent\phantom{0}88.9 & \noindent\phantom{0}13.6 & \noindent\phantom{00}0.0 & \noindent\phantom{0}90.8 & \noindent\phantom{0}56.4 & \noindent\phantom{00}0.0 \\
CEDA & \noindent\phantom{0}88.6 & \noindent\phantom{0}91.8 & \noindent\phantom{0}31.9 & \noindent\phantom{00}0.0 & \noindent\phantom{0}\textbf{97.9} & \noindent\phantom{0}25.7 & \noindent\phantom{00}0.0 & \noindent\phantom{0}98.9 & \noindent\phantom{0}53.9 & \noindent\phantom{00}0.0 & \noindent\phantom{0}97.3 & \noindent\phantom{0}70.5 & \noindent\phantom{00}0.0 \\
OE & \noindent\phantom{0}90.7 & \noindent\phantom{0}92.4 & \noindent\phantom{0}11.0 & \noindent\phantom{00}0.0 & \noindent\phantom{0}97.6 & \noindent\phantom{00}3.7 & \noindent\phantom{00}0.0 & \noindent\phantom{0}98.9 & \noindent\phantom{0}20.0 & \noindent\phantom{00}0.0 & \noindent\phantom{0}98.7 & \noindent\phantom{0}75.7 & \noindent\phantom{00}0.0 \\
ACET & \noindent\phantom{0}89.3 & \noindent\phantom{0}90.7 & \noindent\phantom{0}\textbf{74.5} & \noindent\phantom{00}0.0 & \noindent\phantom{0}96.6 & \noindent\phantom{0}\textbf{88.0} & \noindent\phantom{00}0.0 & \noindent\phantom{0}98.3 & \noindent\phantom{0}\textbf{91.2} & \noindent\phantom{00}0.0 & \noindent\phantom{0}99.7 & \noindent\phantom{0}98.9 & \noindent\phantom{00}0.0 \\
CCU & \noindent\phantom{0}\textbf{91.6} & \noindent\phantom{0}\textbf{93.0} & \noindent\phantom{0}23.3 & \noindent\phantom{00}0.0 & \noindent\phantom{0}97.1 & \noindent\phantom{0}14.8 & \noindent\phantom{00}0.0 & \noindent\phantom{0}\textbf{99.3} & \noindent\phantom{0}38.2 & \noindent\phantom{00}0.0 &\textbf{100.0} & \textbf{100.0} & \noindent\phantom{00}0.0 \\
GOOD\textsubscript{0} & \noindent\phantom{0}89.8 & \noindent\phantom{0}92.9 & \noindent\phantom{0}22.5 & \noindent\phantom{00}0.0 & \noindent\phantom{0}97.0 & \noindent\phantom{0}12.8 & \noindent\phantom{00}0.0 & \noindent\phantom{0}98.3 & \noindent\phantom{0}48.4 & \noindent\phantom{00}0.0 & \noindent\phantom{0}96.3 & \noindent\phantom{0}95.6 & \noindent\phantom{00}0.0 \\
GOOD\textsubscript{20} & \noindent\phantom{0}88.5 & \noindent\phantom{0}90.3 & \noindent\phantom{0}32.4 & \noindent\phantom{0}11.8 & \noindent\phantom{0}95.9 & \noindent\phantom{0}28.3 & \noindent\phantom{0}15.8 & \noindent\phantom{0}98.2 & \noindent\phantom{0}48.2 & \noindent\phantom{00}3.4 & \noindent\phantom{0}99.4 & \noindent\phantom{0}97.6 & \noindent\phantom{0}87.5 \\
GOOD\textsubscript{40} & \noindent\phantom{0}89.5 & \noindent\phantom{0}89.6 & \noindent\phantom{0}38.2 & \noindent\phantom{0}24.8 & \noindent\phantom{0}95.4 & \noindent\phantom{0}38.0 & \noindent\phantom{0}24.9 & \noindent\phantom{0}96.0 & \noindent\phantom{0}62.0 & \noindent\phantom{0}27.4 & \noindent\phantom{0}92.1 & \noindent\phantom{0}89.9 & \noindent\phantom{0}89.8 \\
GOOD\textsubscript{60} & \noindent\phantom{0}90.2 & \noindent\phantom{0}88.6 & \noindent\phantom{0}42.6 & \noindent\phantom{0}34.9 & \noindent\phantom{0}95.6 & \noindent\phantom{0}44.4 & \noindent\phantom{0}39.0 & \noindent\phantom{0}97.0 & \noindent\phantom{0}67.6 & \noindent\phantom{0}49.1 & \noindent\phantom{0}91.8 & \noindent\phantom{0}91.3 & \noindent\phantom{0}91.2 \\
\rowcolor{lightgrey} GOOD\textsubscript{80} & \noindent\phantom{0}90.1 & \noindent\phantom{0}85.6 & \noindent\phantom{0}48.2 & \noindent\phantom{0}42.3 & \noindent\phantom{0}94.0 & \noindent\phantom{0}41.4 & \noindent\phantom{0}38.0 & \noindent\phantom{0}93.3 & \noindent\phantom{0}66.9 & \noindent\phantom{0}55.2 & \noindent\phantom{0}95.8 & \noindent\phantom{0}95.4 & \noindent\phantom{0}95.3 \\
GOOD\textsubscript{90} & \noindent\phantom{0}90.2 & \noindent\phantom{0}81.7 & \noindent\phantom{0}51.5 & \noindent\phantom{0}49.6 & \noindent\phantom{0}91.4 & \noindent\phantom{0}48.7 & \noindent\phantom{0}46.9 & \noindent\phantom{0}90.2 & \noindent\phantom{0}63.5 & \noindent\phantom{0}57.7 & \noindent\phantom{0}89.3 & \noindent\phantom{0}87.7 & \noindent\phantom{0}87.7 \\
GOOD\textsubscript{95} & \noindent\phantom{0}90.4 & \noindent\phantom{0}80.3 & \noindent\phantom{0}52.0 & \noindent\phantom{0}50.8 & \noindent\phantom{0}90.2 & \noindent\phantom{0}44.4 & \noindent\phantom{0}43.3 & \noindent\phantom{0}88.3 & \noindent\phantom{0}62.6 & \noindent\phantom{0}60.3 & \noindent\phantom{0}96.6 & \noindent\phantom{0}95.9 & \noindent\phantom{0}95.8 \\
GOOD\textsubscript{100} & \noindent\phantom{0}90.1 & \noindent\phantom{0}70.0 & \noindent\phantom{0}54.7 & \noindent\phantom{0}\textbf{54.2} & \noindent\phantom{0}75.5 & \noindent\phantom{0}58.9 & \noindent\phantom{0}\textbf{56.9} & \noindent\phantom{0}75.2 & \noindent\phantom{0}61.5 & \noindent\phantom{0}\textbf{61.0} & \noindent\phantom{0}99.5 & \noindent\phantom{0}99.2 & \noindent\phantom{0}\textbf{99.0} \\
\bottomrule
\end{tabularx}
}
\end{sc}
\end{small}
\end{center}
\vskip -0.1in
\end{table*}




%% file: figures/samples.tex
\newcommand{\plotwidth}{.08\textwidth}
\begin{figure*}
\centering
{\footnotesize
\setlength{\tabcolsep}{0.9mm}
\hspace{-.6cm}
\begin{tabular}{lllllllllll}

\includegraphics[width=\plotwidth]{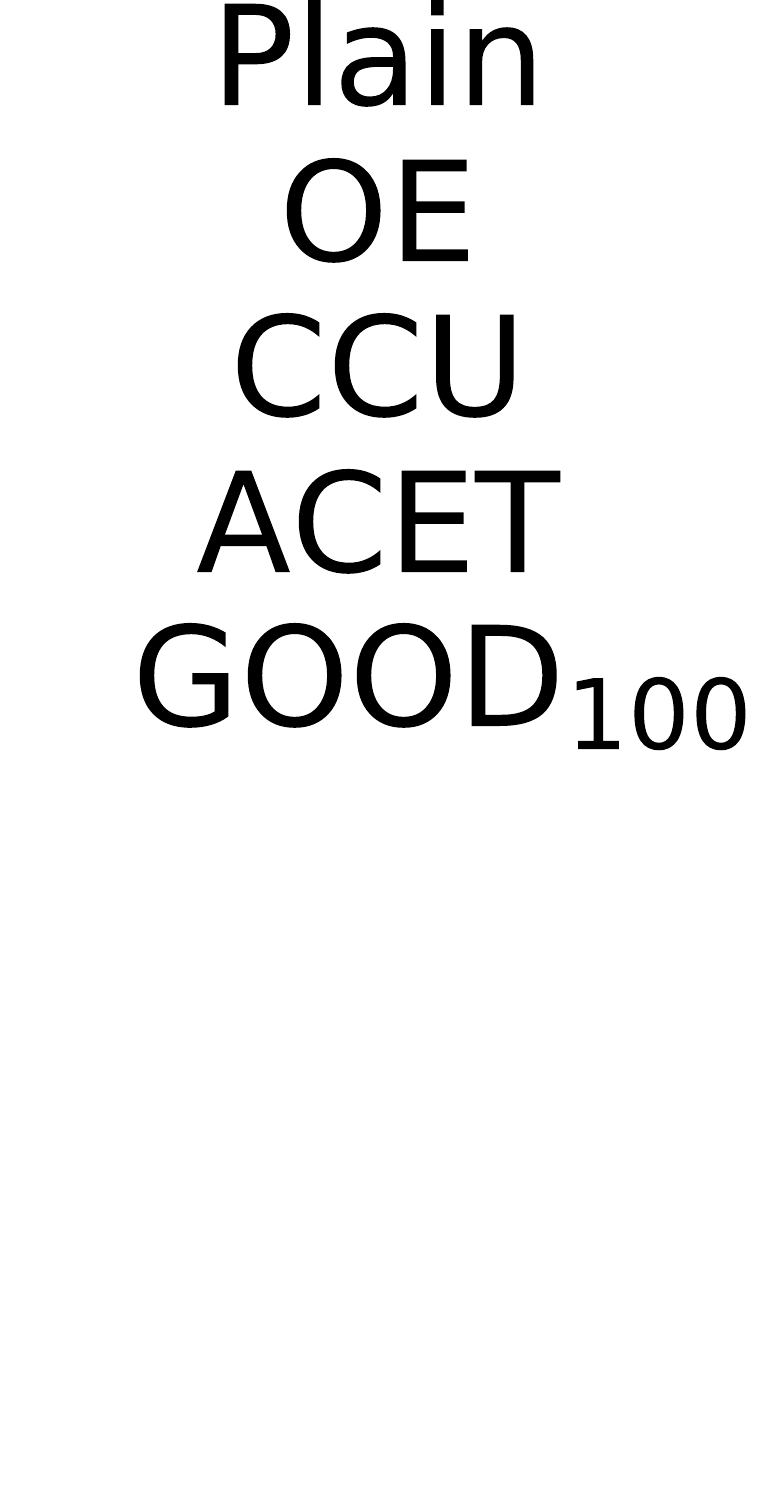} & \includegraphics[width=\plotwidth]{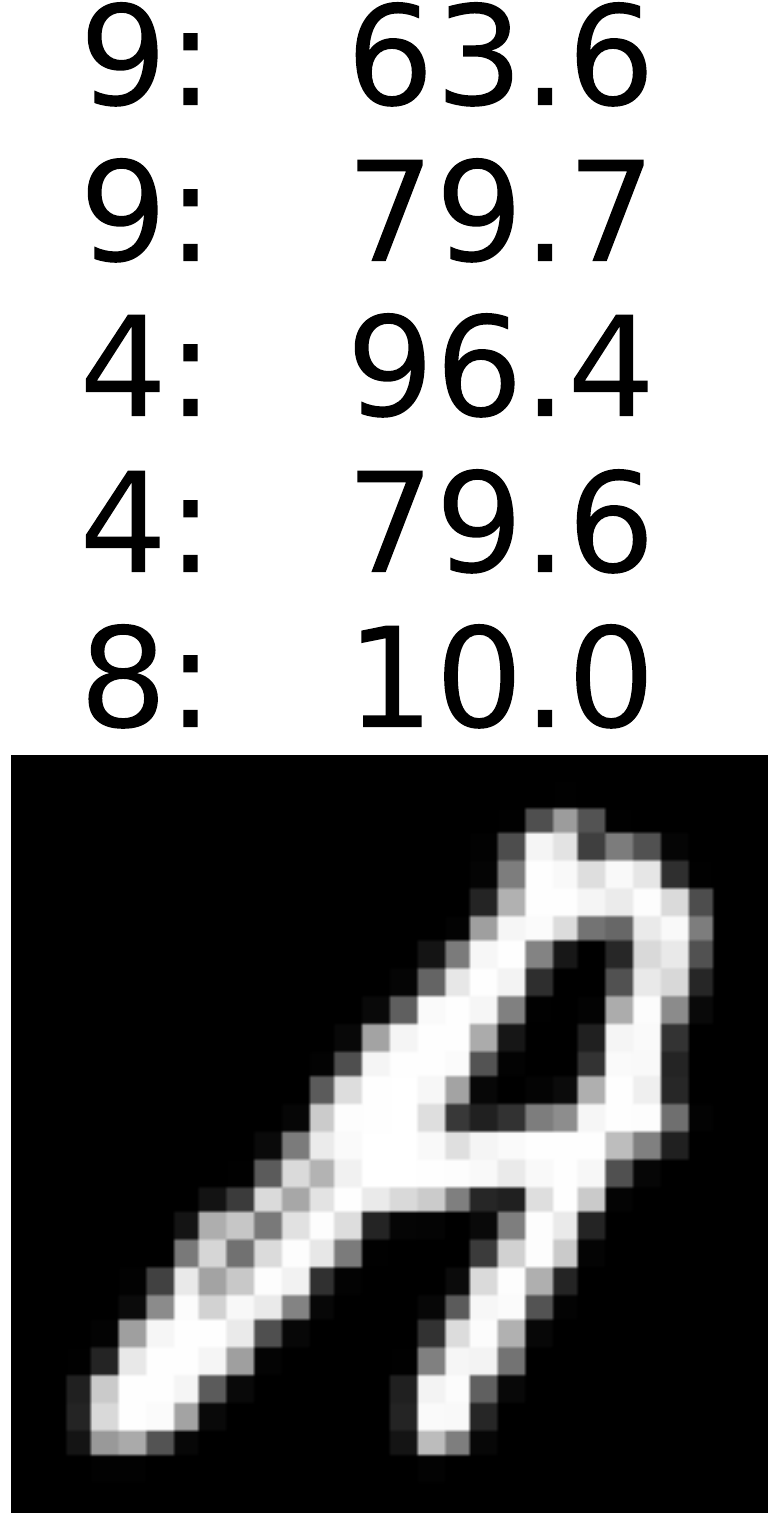} & \includegraphics[width=\plotwidth]{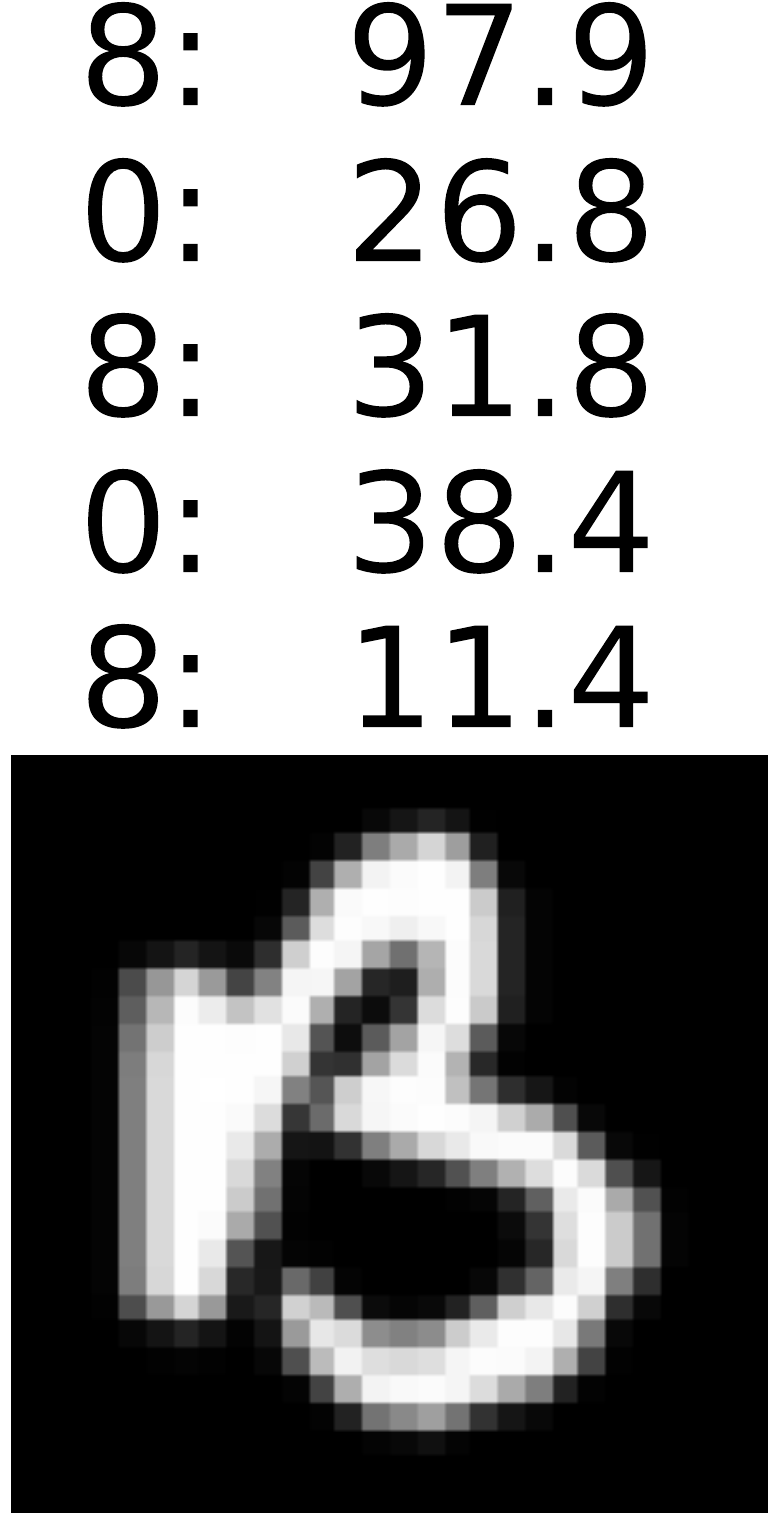} & 
\includegraphics[width=\plotwidth]{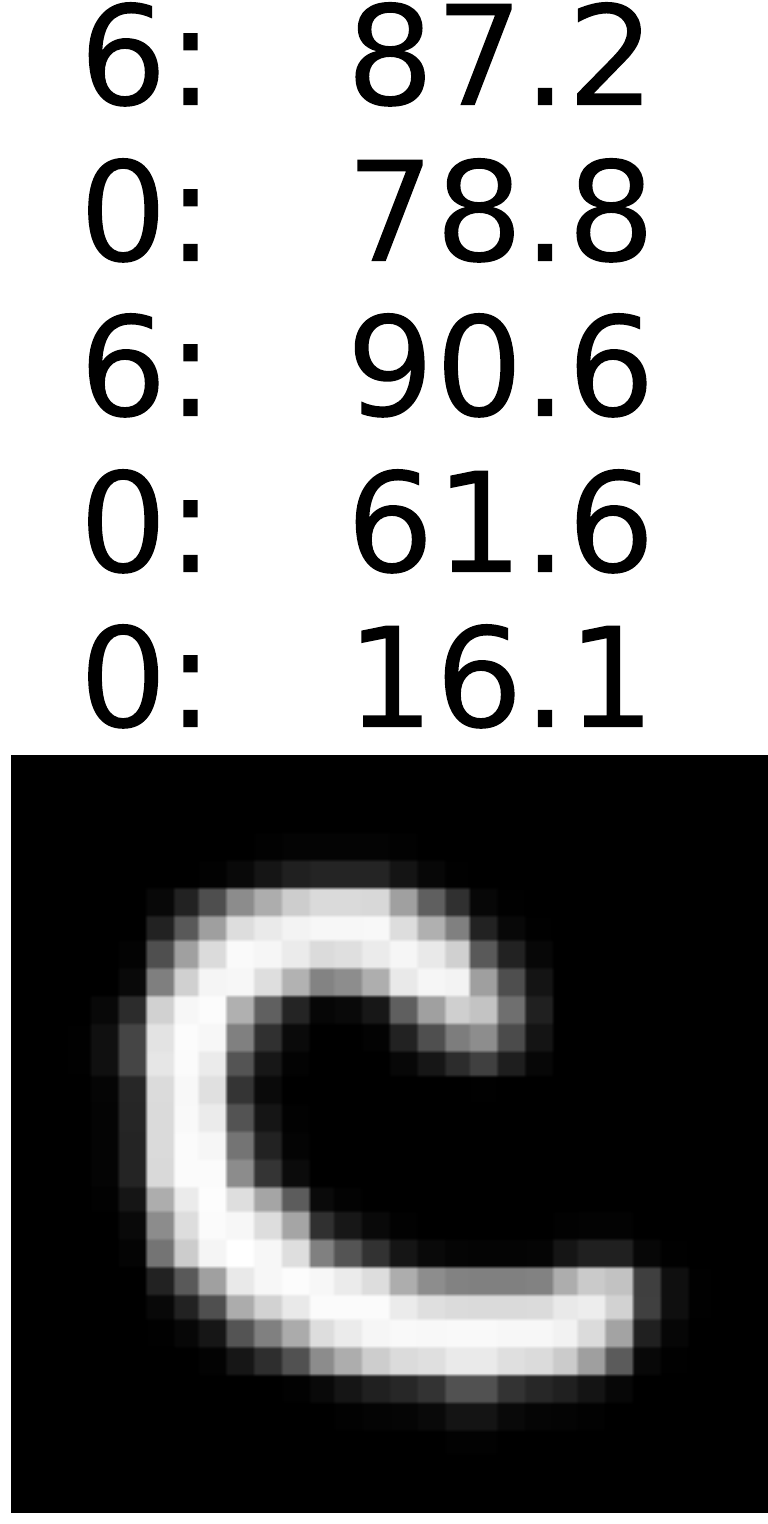} & \includegraphics[width=\plotwidth]{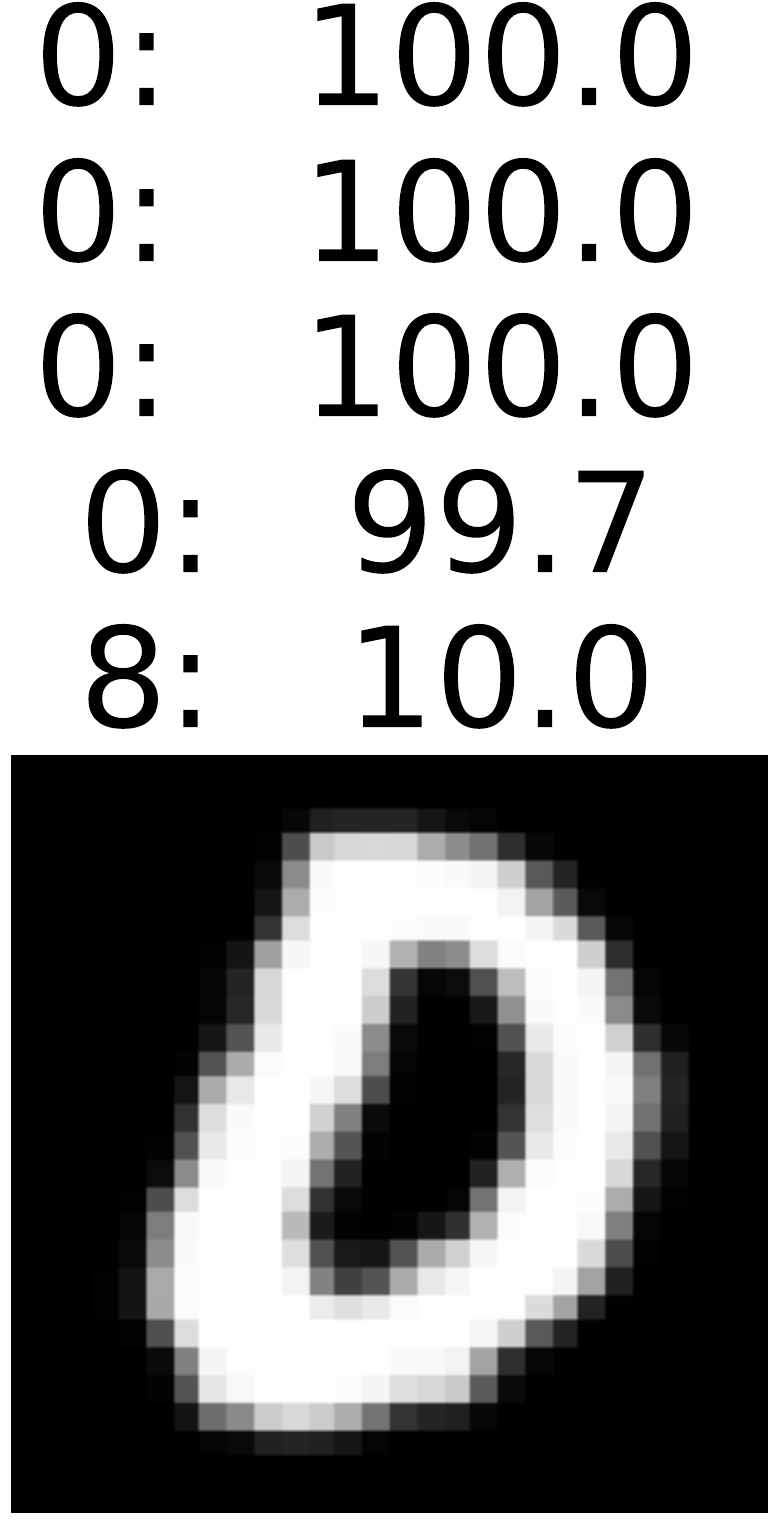} & \includegraphics[width=\plotwidth]{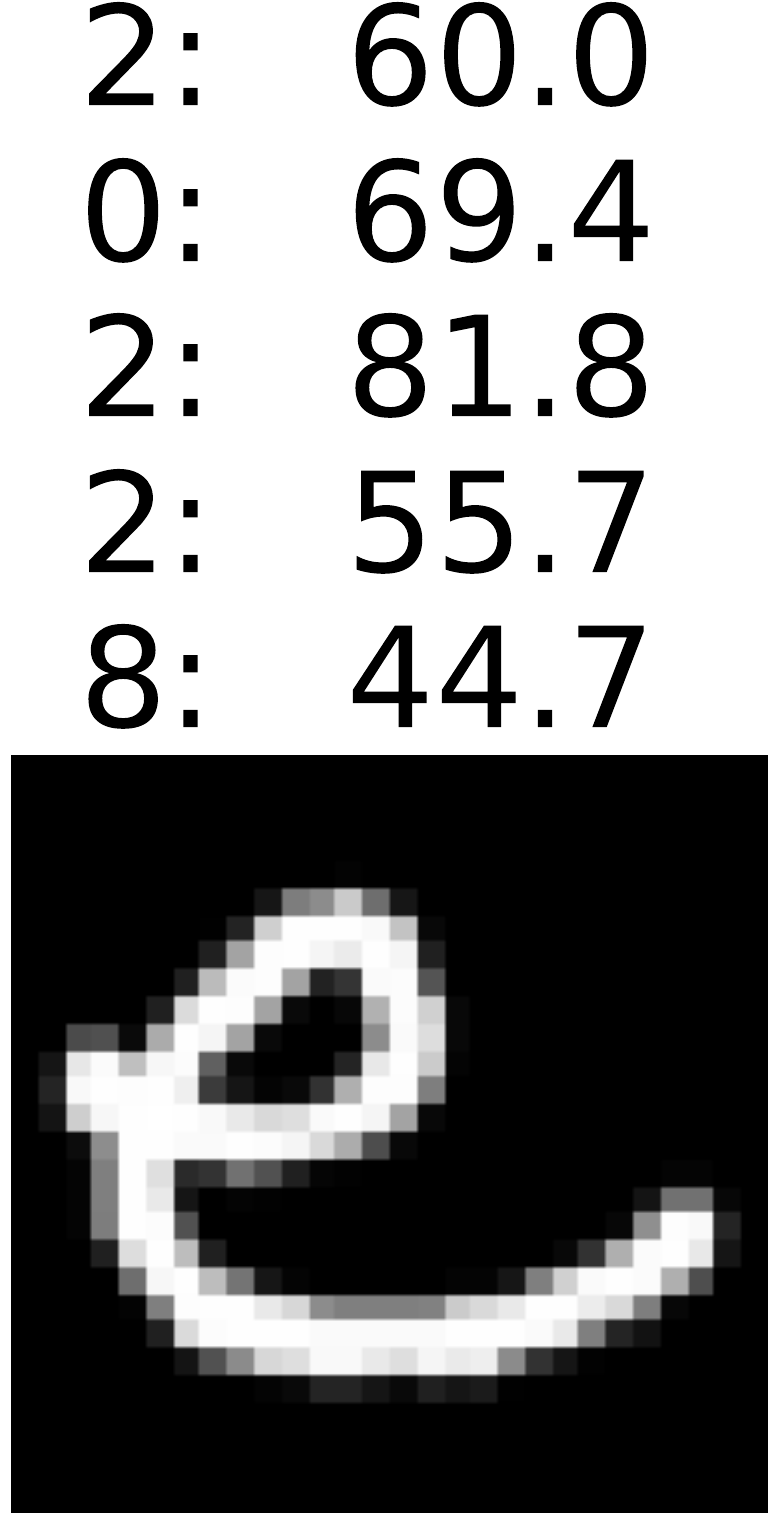} & \includegraphics[width=\plotwidth]{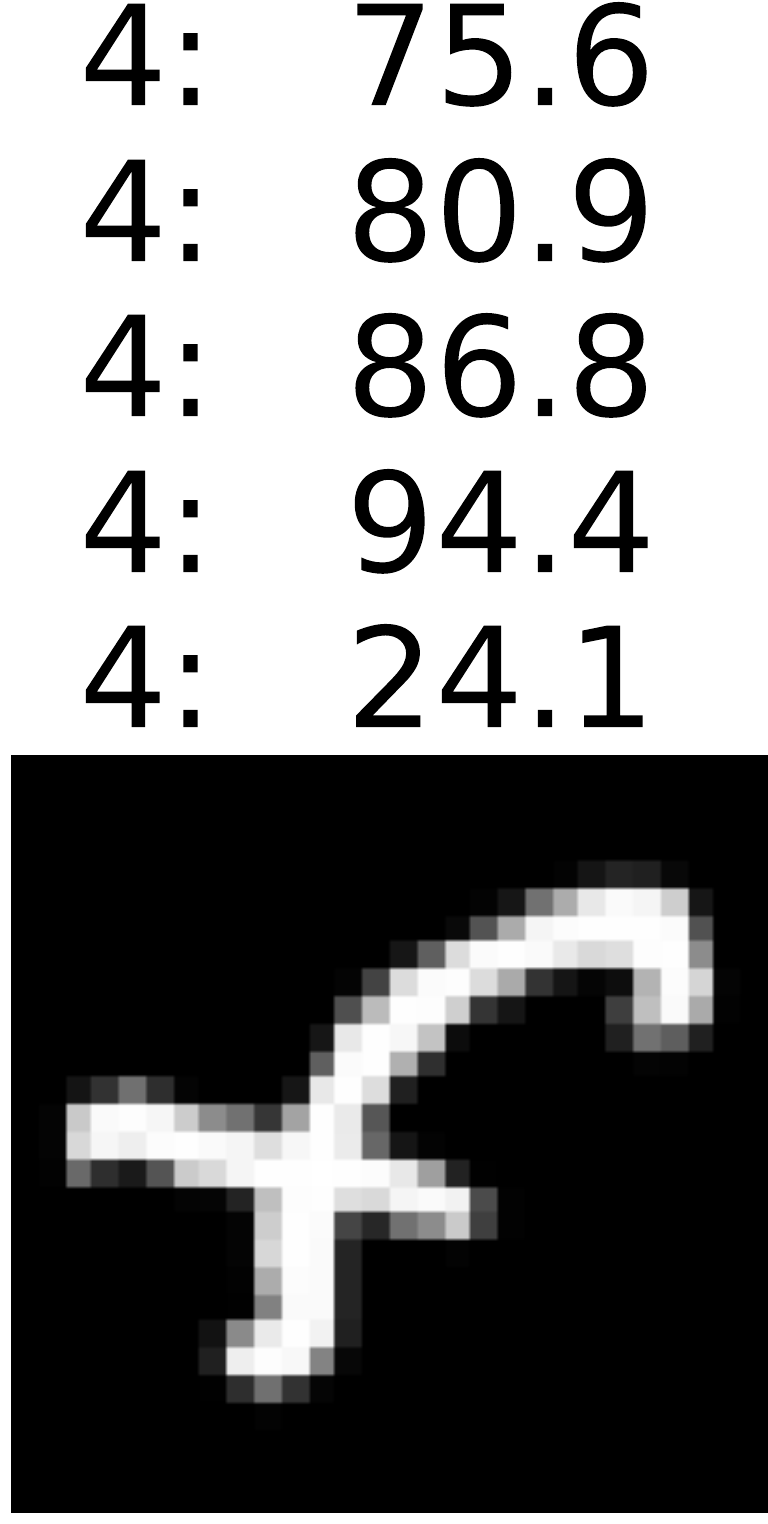} & \includegraphics[width=\plotwidth]{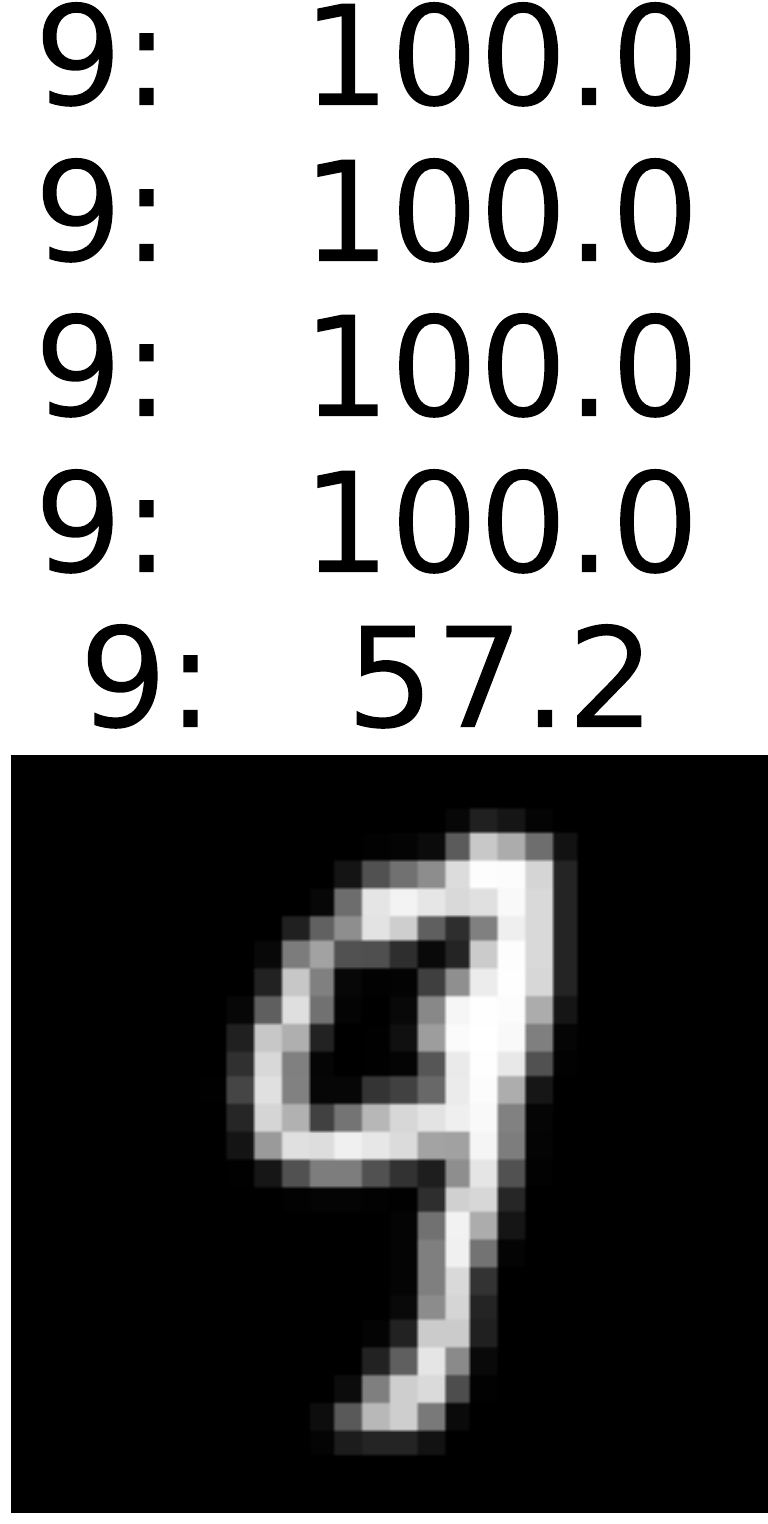} & \includegraphics[width=\plotwidth]{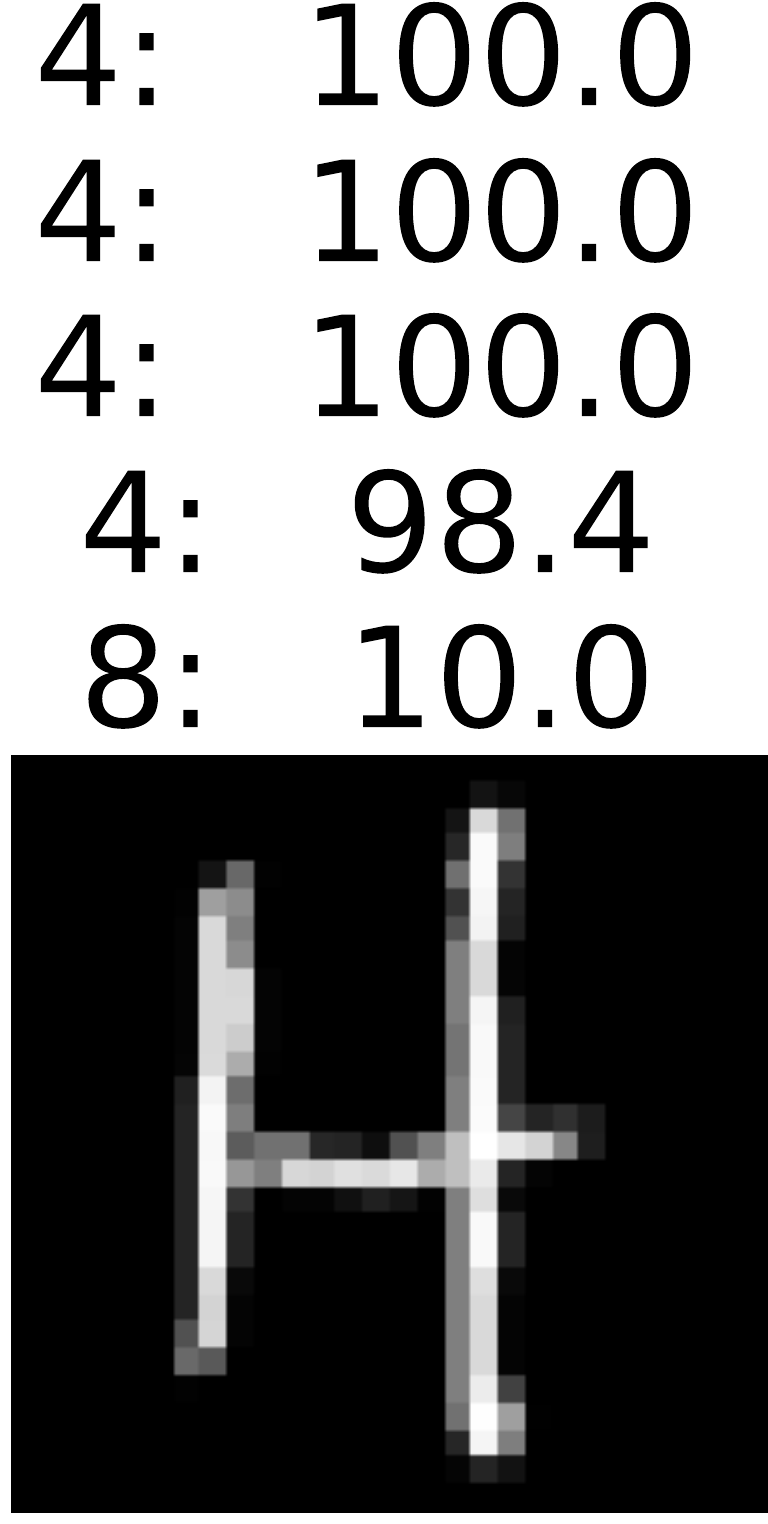} &
\includegraphics[width=\plotwidth]{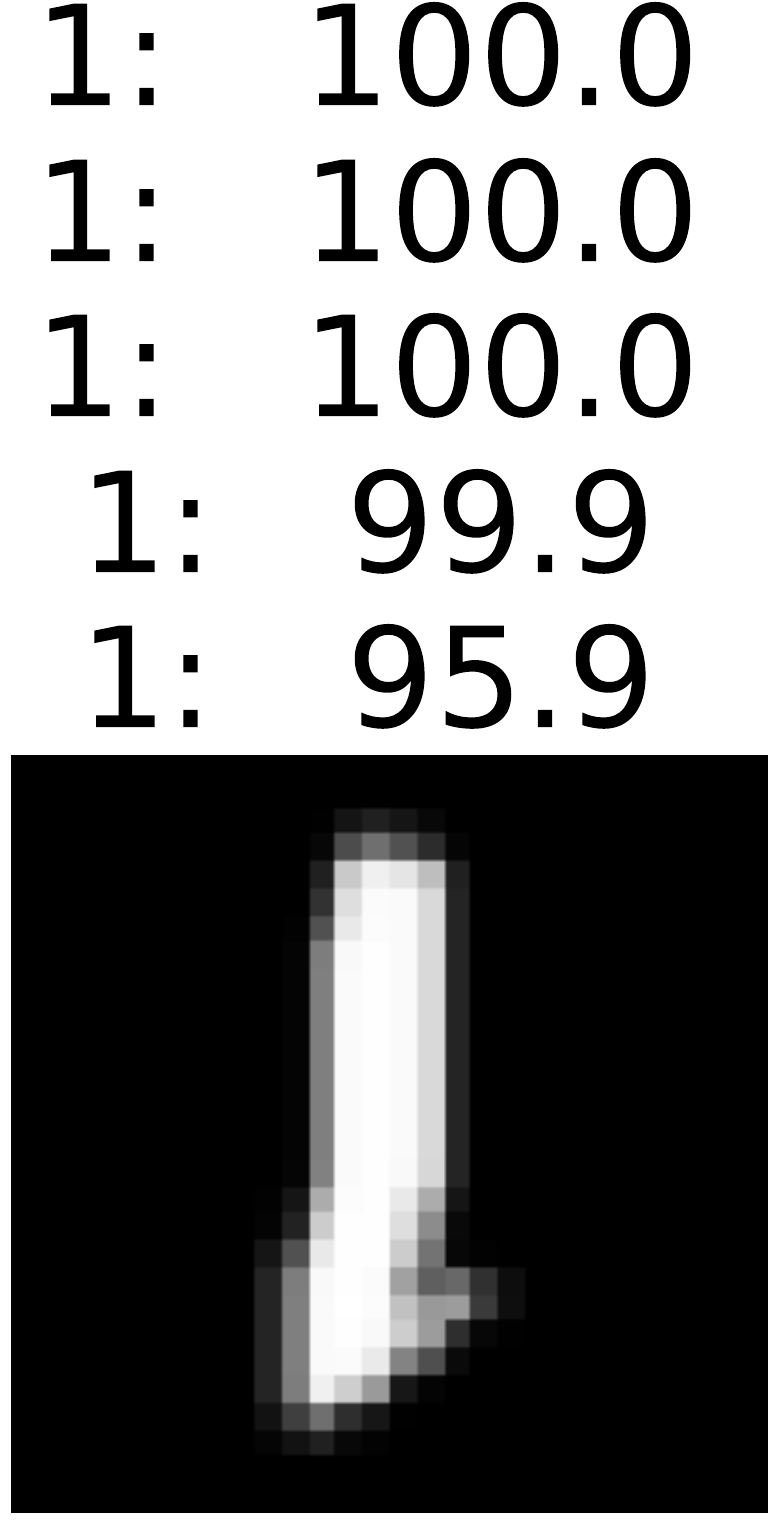} &
\includegraphics[width=\plotwidth]{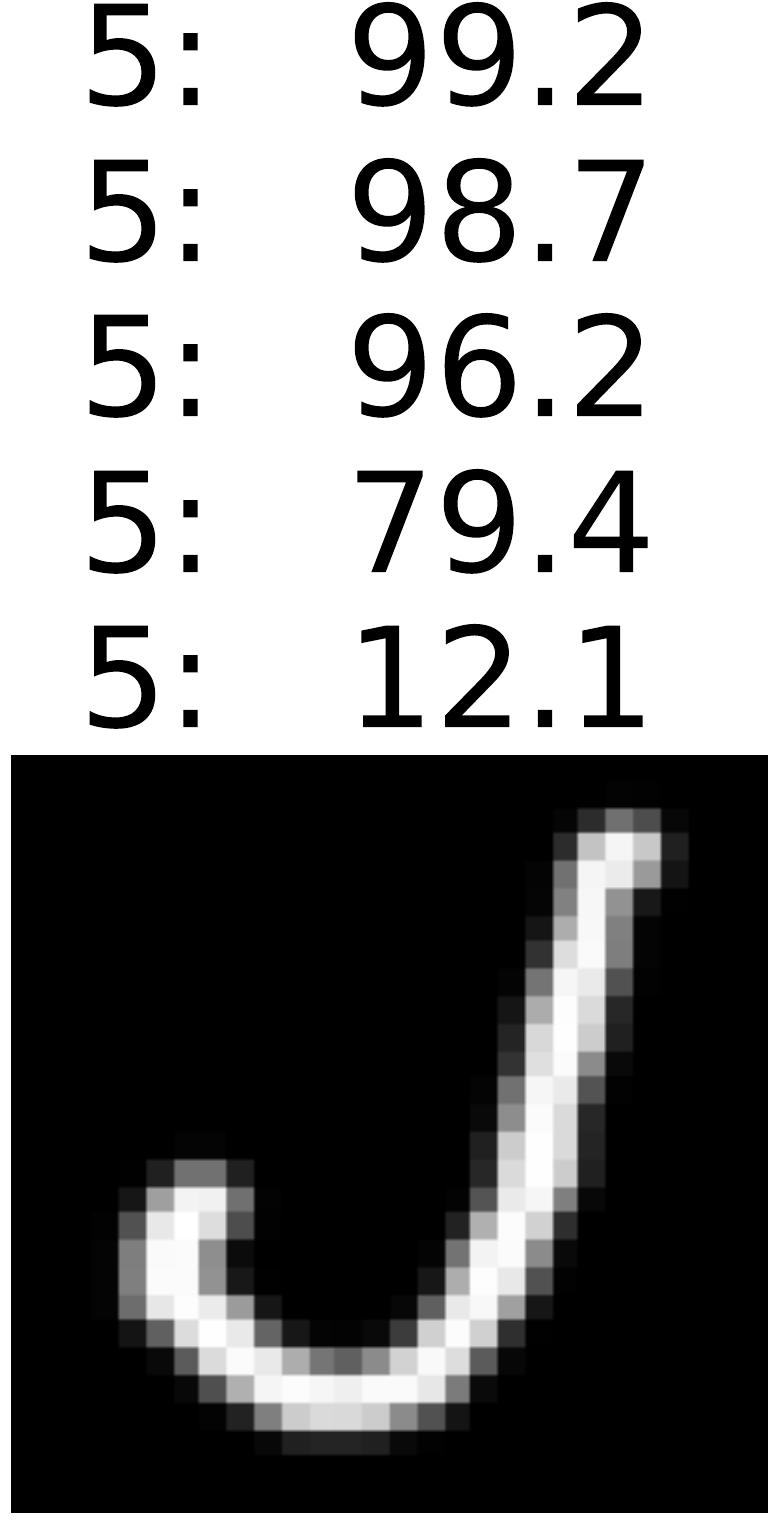} \\

\end{tabular}}

\caption{\label{Fig:Samples} Random samples from 10 letters in the out-distribution dataset EMNIST. The predictions and confidences of all methods trained on MNIST are shown on top. GOOD\textsubscript{100} is the only method which is \textbf{not} overconfident (e.g. ``H'') unless the letter
is indistinguishable from a digit (``I'').}
\end{figure*}

%% file: sections/impact.tex
\section*{Broader Impact}
In order to use machine learning in safety-critical systems it is required that the machine learning system correctly flags its uncertainty. As neural networks have been shown to be overconfident far away from the training data, this work aims at overcoming this issue by not only enforcing low confidence on out-distribution images but even guaranteeing low confidence in a neighborhood around it. As a neural network should not flag that it knows when it does not know, this paper contributes to a safer use of deep learning classifiers.

%% file: sections/acknowledgements.tex
\section*{Acknowledgements}
The authors acknowledge support from the German Federal Ministry of Education and Research (BMBF) through the Tübingen AI Center (FKZ: 01IS18039A) and from the Deutsche Forschungsgemeinschaft (DFG, German Research Foundation) under Germany’s Excellence Strategy (EXC number 2064/1, Project number 390727645), as well as from the DFG TRR 248 (Project number 389792660).
The authors thank the International Max Planck Research School for Intelligent Systems (IMPRS-IS) for supporting Alexander Meinke.

%% file: sections/adversarial.tex
\section{Adversarial attacks on OOD detection} \label{section:adversarial}

It has been demonstrated~\cite{biggio2013evasion,SzeEtAl2014,CarWag2016,Croce_2019_ICCV} that without strong countermeasures, DNNs are very susceptible to adversarial attacks changing the classification result.
The goal of adversarial attacks in our setting is to fool the OOD detection which is based on the confidence in the prediction. Thus the attacker aims at maximizing the confidence in a neighborhood around
a given out-distribution input $x$ so that the adversarially modified image will be wrongly assigned to the in-distribution.
In this paper, we regard as threat model/neighborhood an $l_\infty$-ball of a given radius $\epsilon$, that is $\{z \in [0,1]^d \,|\, \norm{z-x}_\infty \leq \epsilon\}$; note that in our case the disturbed inputs have to be valid images, hence the additional constraint $z \in [0,1]^d$.


For evaluation, we use Auto-PGD \cite{CroHei2020}, which is a state-of-the-art implementation of PGD (projected gradient descent) using adaptive step sizes and random restarts. We use additionally backtracking. Since Auto-PGD has been designed for finding adversarial samples around the in-distribution, we change the objective of Auto-PGD to be the confidence of the classifier. We use Auto-PGD with 500 steps and 5 random restarts which is a quite strong attack. By default, the random initialization is drawn uniformly from the $\epsilon$-ball. However, we found that for MNIST the attack very often got stuck for our GOOD models, because a large random perturbation of size 0.3 would move the sample directly into a region of the input space where the model is completely flat and thus no gradients are available (in this sense adversarial attacks on OOD inputs are more difficult than usual adversarial attacks on the in-distribution). We instead use a modified version of the attack for MNIST which starts within short distance of the original point. 
Thus we use as initialization a random perturbation from  $[-0.01, 0.01]^d$ (note that for our evaluation on CIFAR10, this choice coincides with the default settings).

Nevertheless, for MNIST most out-distribution points lie in regions where the predictions of our GOOD models are flat, i.e. the gradients are exactly zero. Because of this, Auto-PGD is unable to effectively explore the search space around those points. Thus, for MNIST we created an adaptive attack which partially circumvents these issues. First, we use an initialization scheme that mitigates lack of gradients by increasing the contrast as much as the threat model allows. All pixel values $x_i$ that lie above $1-\epsilon$ get set to $x_i=1$ and all values $x_i\leq 1-\epsilon$ get set to $\max \lbrace 0, x_i-\epsilon \rbrace$. In our experience these points are more likely to yield gradients, so we use them as initialization for a 200-step PGD attack with backtracking, adaptive step size selection and momentum of $0.9$. Concretely, we use a step size of $0.1$, and whenever a PGD step does not increase the confidence we backtrack and halve the step size. After every successful gradient step we multiply the step size by $1.1$. Using backtracking and adaptive step size is necessary because otherwise one can easily step into regions where gradient information is no longer available.

Additionally, to further mitigate the problem of gradient-masking at initialization, for each model we use the final best points of all other models and use those as starting points for the same monotone PGD as described before. We use the sample-wise worst-case confidence to compute the final AAUC. Especially CEDA displays much higher apparent robustness if one omits the transfer attacks. Surprisingly, in this respect CEDA behaves very differently from OE, even though they pursue very similar objectives during training.

%% file: sections/ood_conf_reduction.tex
\section{A review of robust OOD detection} \label{section:OOD_conf}

\paragraph{ACET} A method that was proposed in order to achieve adversarially robust low confidence on OOD data is Adversarial Confidence Enhancing Training (ACET)~\citep{HeiAndBit2019} which is based on adversarial training on the out-distribution. However, similar to adversarial training on the in-distribution, typically this does not lead to any guarantees, whereas our goal is to get guarantees on the confidences of worst-case out-distribution inputs.
ACET has 
the following objective:
\begin{align} \label{eq:objective_ACET}
    \frac{1}{N}\sum_{i = 1}^{N} \Lce(x_i^{\text{IN}},y_i^{\text{IN}}) + \frac{\kappa}{M}\sum_{j=1}^{M} \max_{\norm{\hat{x}-x_j^{\text{OUT}}}_\infty \leq \epsilon} \Loe(\hat{x})
    \ .
\end{align}
They use $\Loe = \log\Conf_{\!f}$ with low frequency noise as their training out-distribution.
We found firstly that training an ACET model with 80M as out-distribution yields much better results than the smoothed uniform noise used in \cite{HeiAndBit2019} and 
secondly using the cross-entropy loss with respect to the uniform prediction instead of $\log\Conf_{\!f}$ also leads to improvements.
For training ACET models, we employ a standard PGD attack with 40 steps of size $\frac{2\epsilon}{41}$ with initialization at the target input for maximizing the loss around $x_j^{\text{OUT}}$. As usual for a $l_\infty$-attack, we use the sign of the gradient as direction and project onto the intersection of the image
domain $[0,1]^d$ and the $l_\infty$-ball of radius $\epsilon$ around the target. Finally, the attack returns
the image with the highest confidence found during the iterations. For the attack at training time we use no backtracking or adaptive stepsizes. 
ACET does not provide any guaranteed confidence bounds.

\paragraph{CCU} Certified Certain Uncertainty (CCU)~\citep{meinke2020towards} gives low confidence guarantees around certain OOD data that is far away from the training dataset in a specific metric. Those bounds do hold on such far-away datasets, but do not generalize to inputs relatively close to the in distribution, like for example CIFAR-10 vs. CIFAR-100. 
Moreover, even in the regime where CCU yields meaningful guarantees, they are given in terms of a data-dependent Mahalanobis distance rather than the $l_\infty$-distance.
However, due to norm equivalences, one can still extract $l_\infty$-guarantees from CCU and we evaluated the CCU guarantees as follows.
We use the corollary 3.1 from \cite{meinke2020towards} which states that for a CCU model that is written as 
\begin{equation}
    p(y|x) = \frac{p(y|x,i)p(x|i)+\frac{1}{K} p(x|o) }{p(x|i)+p(x|o)}
\end{equation}
with $p(y|x,i)$ being the softmax output of a neural network and $p(x|i)$ and $p(x|o)$ Gaussian mixture models for in-and out-distribution, one can bound the confidence in a certain neighborhood around any point $x \in \R^d$ via
\begin{equation}\label{Eq:CCU_guarantee}
\underset{d_M(\hat{x},x)\leq R}{\max}p(y|x) \, \leq \,\frac{1}{K} \frac{1 + K\,b(x, R) }{1+ b(x, R)} .
\end{equation}
Here $b:\R^d \times \R_+ \rightarrow \R_+$ is a positive function that increases monotonically in the radius $R$ and that depends on the parameters of the Gaussian mixture models (details in \cite{meinke2020towards}). The metric $d_M: \R^d \times \R^d \rightarrow \R_+$ that they used for their CCU model is given as 
\begin{equation}
    d_M(x,y)=\norm{C^{-\frac{1}{2}}(x-y)},
\end{equation}
where $C$ is a regularized version of the covariance matrix, calculated on the augmented in-distribution data. Note that this Mahalanobis metric is strongly equivalent to the metric induced by the $l_2$-norm and consequently to the metric induced by the $l_\infty$-norm. By computing the equivalence constants between these metrics we can extract the $l_\infty$-guarantees that are implicit in the CCU model. Geometrically speaking, we compute the size a an ellipsoid (its shape determined by the eigenvalues of $C$) that is large enough to fit a cube inside it with a radius given by our threat model $r=0.3$ or $r=0.01$, respectively. Via norm equivalences one has
\begin{equation}
    d_M(x,y) \leq \sqrt{\lambda_1} d_2(x,y) \leq \sqrt{d\lambda_1} d_\infty(x,y) \leq \sqrt{d\lambda_1} r,
\end{equation}
where $\lambda_1$ is the largest eigenvalue of $C$. This means that the confidence upper bounds from \eqref{Eq:CCU_guarantee} on a Mahalanobis-ball of radius $R=(d\lambda) ^{\frac{1}{2}} r$ automatically apply to an $l_\infty$-ball of radius $r$. However, the covariance matrix $C$ is highly ill-conditioned, which means that $\lambda_1$ is fairly high. On top of that, in high dimensions $\sqrt{d}$ is big as well so that in practice the required radius $R$ becomes too large for CCU to certify meaningful guarantees. Even on uniform noise, the upper bounds were larger than the highest confidence on the in-distribution test set, with the consequence that there are no lower-bounds on the AAUC. However, we want to stress that at least for uniform noise the lack of guarantees of CCU is due to the incompatability of the threat models used in our paper and \cite{meinke2020towards}.

Another type of guarantee that certifies a detection rate for OOD samples by applying probably approximately correct (PAC) learning considerations has been proposed in~\cite{liu18e}. Their problem setting and nature of guarantees are not directly comparable to ours, since their guarantees handle behaviour on whole distributions while our guarantees are given for individual datapoints.

%% file: sections/auroc_definition.tex
\section{AUC and Conservative AUC} \label{section:auroc}
As a measure for the separation of in- vs. out-distribution data we use the Area Under the Receiver Operating Characteristic curve (AUROC or AUC) using the confidence of the classifier as the feature. The AUC is 
equal to the empirical probability of a random in-sample to be assigned a higher confidence than a random out-sample, plus one half times the probability of the confidences being equal. Thus, the standard way (as e.g. implemented in scikit-learn \cite{scikit-learn}) to calculate the AUC from given confidence values on sets of in- and out-distribution samples $S_{in}$ and $S_{out}$ is
\begin{align}
    \begin{split}
        \AUROC(f, S_{in}, S_{out}) = \frac{1}{|S_{in}| |S_{out}|} \biggl( 
        &\left| \left\{ x_{in} \in S_{in}, x_{out} \in S_{out} \mid \Conf_{\!f}(x_{in}) > \Conf_{\!f}(x_{out}) \right\} \right| \\
        + \frac{1}{2} &\left| \left\{ x_{in} \in S_{in}, x_{out} \in S_{out} \mid \Conf_{\!f}(x_{in}) = \Conf_{\!f}(x_{out}) \right\} \right| \biggr) \ ,
    \end{split}
\end{align}
where for a set $S$, $|S|$ indicates the number of its elements.
The half-weighted equality term gives this definition certain symmetry properties. 
However, it assigns a positive score to some completely uninformed functions~$f$.
For example, a constant uniform classifier with $p_k(x) = \frac{1}{K}$ receives an AUC value of 50\%.
Similarly, a classifier that assigns 100\% confidence to most in-distribution inputs would have positive AUC and even GAUC statistics, even if it fails to have confidence below 100\% on any OOD inputs.
In order to regard only example pairs where the distributions are positively distinguished, we define the \textbf{Conservative AUC} ($\CAUROC$) by dropping the equality term:
\begin{align} \label{eq:cauroc}
    \CAUROC(f, S_{in}, S_{out}) := \frac{1}{|S_{in}| |S_{out}|}
        &\left| \left\{x_{in} \in S_{in}, x_{out} \in S_{out} \mid \Conf_{\!f}(x_{in}) > \Conf_{\!f}(x_{out}) \right\} \right| \ .
\end{align}
While in general $\CAUROC(f, S_{in}, S_{out}) \leq  \AUROC(f, S_{in}, S_{out})$, the confidences of all models presented in the paper are differentiated enough so that for all shown numbers actually $\CAUROC = \AUROC$.
However, we have experienced that one can have models where the confidences (uniform or one-hot predictions) cannot be distinguished due to limited numerical precision. In these cases the normal AUC definition would indicate a certain discrimination where it is actually impossible to discriminate the confidences.

%% file: sections/experimental_details.tex
\section{Experimental details}\label{section:experimental_details}

The layer compositions of the architectures used for all GOOD and baseline models are laid out in Table \ref{table:architectures}. No normalization of inputs or activations is used. Weight decay ($l_2$) is set to $0.05$ for MNIST and $0.005$ for SVHN and CIFAR-10.
For all runs, we use a batch size of 128 samples from both the in- and the out-distribution (where applicable). 
At \url{https://gitlab.com/Bitterwolf/GOOD} you can find the exact implementation.

\input{tables/table_arch.tex}
For the MNIST experiments, we use as optimizer SGD with 0.9 Nesterov momentum, with an initial learning rate of $\frac{0.005}{128}$ that is divided by 5 after 50, 100, 200, 300 and 350 epochs, with a total number of 420 training epochs.
For the GOOD, CEDA and OE runs, the first two epochs only use in-distribution $\Lce$; over the next 100 epochs, the value of $\kappa$ is ramped up linearly from zero to its final value of $0.3$ for GOOD/OE and $1.0$ for CEDA, where it stays for the remaining 318 epochs. The $\epsilon$ value in the $\LCUB$ loss for GOOD is also increased linearly, starting at epoch 10 and reaching its final value of $0.3$ on epoch 130. CCU is trained using the publicly available code from \cite{meinke2020towards}, where we modify the architecture, learning rate schedule and data augmentation to be the same as OE. The initial learning rate for the Gaussian mixture models is $1e-5/\mathrm{batchsize}$ and gets dropped at the same epochs as the neural network learning rate. Our more aggressive data augmentation implies that our underlying Mahalanobis metric is not the same as they used in \cite{meinke2020towards}.
The ACET model for MNIST is warmed up with two epochs on the in-distribution only, then four with $\kappa = 1.0$ and $\epsilon = 0$, and the full ACET loss with $\kappa = 1.0$ and $\epsilon = 0.3$ for the remaining epochs.
The reason why we chose a smaller $\kappa$ of $0.3$ for the MNIST GOOD runs is that considering the large $\epsilon$ for which guarantees are enforced, training with higher $\kappa$ values makes training unstable without improving any validation results.

For the SVHN and CIFAR-10 baseline models, we used the ADAM optimizer \cite{KinEtAl2014} with initial learning rate $\frac{0.01}{128}$ for SVHN and $\frac{0.1}{128}$ for CIFAR-10 that was divided by 5 after 30 and 100 epochs, with a total number of 420 training epochs.
For OE, $\kappa$ is increased linearly from zero to one between epochs 60 and 360. The same holds for CCU which again uses the same hyperparameters as OE.
Again, ACET is warmed up with two in-distribution-only and four OE epochs. Then it is trained with $\kappa = 1.0$ and $\epsilon = 0.03/0.01$ (SVHN/CIFAR-10), with a shorter training time of 100 epochs (the same number as used in~\cite{HeiAndBit2019}).\\
In line with the experiences reported in \cite{gowal2018effectiveness} and \cite{zhang2020towards}, for GOOD training on SVHN and CIFAR-10 longer training schedules with slower ramping up of the $\LCUB$ loss are necessary, as adding the out-distribution loss defined in Equation \eqref{eq:cub_loss} to the training objective at once will overwhelm the in-distribution cross-entropy loss and cause the model to collapse to uniform predictions for all inputs, without recovery.
In order to reduce warm-up time, we use a pre-trained CEDA model for initialization and train for 900 epochs.
The learning rate is 10\textsuperscript{-4} in the beginning and is divided by 5 after epochs 450, 750 and 850.
Due to the pre-training, we begin training with a small $\kappa$ and already start with non-zero $\epsilon$ after epoch 4. Then, $\epsilon$ is increased linearly to its final value of $0.03$ for SVHN and $0.01$ for CIFAR-10, which is reached at epoch 204. Simultaneously, $\kappa$ is increased linearly with a virtual starting point at epoch {-2} to its final value of $1.0$ at epoch 298.

Due to the tendency of IBP based training towards instabilities, the selection of hyper-parameters was based on finding settings where training is reliably stable while guaranteed bounds over meaningful $\epsilon$-radii are possible. 

For the accuracy, AUC and GAUC evaluations in Table \ref{table:all_aucs} the test splits of each (non-noise) dataset were used, with the following numbers of samples: 10,000 for MNIST, FashionMNIST, CIFAR-10, CIFAR-100 and Uniform Noise; 20,800 for EMNIST Letters; 26,032 for SVHN; 300 for LSUN Classroom. Due to the computational cost of the employed attacks, the AAUC values are based on subsets of 1000 samples for each dataset.

All experiments were run on Nvidia Tesla P100 and V100 GPUs, with GPU memory requirement below 16GB.

%% file: tables/table_arch.tex
\begin{table}[hb]
\caption{Model architectures used for MNIST (L), SVHN (XL) and CIFAR-10 (XL) experiments. Each convolutional and non-final affine layer is followed by a ReLU activation. All convolutions use a kernel size of 3, a padding of 1, and stride of 1, except for the third convolution which has stride=2.}
\label{table:architectures}
\vskip 0.15in
\begin{center}
\begin{small}
\begin{sc}
\begin{tabular}{l|l}
\toprule
L & XL     \\
\midrule
Conv2d(64)
& Conv2d(128)\\
Conv2d(64)
& Conv2d(128)\\
Conv2d(128)\textsubscript{s=2}
& Conv2d(256)\textsubscript{s=2}\\
Conv2d(128)
& Conv2d(256)\\
Conv2d(128)
& Conv2d(256)\\
Linear(512)
& Linear(512)\\
Linear(10)
& Linear(512)\\
& Linear(10) \\
\bottomrule
\end{tabular}
\end{sc}
\end{small}
\end{center}
\vskip -0.1in
\end{table}

%% file: sections/quantile_depiction.tex
\section{Depiction of GOOD Quantile-loss}
In Quantile-GOOD training, the out-distribution part of each batch is split up into  ``harder'' and ``easier'' parts, since trying to enforce low confidence guarantees on out-distribution inputs that are very close to the in-distribution leads to low confidences in general, even on the in-distribution.
In Table \ref{table:quantile_images}, we show example batches of GOOD\textsubscript{60} models with MNIST, SVHN and CIFAR-10 as in-distribution near the end of training (from epochs 410, 890 and 890, respectively).
Even though the actual CIFAR images were filtered out, some images containing objects from CIFAR-classes are still present. For the CIFAR-10 model, such samples (among others) get sorted above the quantile. For MNIST, lower brightness images appear to be more difficult, while for SVHN images with fewer objects seem to be comparably hardest to distinguish from the house numbers of the in-distribution.
\input{tables/table_quantile_images}

\section{Confidences on EMNIST}
Figure \ref{Fig:Samples_cont} shows samples of the letters ``k'' through ``z'' together with the predictions and confidences of the GOOD\textsubscript{100} MNIST model and four baseline models, complementing Figure \ref{Fig:Samples}. 
We see that GOOD\textsubscript{100} produces low confidences for most letters when they show no digit-specific features. Interestingly it even rejects some letters that could easily be mistaken for digits by humans (``o'').
The mean confidence values of the same selection of MNIST models for each letter of the alphabet for EMNIST are plotted in Figure \ref{Fig:EMNIST_conf}.
We observe that the mean confidence often aligns with the intuitive likeness of a letter with some digit: GOOD\textsubscript{100} has the highest mean confidence on the letter inputs ``i'' and ``l'', which in many cases do look like the digit ``1''. Again, the confidence of GOOD\textsubscript{100} on the letter ``o'', which even humans often cannot distinguish from a digit ``0'', is generally low. On the other hand, ``y'' receives a surprisingly high confidence, compared to other letters, so we conclude that GOOD\textsubscript{100} uses different features than humans in order to achieve its impressive performance on EMNIST.

\renewcommand{\plotwidth}{.08\textwidth}
\begin{figure*}[ht]
\centering
{\footnotesize
\setlength{\tabcolsep}{1.6mm}
\hspace{-.6cm}
\begin{tabular}{lllllllllll}

\includegraphics[width=\plotwidth]{images/samples/blank_2.pdf} & \includegraphics[width=\plotwidth]{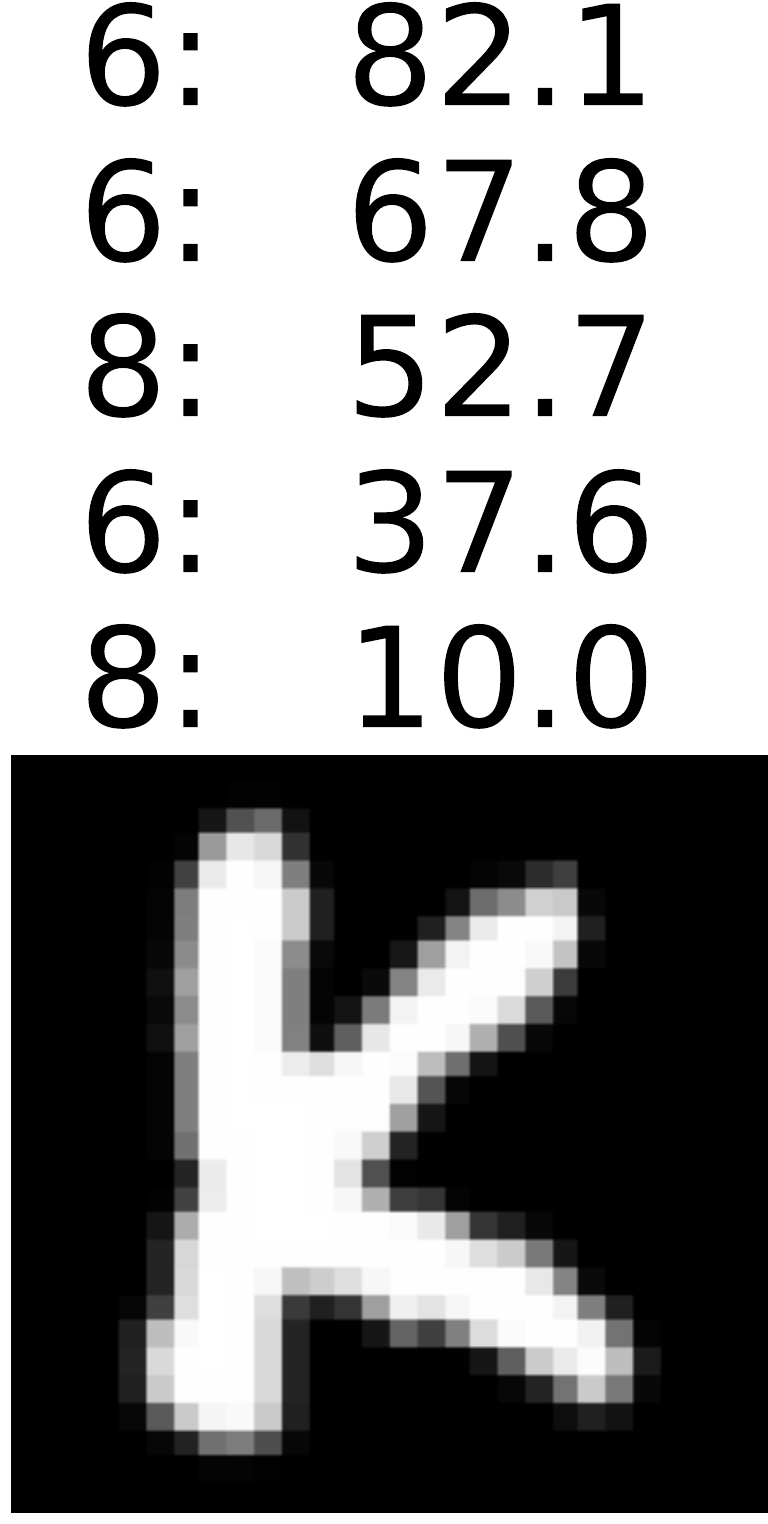} & \includegraphics[width=\plotwidth]{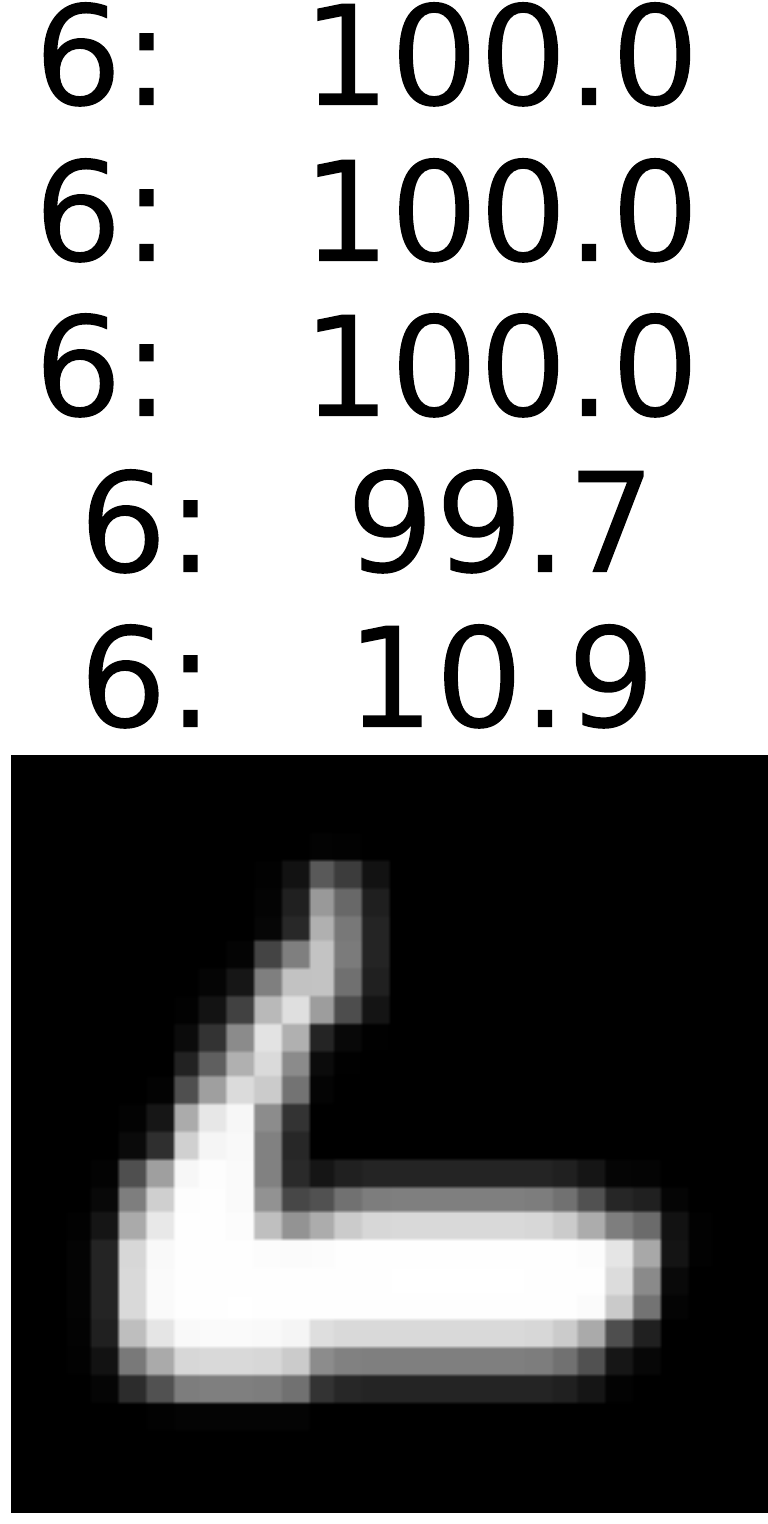} & 
\includegraphics[width=\plotwidth]{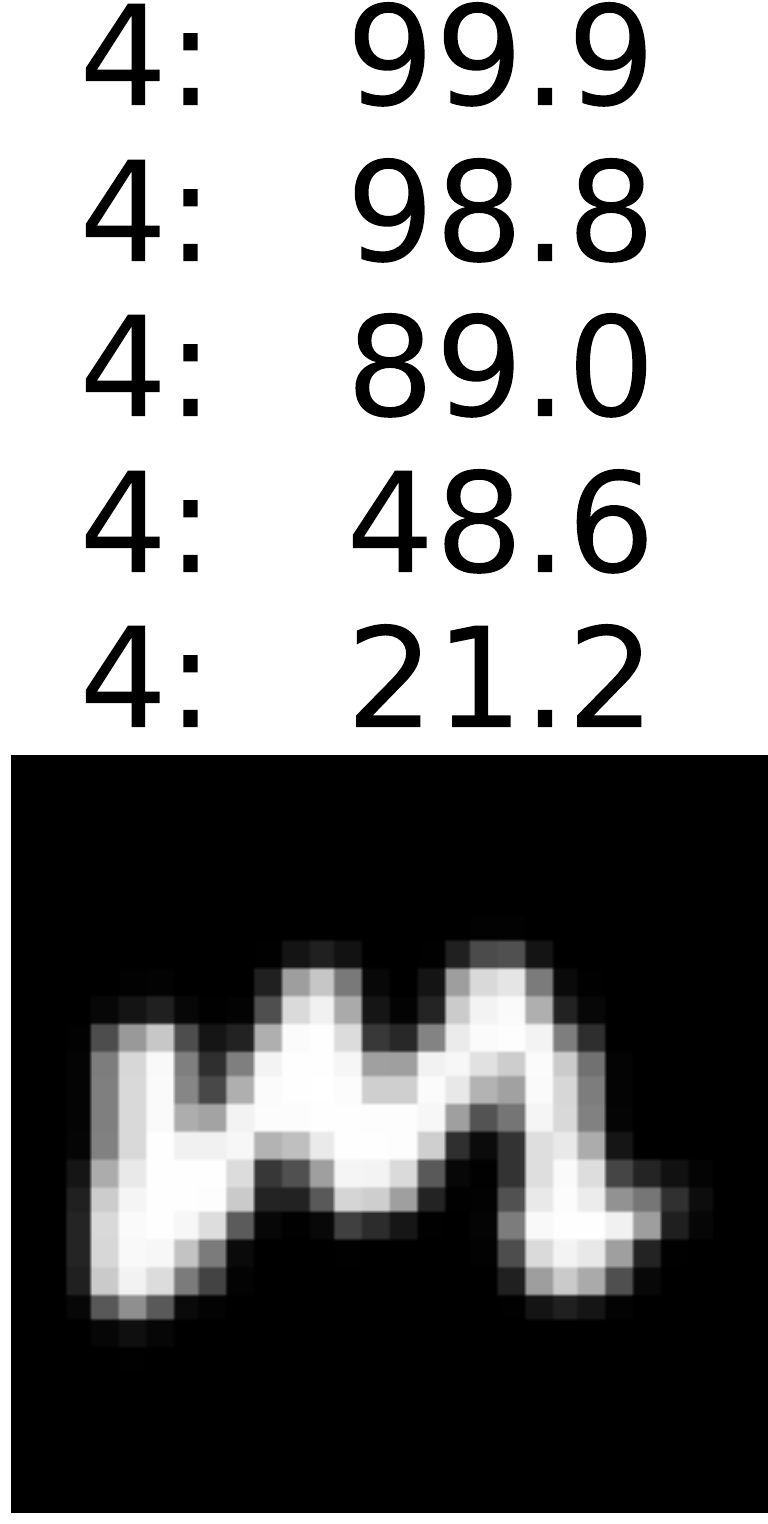} & \includegraphics[width=\plotwidth]{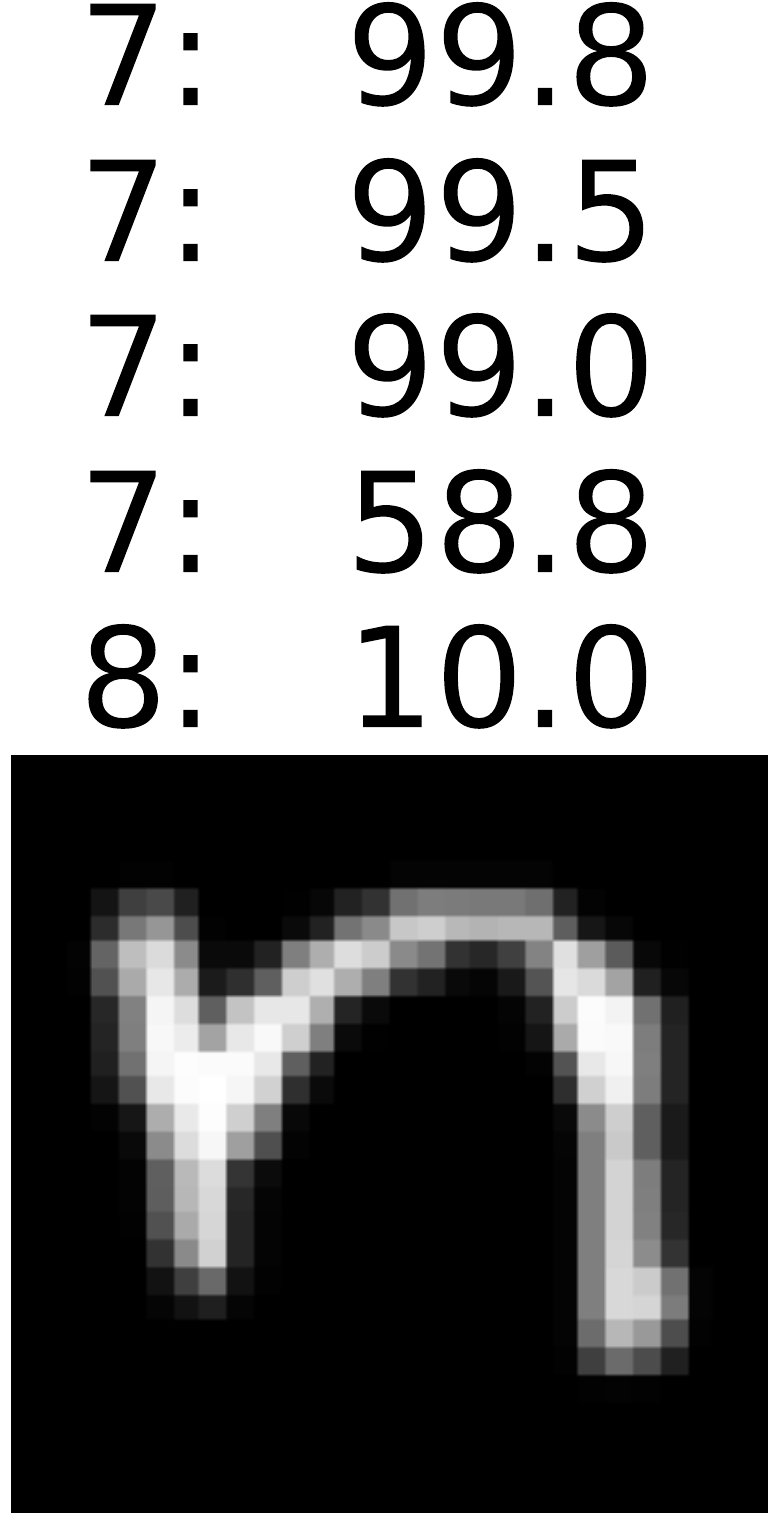} & \includegraphics[width=\plotwidth]{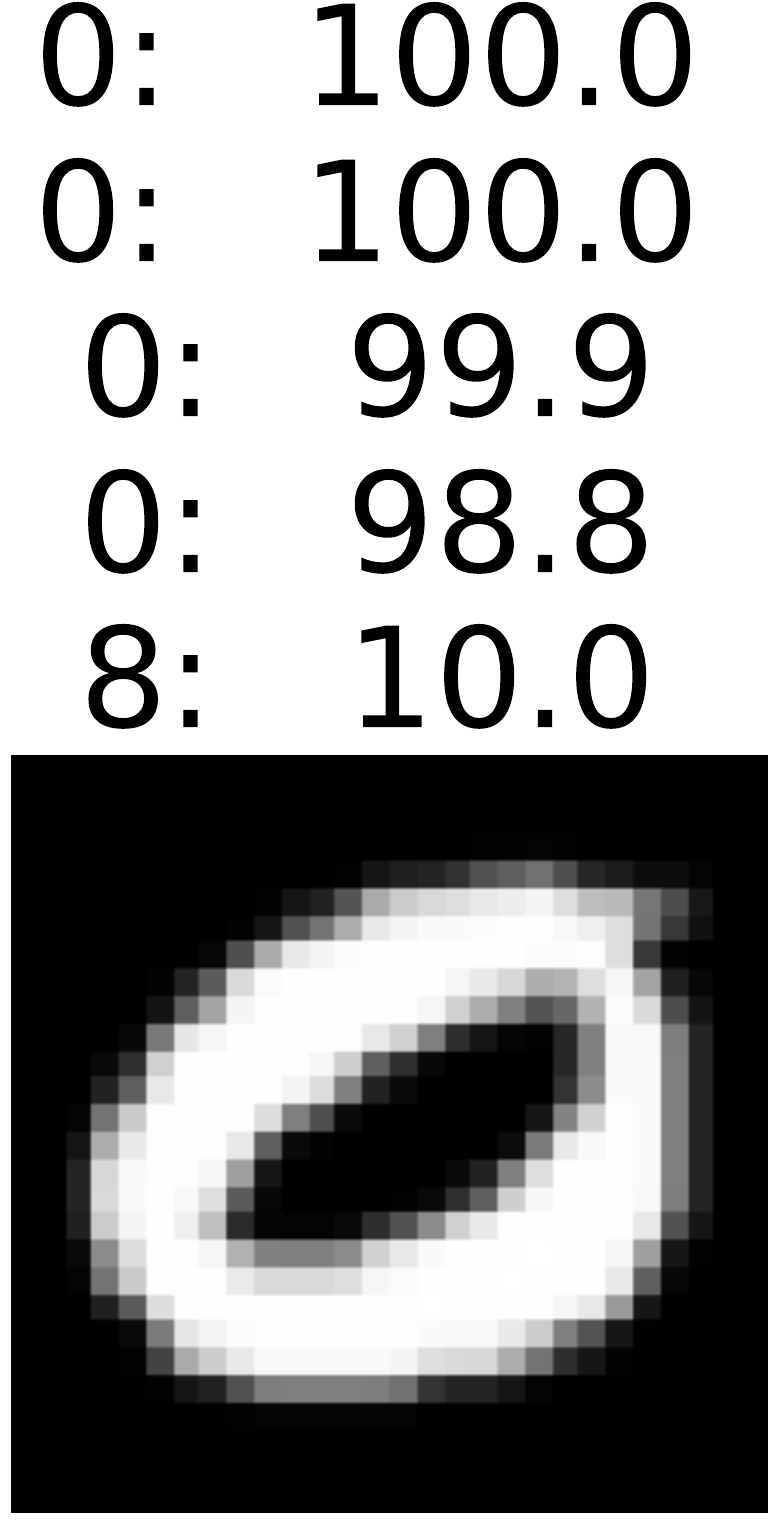} & \includegraphics[width=\plotwidth]{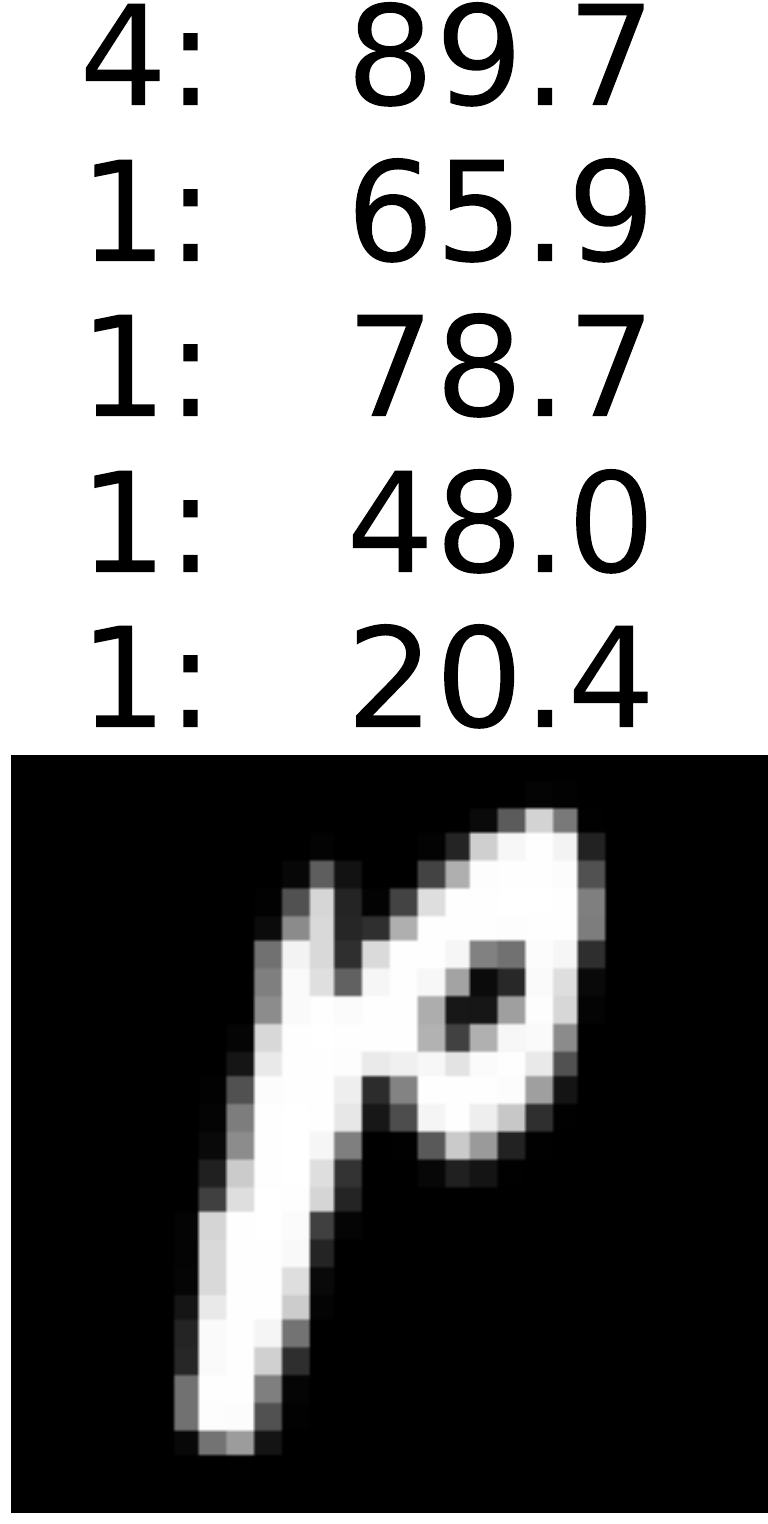} & \includegraphics[width=\plotwidth]{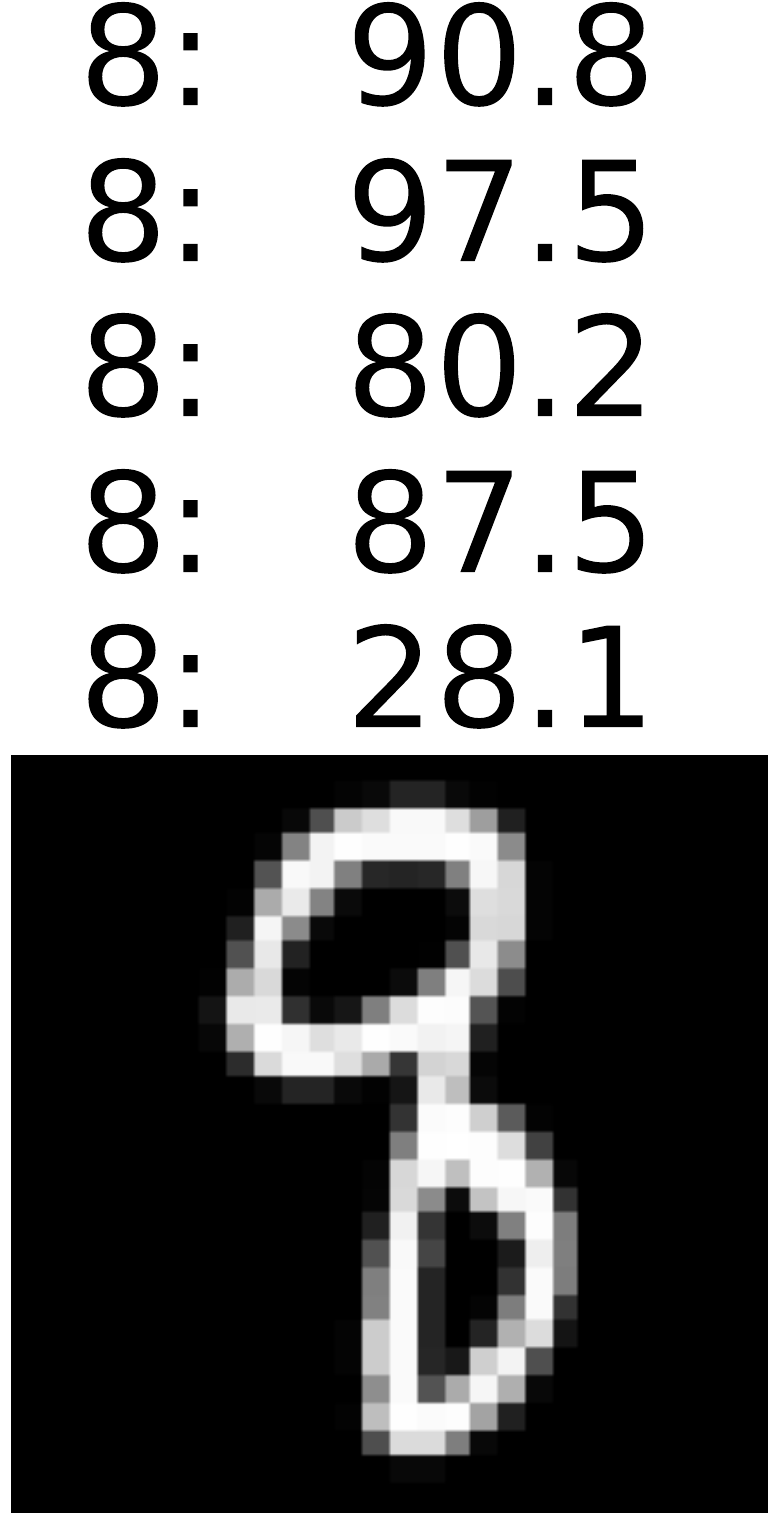} & \includegraphics[width=\plotwidth]{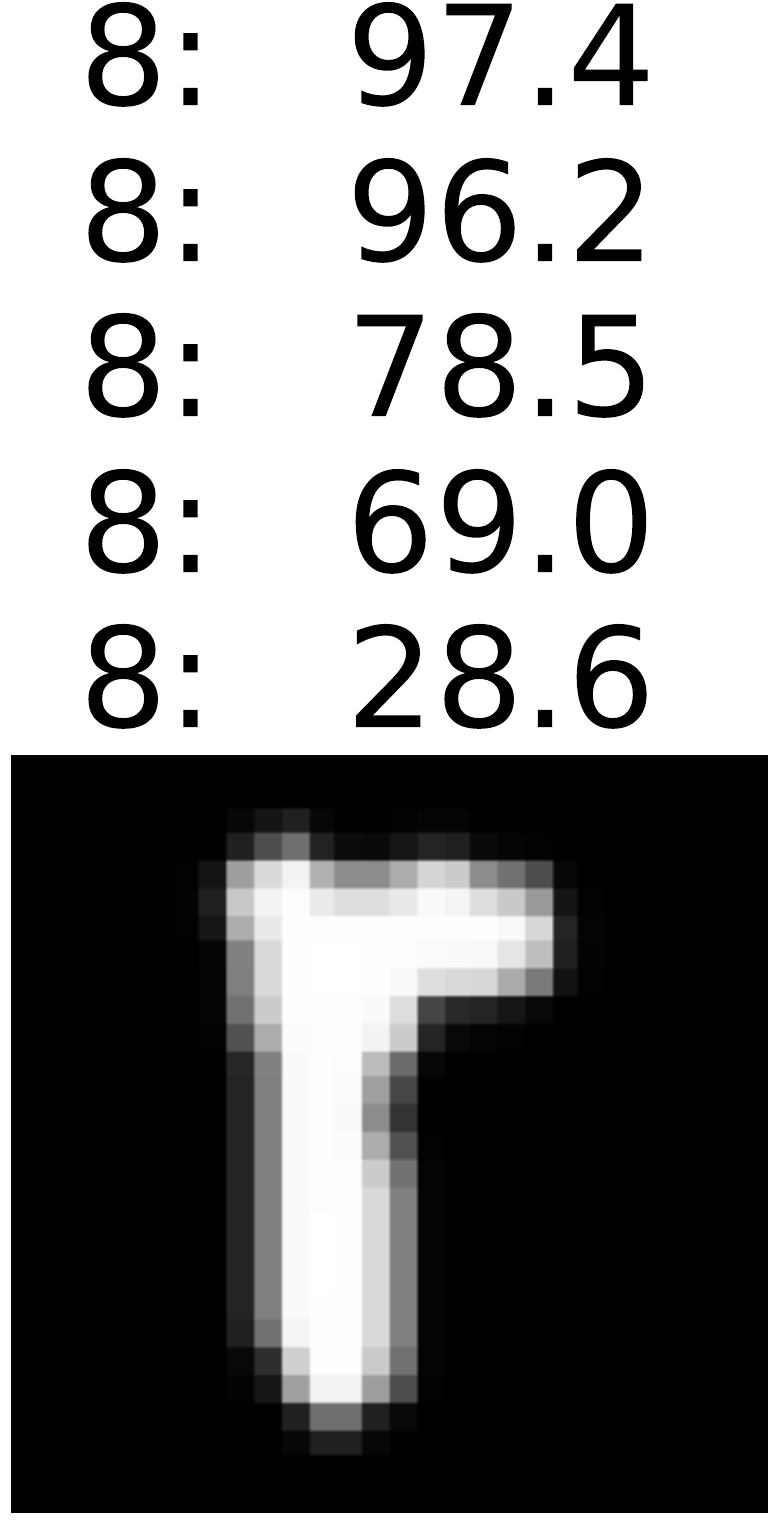} \\[3mm]
\includegraphics[width=\plotwidth]{images/samples/blank_2.pdf}  & \includegraphics[width=\plotwidth]{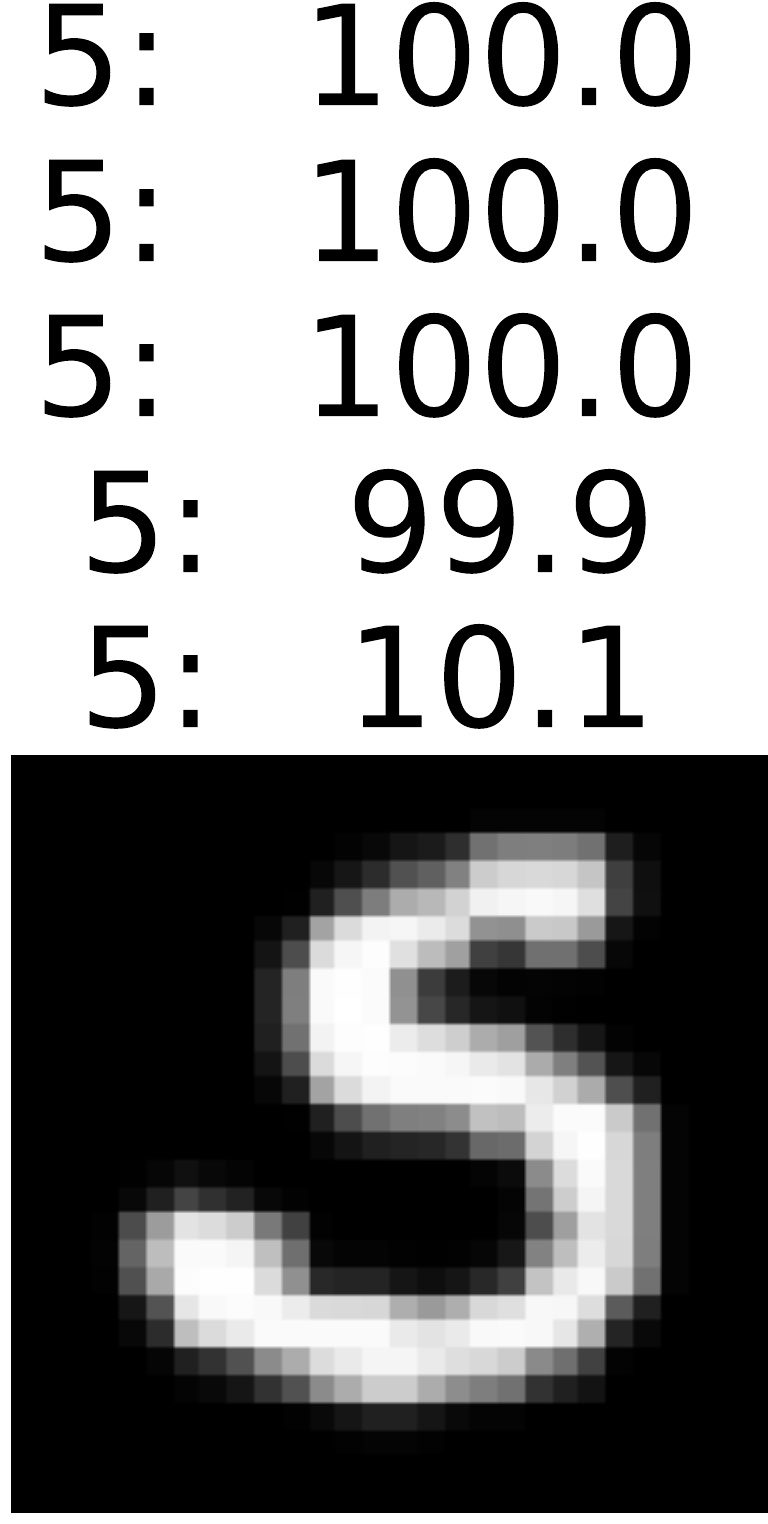} & 
\includegraphics[width=\plotwidth]{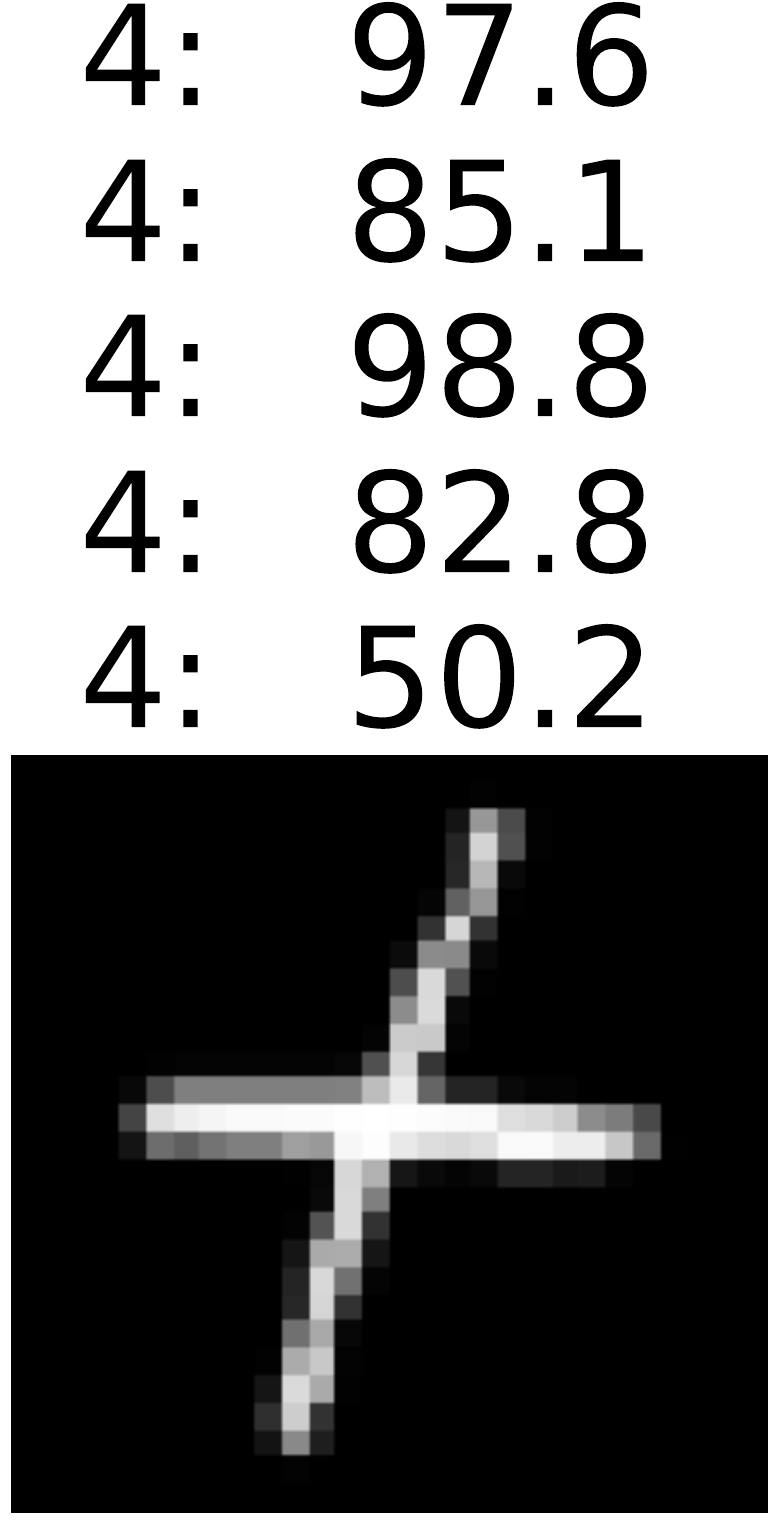} & \includegraphics[width=\plotwidth]{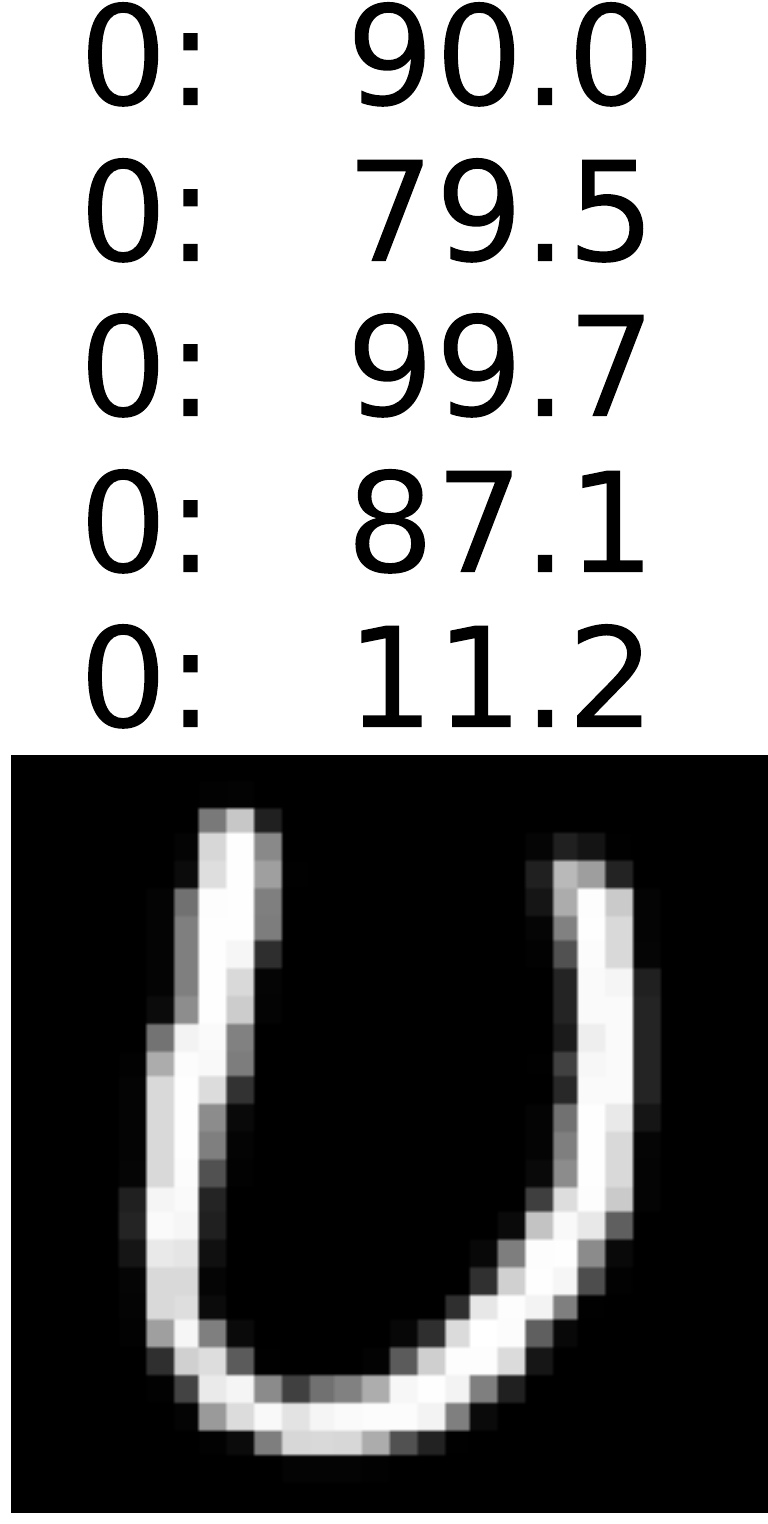} & \includegraphics[width=\plotwidth]{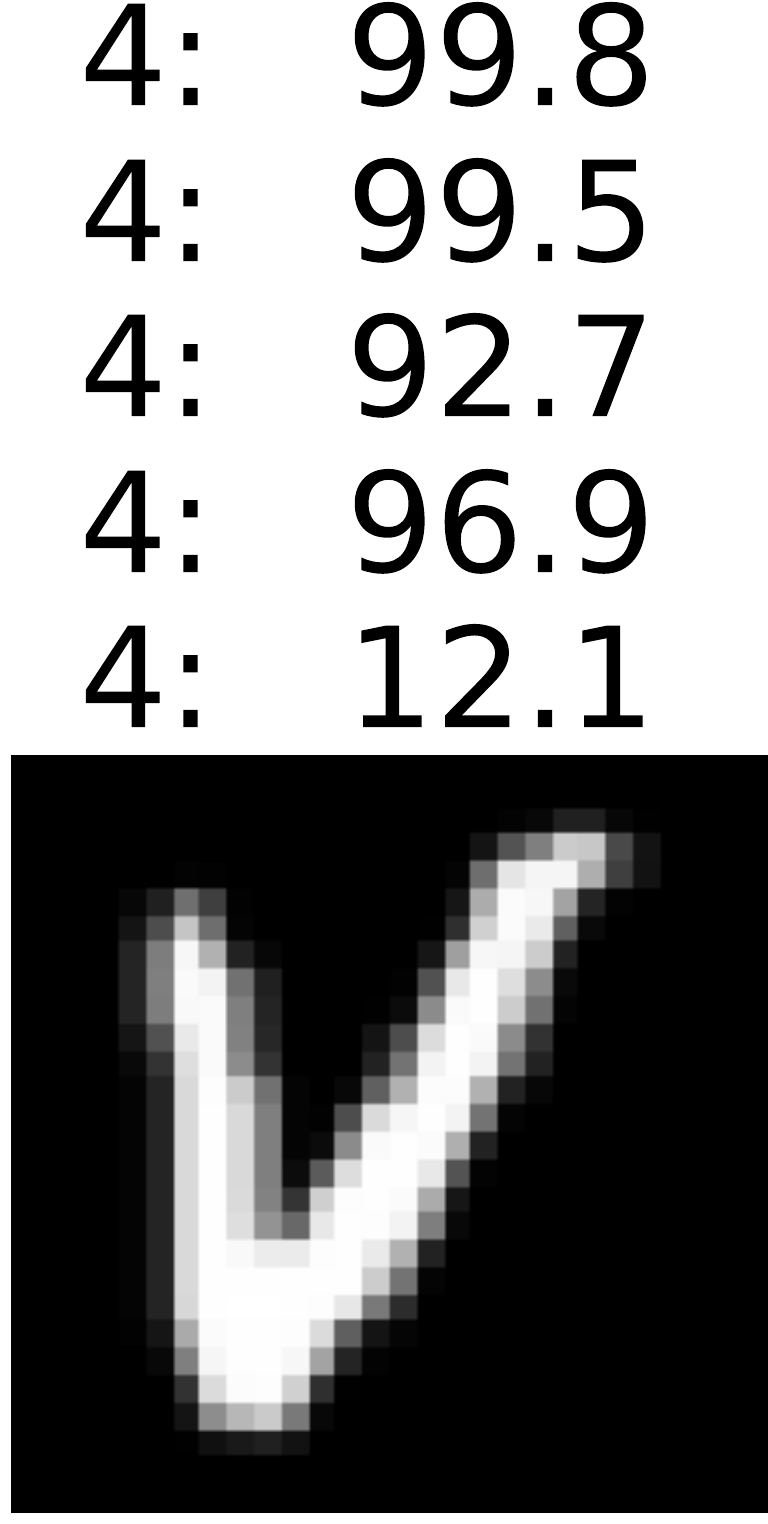} & \includegraphics[width=\plotwidth]{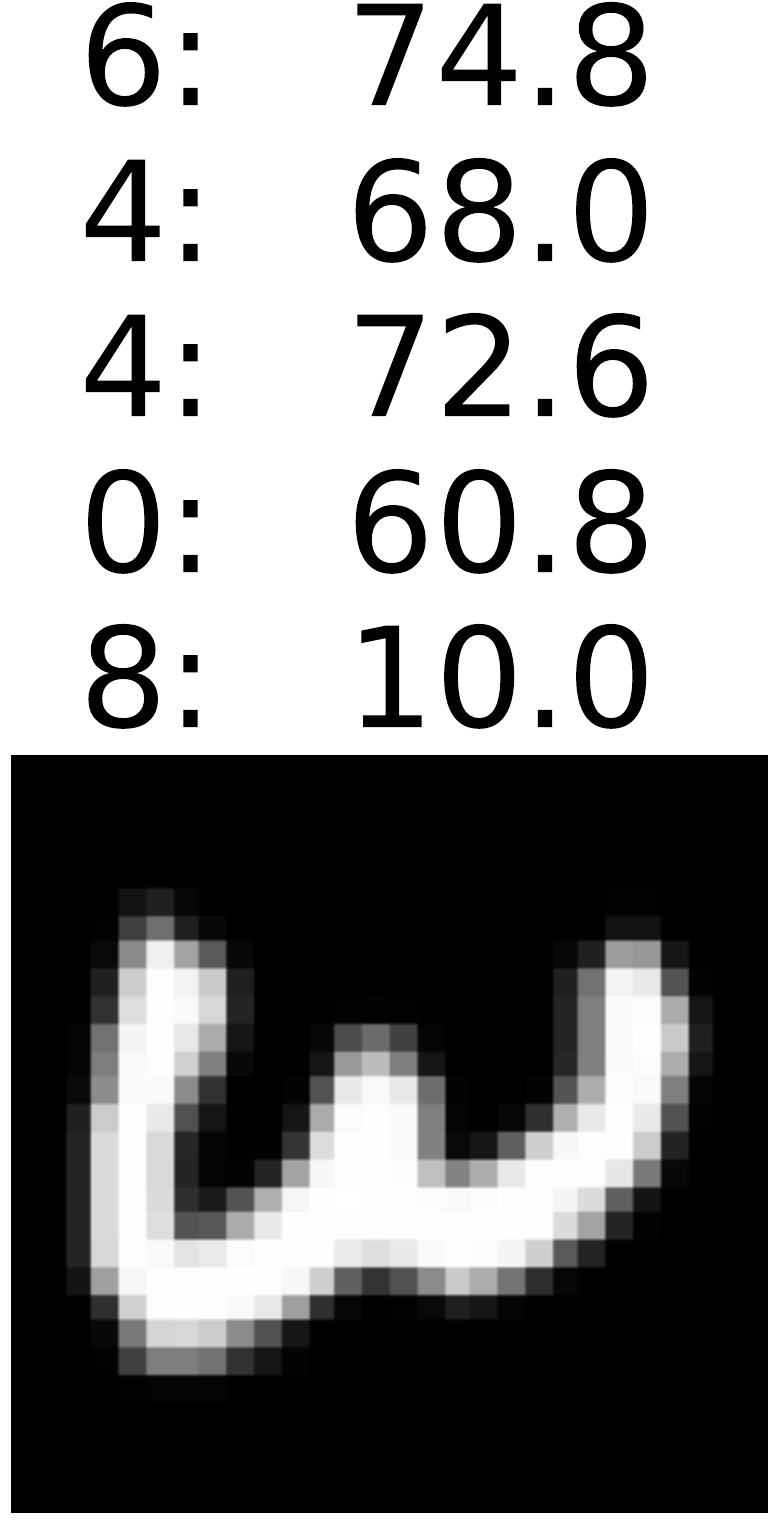} & \includegraphics[width=\plotwidth]{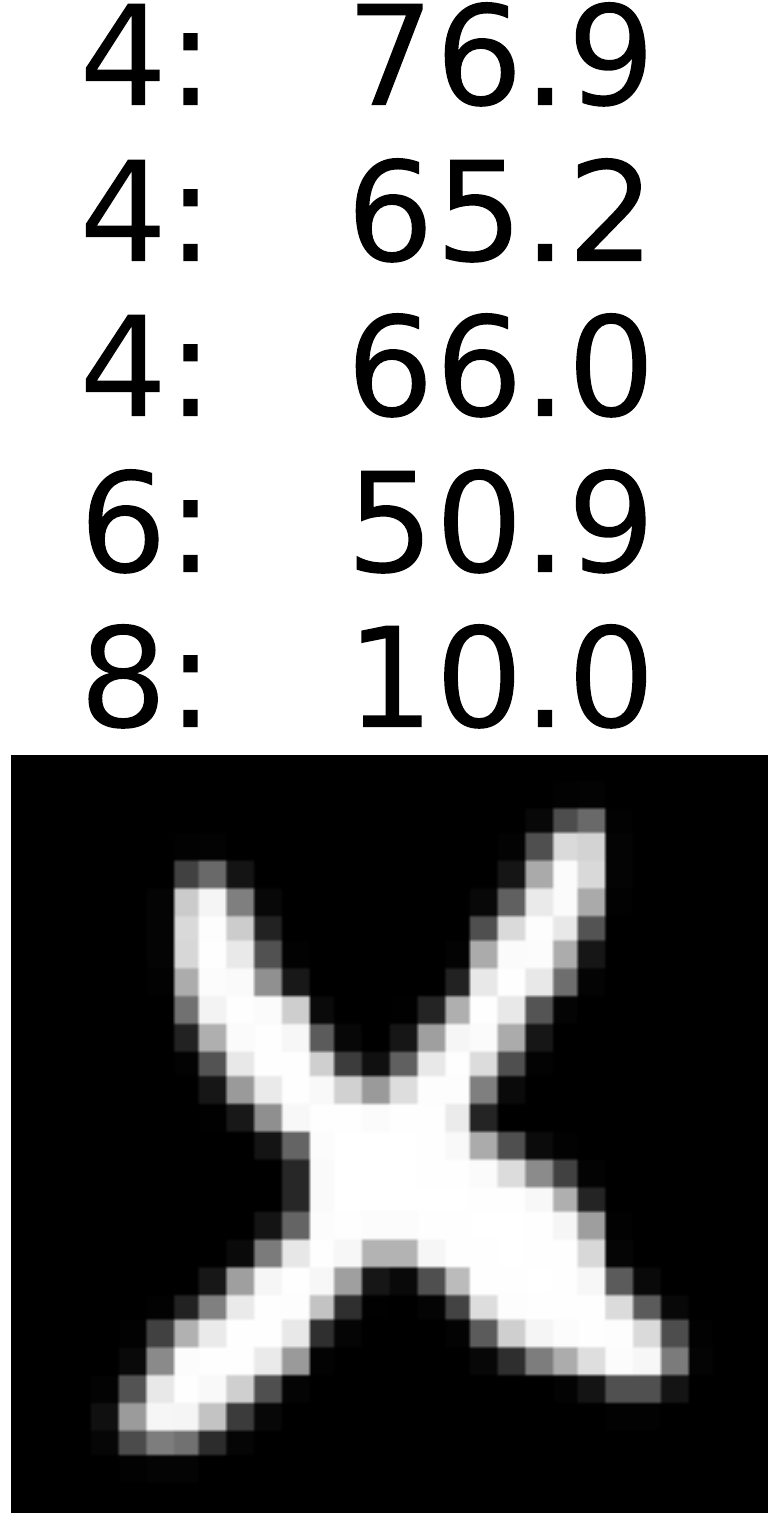} & \includegraphics[width=\plotwidth]{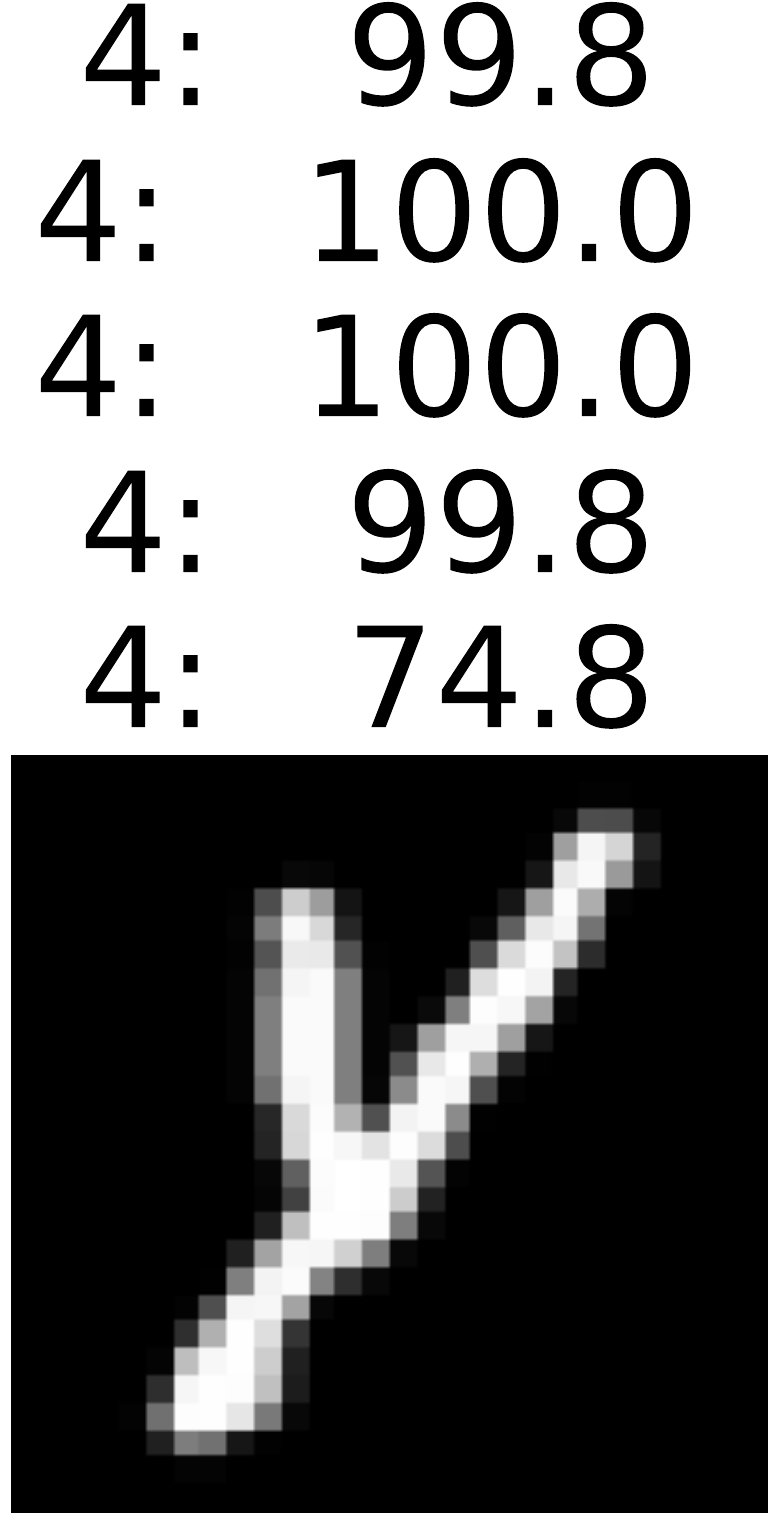} &
\includegraphics[width=\plotwidth]{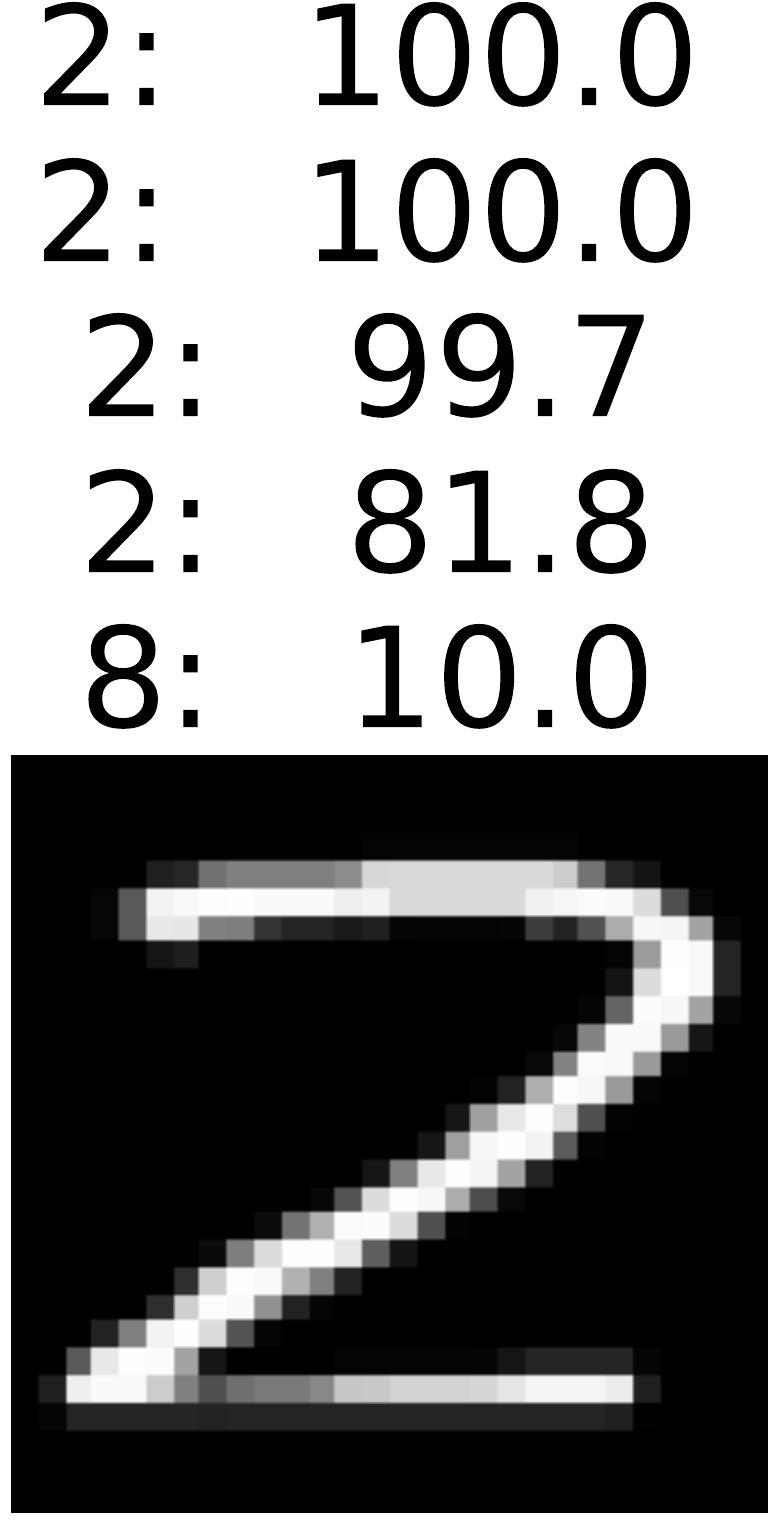} \\
\end{tabular}}

\caption{\label{Fig:Samples_cont} Continuation of Figure \ref{Fig:Samples}. Random samples from the remaining letters in the out-distribution dataset EMNIST. The predictions and confidences of different methods trained on MNIST are shown on top. }
\end{figure*}

\input{figures/EMNIST_conf.tex}

%% file: tables/table_quantile_images.tex
\begin{table*}[hb]
\caption{Exemplary batch of out-distribution 80M Tiny Images (after augmentation) towards the end of training of GOOD\textsubscript{60} models. \textbf{Top:} The 52 Images with highest confidence upper bound. On these, loss is based on standard output. \textbf{Bottom:} The remaining 76 Images with lowest confidence upper bound. Here, loss is based on upper bounds within the $\epsilon$-ball.}
\label{table:quantile_images}
\setlength\tabcolsep{.5pt} 
\vskip 0.15in
\begin{center}
\begin{small}
\begin{sc}
\makebox[\textwidth][c]{
\begin{tabularx}{1.0\textwidth}{CCC}
\textbf{in: MNIST} & \textbf{in: SVHN} & \textbf{in: CIFAR-10} \\
  \includegraphics[width=0.99\linewidth]{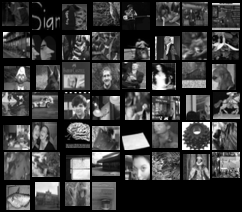}
& \includegraphics[width=0.99\linewidth]{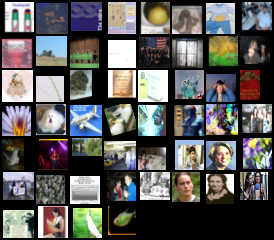}
& \includegraphics[width=0.99\linewidth]{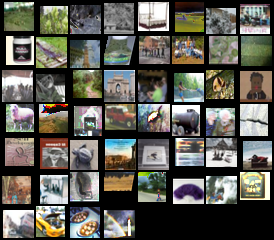}
 \\
  \includegraphics[width=0.99\linewidth]{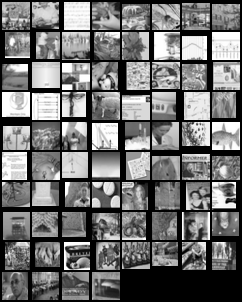}
& \includegraphics[width=0.99\linewidth]{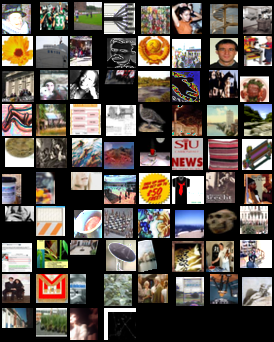}
& \includegraphics[width=0.99\linewidth]{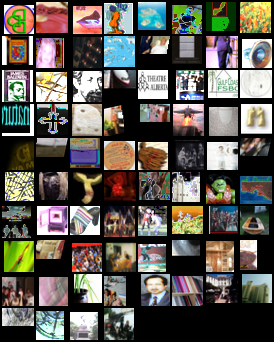}
\end{tabularx}
}
\end{sc}
\end{small}
\end{center}
\vskip -0.1in
\end{table*}

%% file: figures/EMNIST_conf.tex
\begin{figure*}[ht]
\centering \includegraphics[height=.25\textwidth]{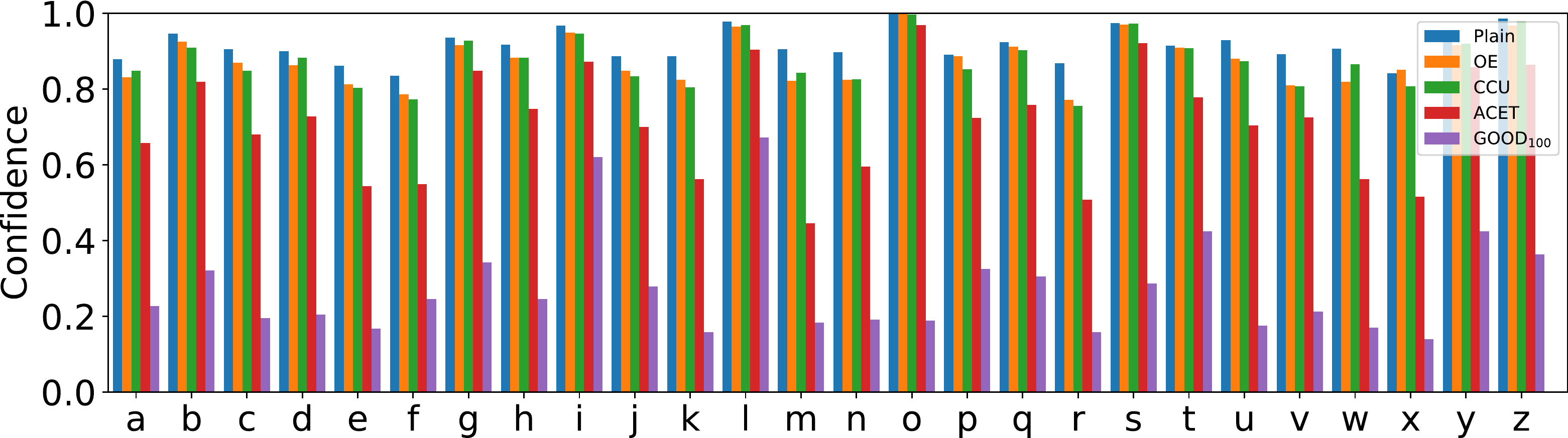}
\caption{\label{Fig:EMNIST_conf} Mean confidence of different models across the classes of EMNIST-Letters. GOOD\textsubscript{100} only has high mean confidence on letters that can easily be mistaken for digits.}
\end{figure*}

%% file: sections/distribution_of_confidences.tex
\section{Distributions of confidences and confidence upper bounds}
Table~\ref{tab:MMC} shows the mean confidences of all models on the in-distribution as well as the mean confidences and the mean guaranteed upper bounds on the worst-case confidences on the evaluated out-distributions.
As discussed, GOOD\textsubscript{100} training can reduce the confidence on the in-distribution, with a particularly strong effect for CIFAR-10.
By adjusting the loss quantile, this effect can be significantly reduced while maintaining non-trivial guarantees.

The histograms of mean confidences on the in-distribution and mean guaranteed upper bounds on the worst-case confidences on the samples from the evaluated out-distribution test sets for seven models are shown in Tables~\ref{table:ub_conf_histograms_mnist}~(MNIST), \ref{table:ub_conf_histograms_svhn}~(SVHN) and \ref{table:ub_conf_histograms_cifar10} (CIFAR-10).
A higher GOOD loss quantile generally shifts the distribution of the upper bounds on the worst-case confidence towards smaller values, but in some cases, especially for GOOD\textsubscript{100} on CIFAR-10, strongly lowers confidences in in-distribution predictions as well.

\begin{table*}[!htbp]
\caption{Mean confidence on the in-distribution and mean confidence / mean upper bounds on the confidence within the $l_\infty$-balls of radius $\epsilon$ on the evaluated out-distribution datasets.}
\label{tab:MMC}
\setlength\tabcolsep{.5pt} 
\vskip 0.15in
\begin{center}
\begin{small}
\begin{sc}
\makebox[\textwidth][c]{
\begin{tabularx}{1.01\textwidth}{l|C|CCCC}
\toprule
\multicolumn{6}{c}{in: MNIST \hspace{.8cm} $\epsilon = 0.3$} \\
\midrule
 Method \ \ & MNIST & FashionMNIST & EMNIST Letters & CIFAR-10 & Uniform Noise \\ \midrule
Plain
 & 99.7
 & 79.2\ / 100.0
 & 91.5\ / 100.0
 & 77.2\ / 100.0
 & 79.6\ / 100.0\\
CEDA
 & 99.7
 & 22.0\ / 100.0
 & 88.3\ / 100.0
 & 10.0\ / 100.0
 & 10.0\ / 100.0\\
OE
 & 99.7
 & 25.4\ / 100.0
 & 87.9\ / 100.0
 & 10.1\ / 100.0
 & 10.0\ / 100.0\\
ACET
 & 99.6
 & 12.3\ / 100.0
 & 75.0\ / 100.0
 & 10.0\ / 100.0
 & 10.0\ / 100.0\\
CCU
 & 99.7
 & 17.5\ / 100.0
 & 87.4\ / 100.0
 & 10.0\ / 100.0
 & 10.0\ / 100.0
\\
GOOD\textsubscript{0}
 & 99.7
 & 20.6\ / 100.0
 & 87.9\ / 100.0
 & 10.0\ / 100.0
 & 10.0\ / 100.0\\
GOOD\textsubscript{20}
 & 99.5
 & 19.4\ / \phantom{0}93.0
 & 70.2\ / 100.0
 & 10.0\ / \phantom{0}76.6
 & 10.0\ / \phantom{0}10.0\\
GOOD\textsubscript{40}
 & 99.3
 & 17.7\ / \phantom{0}76.8
 & 58.2\ / 100.0
 & 10.0\ / \phantom{0}43.7
 & 10.0\ / \phantom{0}10.0\\
GOOD\textsubscript{60}
 & 99.2
 & 15.8\ / \phantom{0}66.0
 & 51.7\ / 100.0
 & 10.0\ / \phantom{0}24.8
 & 10.0\ / \phantom{0}10.0\\
GOOD\textsubscript{80}
 & 99.0
 & 16.3\ / \phantom{0}55.1
 & 40.7\ / \phantom{0}98.6
 & 10.0\ / \phantom{0}15.7
 & 10.0\ / \phantom{0}10.0\\
GOOD\textsubscript{90}
 & 98.8
 & 14.2\ / \phantom{0}47.5
 & 38.3\ / \phantom{0}98.2
 & 10.0\ / \phantom{0}12.7
 & 10.0\ / \phantom{0}10.0\\
GOOD\textsubscript{95}
 & 98.7
 & 13.1\ / \phantom{0}42.6
 & 32.2\ / \phantom{0}98.2
 & 10.0\ / \phantom{0}11.6
 & 10.0\ / \phantom{0}10.0\\
\rowcolor{lightgrey}GOOD\textsubscript{100}
 & 98.4
 & 10.8\ / \phantom{0}40.8
 & 27.1\ / \phantom{0}99.2
 & 10.0\ / \phantom{0}11.0
 & 10.0\ / \phantom{0}10.0\\
\bottomrule
\toprule
\multicolumn{6}{c}{in: SVHN \hspace{1cm} $\epsilon = 0.03$} \\
\midrule
Method & SVHN & CIFAR-100 & CIFAR-10 & LSUN Classroom & Uniform Noise \\ \midrule
Plain
 & 97.7
 & 70.8\ / 100.0
 & 70.5\ / 100.0
 & 66.8\ / 100.0
 & 40.5\ / 100.0
\\
CEDA
 & 97.1
 & 10.2\ / 100.0
 & 10.1\ / 100.0
 & 10.0\ / 100.0
 & 10.0\ / 100.0\\
OE
 & 97.0
 & 10.7\ / 100.0
 & 10.5\ / 100.0
 & 10.3\ / 100.0
 & 10.2\ / 100.0\\
ACET
 & 93.5
 & 10.2\ / 100.0
 & 10.1\ / 100.0
 & 10.1\ / 100.0
 & 10.5\ / 100.0\\
CCU
 & 97.2
 & 10.8\ / 100.0
 & 10.6\ / 100.0
 & 10.4\ / 100.0
 & 10.0\ / 100.0
\\
GOOD\textsubscript{0}
 & 98.7
 & 10.0\ / 100.0
 & 10.0\ / 100.0
 & 10.0\ / 100.0
 & 10.0\ / 100.0\\
GOOD\textsubscript{20}
 & 97.6
 & 10.1\ / \phantom{0}78.1
 & 10.1\ / \phantom{0}81.9
 & 10.0\ / \phantom{0}80.7
 & 10.0\ / \phantom{0}10.0\\
GOOD\textsubscript{40}
 & 97.6
 & 10.1\ / \phantom{0}61.4
 & 10.1\ / \phantom{0}57.4
 & 10.0\ / \phantom{0}54.0
 & 10.0\ / \phantom{0}10.2\\
GOOD\textsubscript{60}
 & 97.4
 & 10.1\ / \phantom{0}44.2
 & 10.1\ / \phantom{0}39.8
 & 10.0\ / \phantom{0}33.7
 & 10.0\ / \phantom{0}10.1\\
GOOD\textsubscript{80}
 & 96.1
 & 10.1\ / \phantom{0}28.1
 & 10.1\ / \phantom{0}23.6
 & 10.0\ / \phantom{0}17.4
 & 10.0\ / \phantom{0}10.2\\
GOOD\textsubscript{90}
 & 94.7
 & 10.1\ / \phantom{0}20.9
 & 10.0\ / \phantom{0}17.7
 & 10.0\ / \phantom{0}14.0
 & 10.0\ / \phantom{0}10.0\\
GOOD\textsubscript{95}
 & 93.4
 & 10.2\ / \phantom{0}18.2
 & 10.1\ / \phantom{0}15.7
 & 10.1\ / \phantom{0}12.6
 & 10.0\ / \phantom{0}10.0\\
\rowcolor{lightgrey}GOOD\textsubscript{100}
 & 91.5
 & 10.7\ / \phantom{0}16.7
 & 10.3\ / \phantom{0}14.5
 & 10.1\ / \phantom{0}12.1
 & 10.0\ / \phantom{0}10.1\\
\bottomrule
\toprule
\multicolumn{6}{c}{in: CIFAR-10 \hspace{.5cm} $\epsilon = 0.01$} \\
\midrule
Method & CIFAR-10 & CIFAR-100 & SVHN & LSUN Classroom & Uniform Noise \\ \midrule
Plain
 & 95.1
 & 79.0\ / 100.0
 & 75.8\ / 100.0
 & 73.9\ / 100.0
 & 73.2\ / 100.0
\\
CEDA
 & 87.0
 & 29.0\ / 100.0
 & 12.1\ / 100.0
 & 10.5\ / 100.0
 & 11.9\ / 100.0\\
OE
 & 85.1
 & 31.6\ / 100.0
 & 19.1\ / 100.0
 & 14.6\ / 100.0
 & 15.6\ / 100.0\\
ACET
 & 71.8
 & 25.3\ / 100.0
 & 16.7\ / 100.0
 & 13.7\ / 100.0
 & 11.2\ / 100.0\\
CCU
 & 89.4
 & 32.5\ / 100.0
 & 20.5\ / 100.0
 & 12.6\ / 100.0
 & 10.0\ / 100.0
\\
GOOD\textsubscript{0}
 & 81.0
 & 18.9\ / 100.0
 & 10.8\ / 100.0
 & 10.1\ / 100.0
 & 10.0\ / 100.0\\
GOOD\textsubscript{20}
 & 78.9
 & 23.8\ / \phantom{0}91.4
 & 13.0\ / \phantom{0}87.8
 & 10.7\ / \phantom{0}97.9
 & 10.1\ / \phantom{0}22.7\\
GOOD\textsubscript{40}
 & 77.1
 & 21.4\ / \phantom{0}84.7
 & 11.2\ / \phantom{0}85.4
 & 10.7\ / \phantom{0}89.5
 & 11.7\ / \phantom{0}12.4\\
GOOD\textsubscript{60}
 & 71.7
 & 21.7\ / \phantom{0}75.4
 & 11.5\ / \phantom{0}72.0
 & 10.5\ / \phantom{0}67.3
 & 13.2\ / \phantom{0}13.4\\
\rowcolor{lightgrey}
GOOD\textsubscript{80}
 & 64.1
 & 23.1\ / \phantom{0}64.4
 & 13.3\ / \phantom{0}67.5
 & 13.5\ / \phantom{0}51.8
 & 12.0\ / \phantom{0}12.3\\
GOOD\textsubscript{90}
 & 55.6
 & 24.2\ / \phantom{0}54.8
 & 15.4\ / \phantom{0}56.2
 & 16.1\ / \phantom{0}44.9
 & 17.2\ / \phantom{0}18.1\\
GOOD\textsubscript{95}
 & 53.1
 & 25.8\ / \phantom{0}52.0
 & 16.9\ / \phantom{0}57.2
 & 18.1\ / \phantom{0}43.6
 & 12.6\ / \phantom{0}12.6\\
GOOD\textsubscript{100}
 & 49.6
 & 34.7\ / \phantom{0}46.0
 & 30.4\ / \phantom{0}44.0
 & 30.6\ / \phantom{0}41.5
 & 11.6\ / \phantom{0}12.0\\
\bottomrule
\end{tabularx}
}
\end{sc}
\end{small}
\end{center}
\vskip -0.1in
\end{table*}

\input{tables/table_ub_conf_histograms}

%% file: tables/table_ub_conf_histograms.tex
\renewcommand{\tabularxcolumn}[1]{>{\small}m{#1}}

\begin{table*}[t]
\caption{Histograms of the confidences on the \textbf{MNIST} in-distribution and guaranteed upper bounds on the confidences on OOD datasets within the $l_\infty$-ball of radius $0.3$.
Each histogram uses 50 bins between 0.1 and 1.0. For better readability, the scale is zoomed in by a factor 10 for numbers below one fifth of the total number of datapoints of the shown datasets.
The vertical dotted line shows the mean value of the histogram's data.
}
\label{table:ub_conf_histograms_mnist}
\setlength\tabcolsep{.5pt} 
\vskip 0.15in
\begin{center}
\begin{small}
\begin{sc}
\makebox[\textwidth][c]{
\begin{tabularx}{1.4\textwidth}{cCCCCC}
Model   &   MNIST &    FashionMNIST gub   &   EMNIST Letters gub   & CIFAR-10 gub  & Uniform gub   \\
Plain
 &  \includegraphics[width=1.0\linewidth]{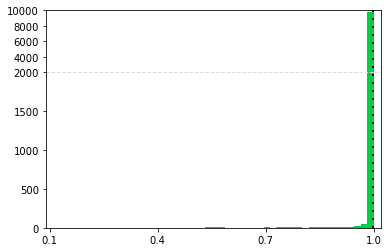}
 &  \includegraphics[width=1.0\linewidth]{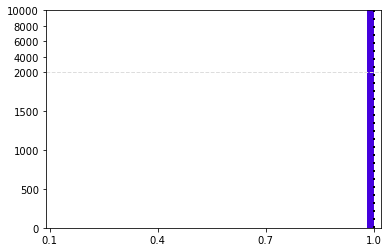}
 &  \includegraphics[width=1.0\linewidth]{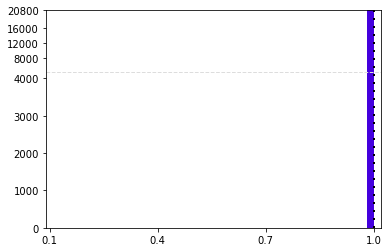}
 &  \includegraphics[width=1.0\linewidth]{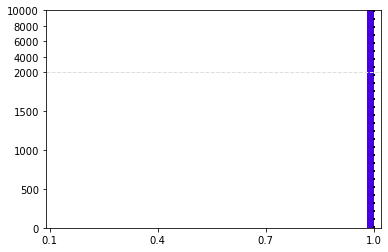}
 &  \includegraphics[width=1.0\linewidth]{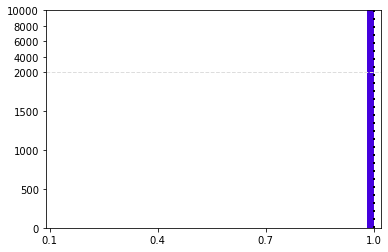}
\\
OE
 &  \includegraphics[width=1.0\linewidth]{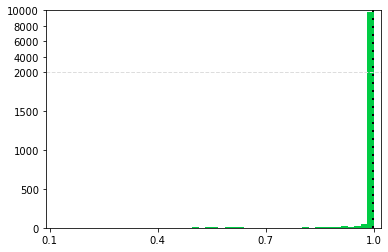}
 &  \includegraphics[width=1.0\linewidth]{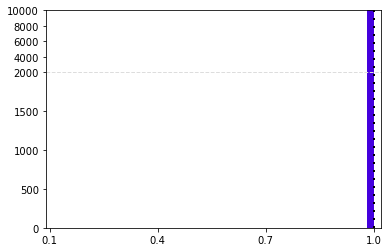}
 &  \includegraphics[width=1.0\linewidth]{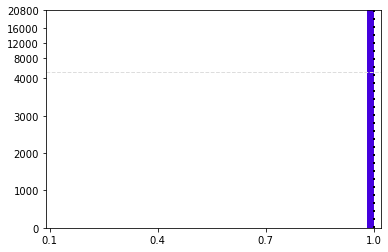}
 &  \includegraphics[width=1.0\linewidth]{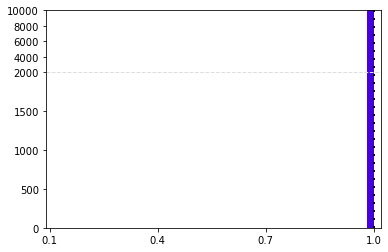}
 &  \includegraphics[width=1.0\linewidth]{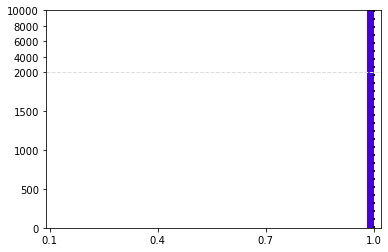}
\\
ACET
 &  \includegraphics[width=1.0\linewidth]{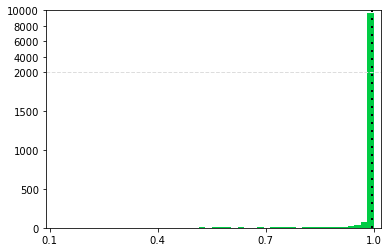}
 &  \includegraphics[width=1.0\linewidth]{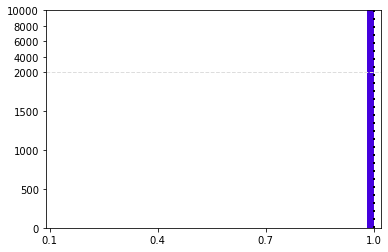}
 &  \includegraphics[width=1.0\linewidth]{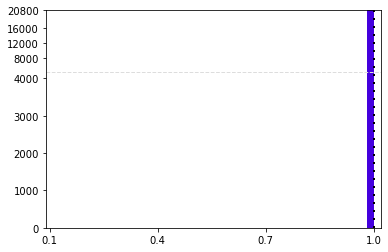}
 &  \includegraphics[width=1.0\linewidth]{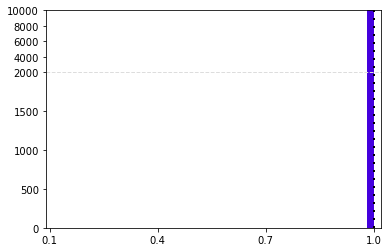}
 &  \includegraphics[width=1.0\linewidth]{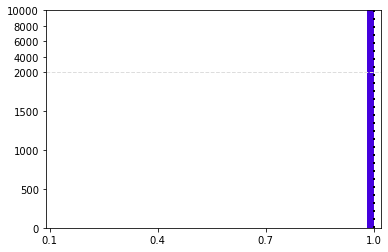}
\\
GOOD\textsubscript{40}
 &  \includegraphics[width=1.0\linewidth]{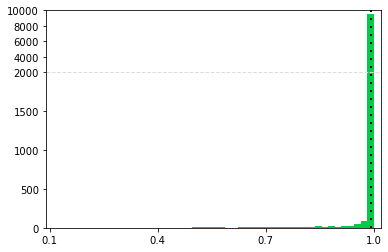}
 &  \includegraphics[width=1.0\linewidth]{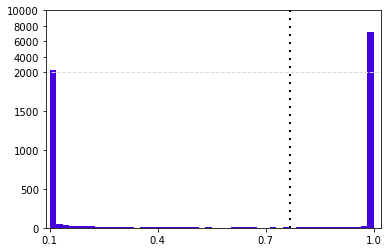}
 &  \includegraphics[width=1.0\linewidth]{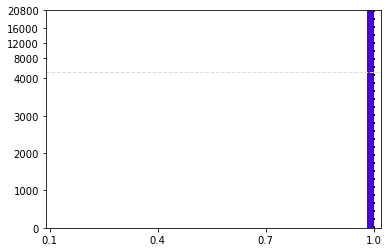}
 &  \includegraphics[width=1.0\linewidth]{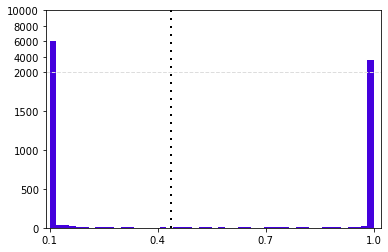}
 &  \includegraphics[width=1.0\linewidth]{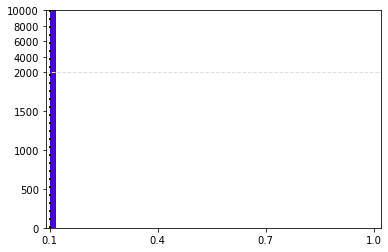}
\\
GOOD\textsubscript{80}
 &  \includegraphics[width=1.0\linewidth]{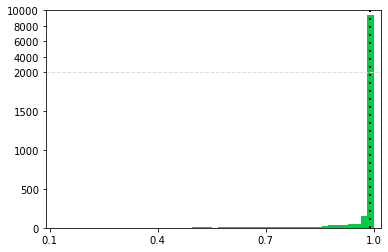}
 &  \includegraphics[width=1.0\linewidth]{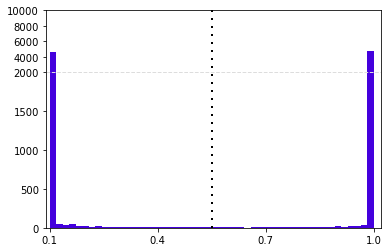}
 &  \includegraphics[width=1.0\linewidth]{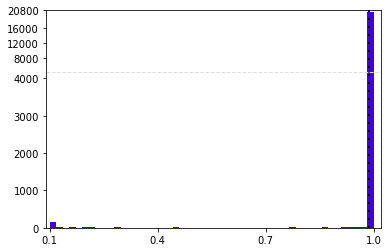}
 &  \includegraphics[width=1.0\linewidth]{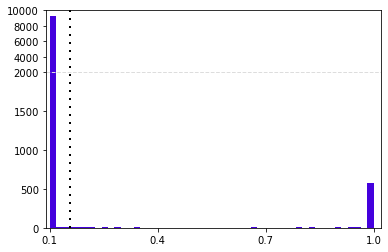}
 &  \includegraphics[width=1.0\linewidth]{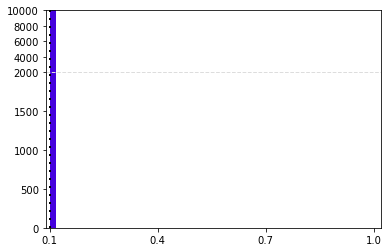}
\\
GOOD\textsubscript{90}
 &  \includegraphics[width=1.0\linewidth]{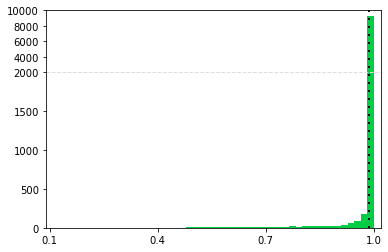}
 &  \includegraphics[width=1.0\linewidth]{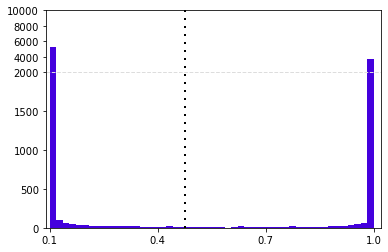}
 &  \includegraphics[width=1.0\linewidth]{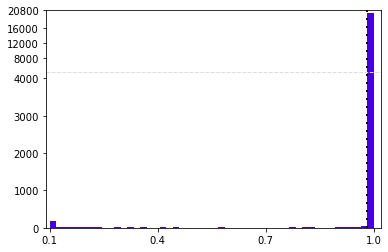}
 &  \includegraphics[width=1.0\linewidth]{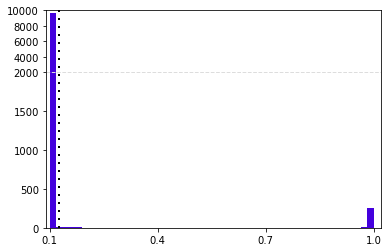}
 &  \includegraphics[width=1.0\linewidth]{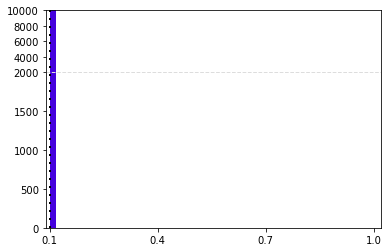}
\\
GOOD\textsubscript{100}
 &  \includegraphics[width=1.0\linewidth]{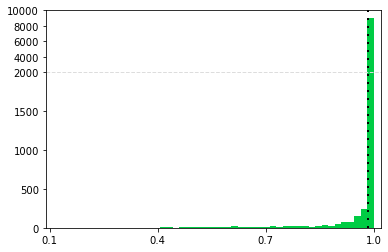}
 &  \includegraphics[width=1.0\linewidth]{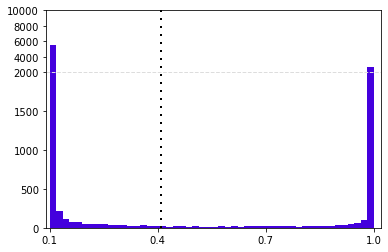}
 &  \includegraphics[width=1.0\linewidth]{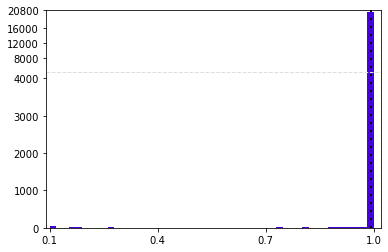}
 &  \includegraphics[width=1.0\linewidth]{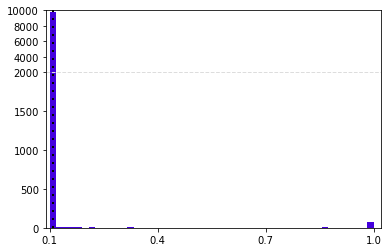}
 &  \includegraphics[width=1.0\linewidth]{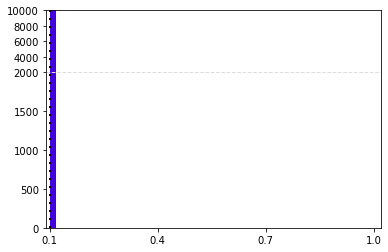}
\\
\end{tabularx}
}
\end{sc}
\end{small}
\end{center}
\vskip -0.1in
\end{table*}

\begin{table*}[t]
\caption{Histograms of the confidences on the \textbf{SVHN} in-distribution and guaranteed upper bounds on the confidences on OOD datasets within the $l_\infty$-ball of radius $0.03$.
Each histogram uses 50 bins between 0.1 and 1.0. For better readability, the scale is zoomed in by a factor 10 for numbers below one fifth of the total number of datapoints of the shown datasets.
The vertical dotted line shows the mean value of the histogram's data.
}
\label{table:ub_conf_histograms_svhn}
\setlength\tabcolsep{.5pt} 
\vskip 0.15in
\begin{center}
\begin{small}
\begin{sc}
\makebox[\textwidth][c]{
\begin{tabularx}{1.4\textwidth}{cCCCCC}
Model   &   SVHN &    CIFAR-100 gub   &   CIFAR-10 gub   & LSUN Classroom gub  & Uniform gub   \\
Plain
 &  \includegraphics[width=1.0\linewidth]{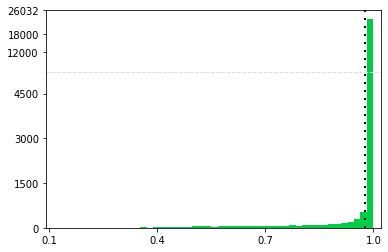}
 &  \includegraphics[width=1.0\linewidth]{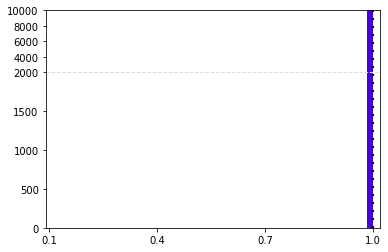} 
 &  \includegraphics[width=1.0\linewidth]{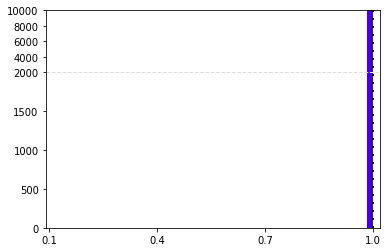} 
 &  \includegraphics[width=1.0\linewidth]{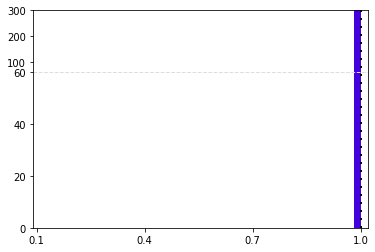} 
 &  \includegraphics[width=1.0\linewidth]{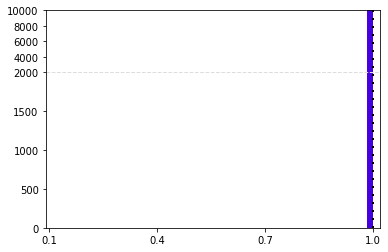} 
\\  
OE
 &  \includegraphics[width=1.0\linewidth]{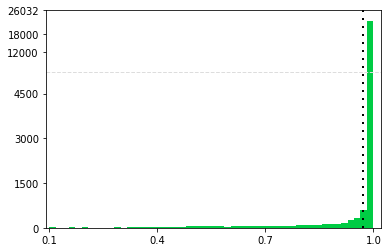}
 &  \includegraphics[width=1.0\linewidth]{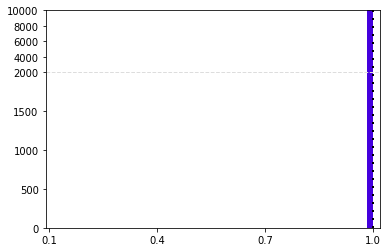}
 &  \includegraphics[width=1.0\linewidth]{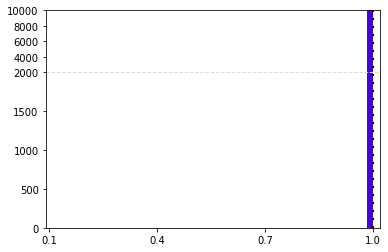}
 &  \includegraphics[width=1.0\linewidth]{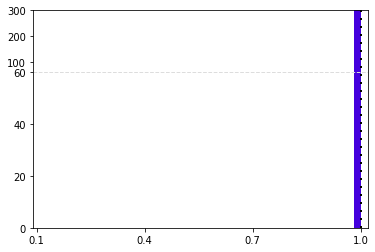}
 &  \includegraphics[width=1.0\linewidth]{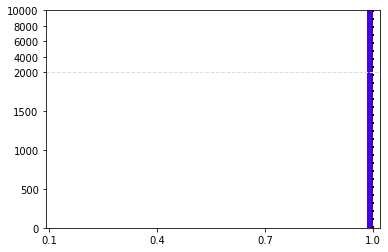}
\\
ACET
 &  \includegraphics[width=1.0\linewidth]{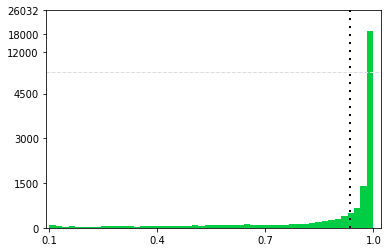}
 &  \includegraphics[width=1.0\linewidth]{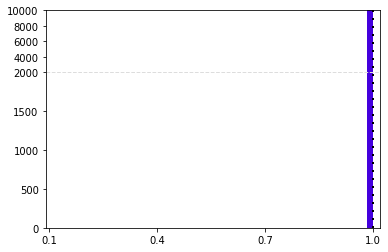}
 &  \includegraphics[width=1.0\linewidth]{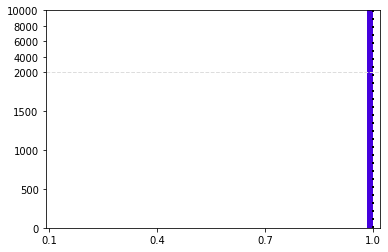}
 &  \includegraphics[width=1.0\linewidth]{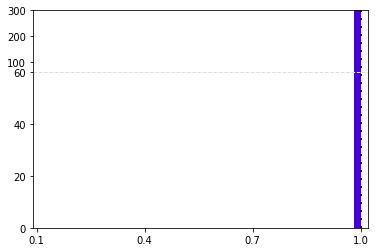}
 &  \includegraphics[width=1.0\linewidth]{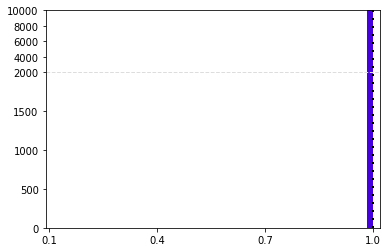}
\\
GOOD\textsubscript{40}
 &  \includegraphics[width=1.0\linewidth]{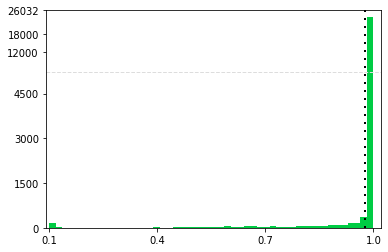}
 &  \includegraphics[width=1.0\linewidth]{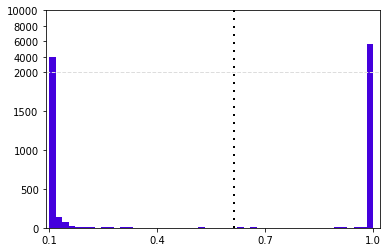}
 &  \includegraphics[width=1.0\linewidth]{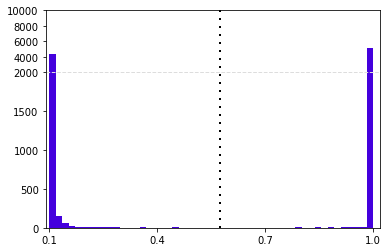}
 &  \includegraphics[width=1.0\linewidth]{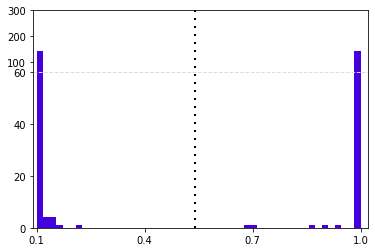}
 &  \includegraphics[width=1.0\linewidth]{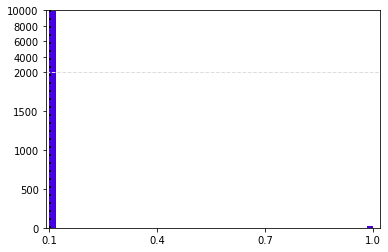}
\\
GOOD\textsubscript{80}
 &  \includegraphics[width=1.0\linewidth]{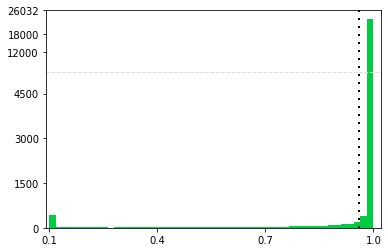}
 &  \includegraphics[width=1.0\linewidth]{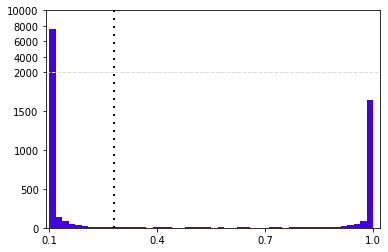}
 &  \includegraphics[width=1.0\linewidth]{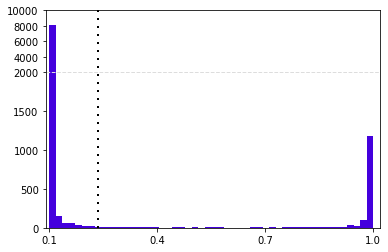}
 &  \includegraphics[width=1.0\linewidth]{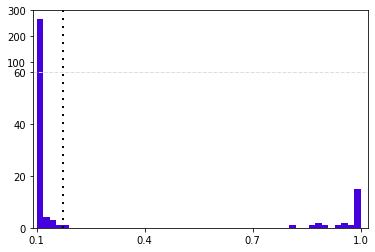}
 &  \includegraphics[width=1.0\linewidth]{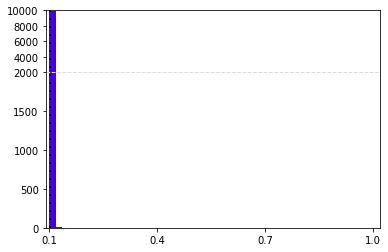}
\\
GOOD\textsubscript{90}
 &  \includegraphics[width=1.0\linewidth]{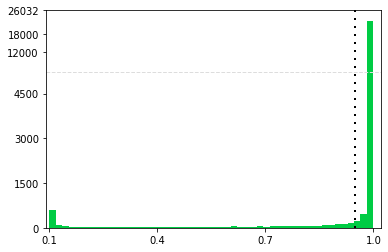}
 &  \includegraphics[width=1.0\linewidth]{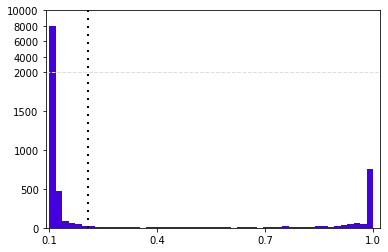}
 &  \includegraphics[width=1.0\linewidth]{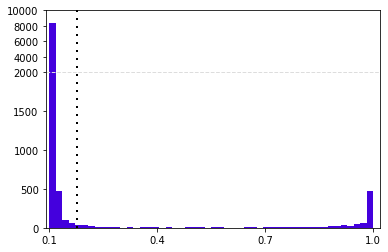}
 &  \includegraphics[width=1.0\linewidth]{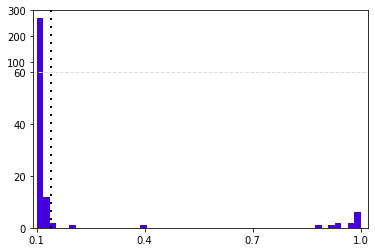}
 &  \includegraphics[width=1.0\linewidth]{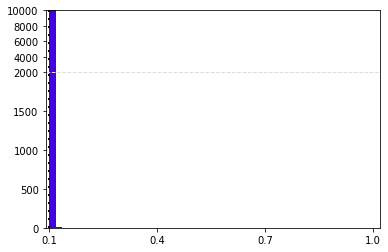}
\\
GOOD\textsubscript{100}
 &  \includegraphics[width=1.0\linewidth]{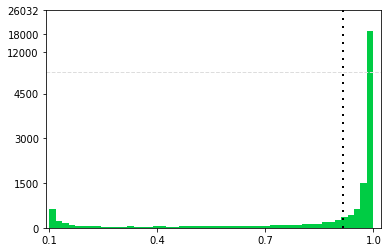}
 &  \includegraphics[width=1.0\linewidth]{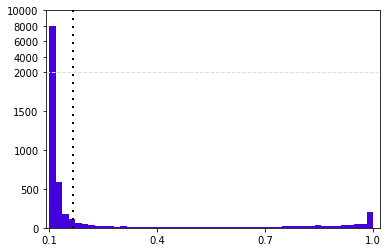}
 &  \includegraphics[width=1.0\linewidth]{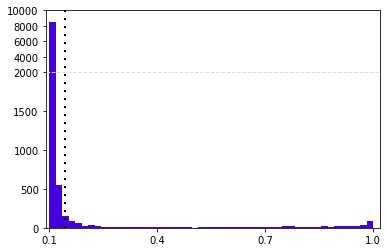}
 &  \includegraphics[width=1.0\linewidth]{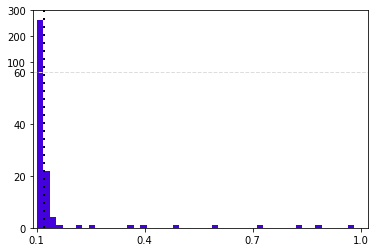}
 &  \includegraphics[width=1.0\linewidth]{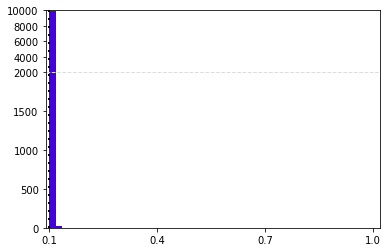}
\\
\end{tabularx}
}
\end{sc}
\end{small}
\end{center}
\vskip -0.1in
\end{table*}

\begin{table*}[t]
\caption{Histograms of the confidences on the \textbf{CIFAR-10} in-distribution and guaranteed upper bounds on the confidences on OOD datasets within the $l_\infty$-ball of radius $0.01$.
Each histogram uses 50 bins between 0.1 and 1.0. For better readability, the scale is zoomed in by a factor 10 for numbers below one fifth of the total number of datapoints of the shown datasets.
The vertical dotted line shows the mean value of the histogram's data.
}
\label{table:ub_conf_histograms_cifar10}
\setlength\tabcolsep{.5pt} 
\vskip 0.15in
\begin{center}
\begin{small}
\begin{sc}
\makebox[\textwidth][c]{
\begin{tabularx}{1.4\textwidth}{cCCCCC}
Model   &   CIFAR-10 &    CIFAR-100 gub   &   SVHN gub   & LSUN Classroom gub  & Uniform gub   \\
Plain
 &  \includegraphics[width=1.0\linewidth]{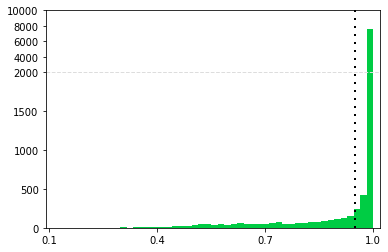}
 &  \includegraphics[width=1.0\linewidth]{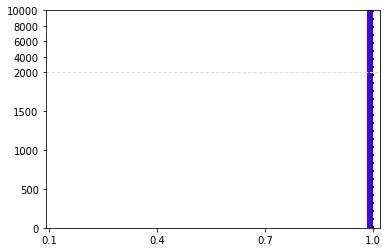} 
 &  \includegraphics[width=1.0\linewidth]{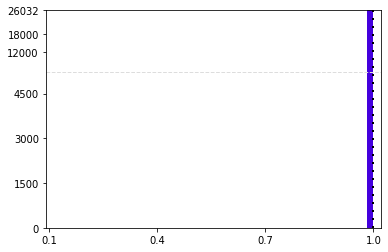} 
 &  \includegraphics[width=1.0\linewidth]{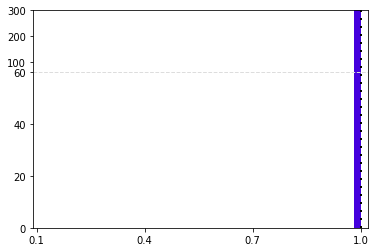} 
 &  \includegraphics[width=1.0\linewidth]{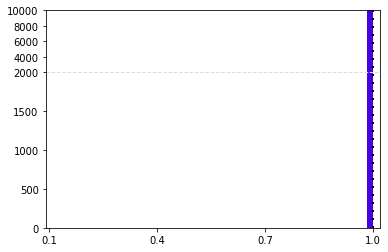} 
\\
OE
 &  \includegraphics[width=1.0\linewidth]{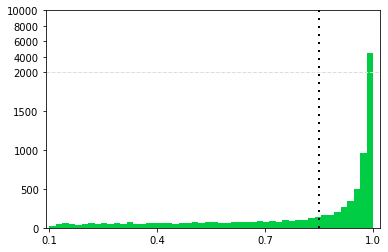}
 &  \includegraphics[width=1.0\linewidth]{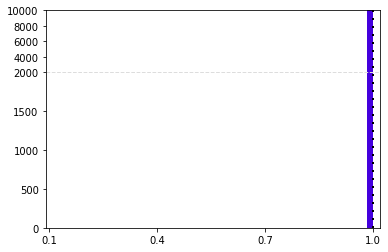}
 &  \includegraphics[width=1.0\linewidth]{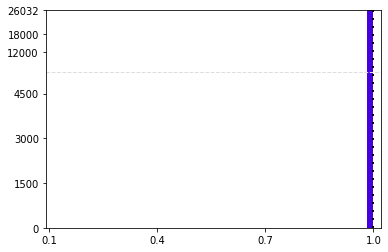}
 &  \includegraphics[width=1.0\linewidth]{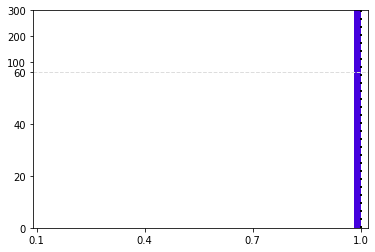}
 &  \includegraphics[width=1.0\linewidth]{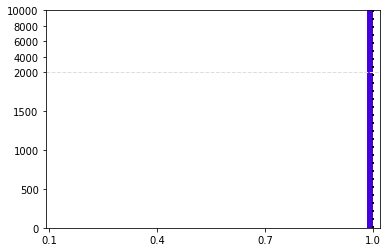}
\\
ACET
 &  \includegraphics[width=1.0\linewidth]{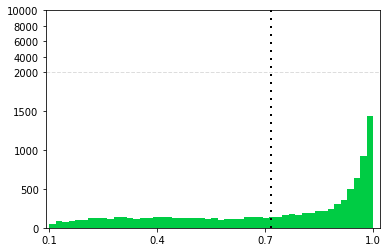}
 &  \includegraphics[width=1.0\linewidth]{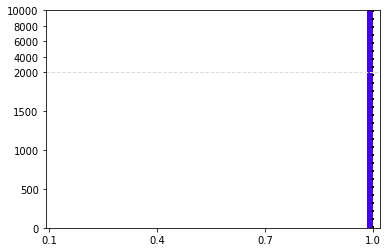}
 &  \includegraphics[width=1.0\linewidth]{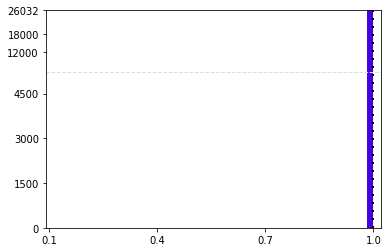}
 &  \includegraphics[width=1.0\linewidth]{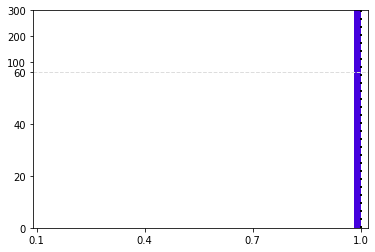}
 &  \includegraphics[width=1.0\linewidth]{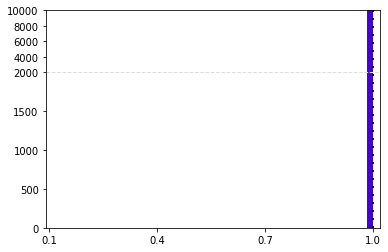}
\\
GOOD\textsubscript{40}
 &  \includegraphics[width=1.0\linewidth]{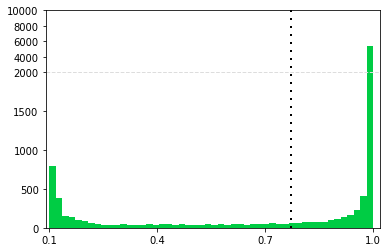}
 &  \includegraphics[width=1.0\linewidth]{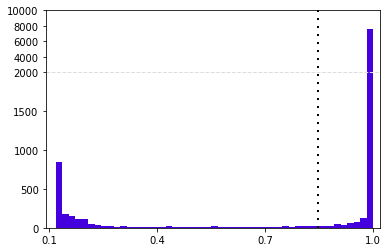}
 &  \includegraphics[width=1.0\linewidth]{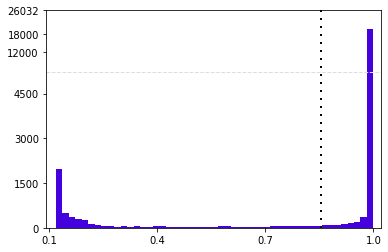}
 &  \includegraphics[width=1.0\linewidth]{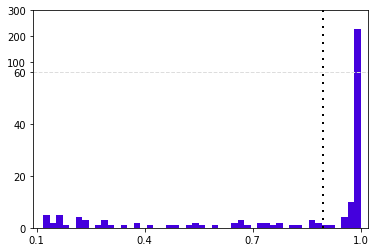}
 &  \includegraphics[width=1.0\linewidth]{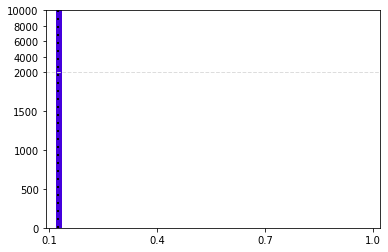}
\\
GOOD\textsubscript{80}
 &  \includegraphics[width=1.0\linewidth]{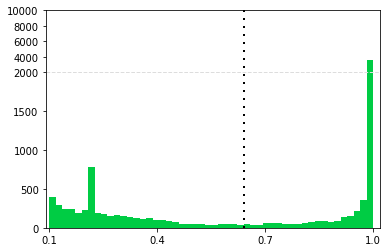}
 &  \includegraphics[width=1.0\linewidth]{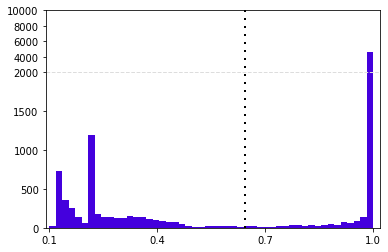}
 &  \includegraphics[width=1.0\linewidth]{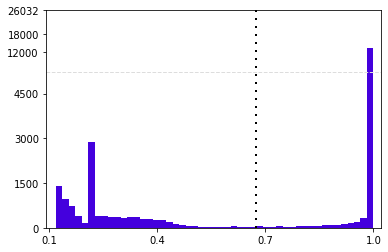}
 &  \includegraphics[width=1.0\linewidth]{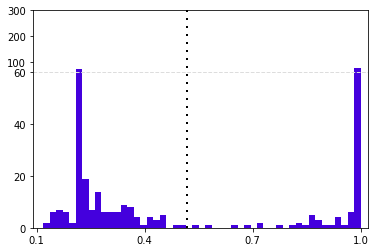}
 &  \includegraphics[width=1.0\linewidth]{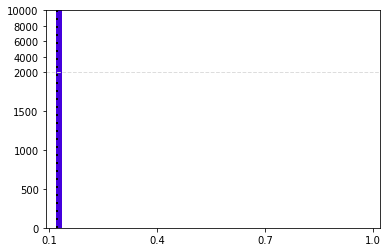}
\\
GOOD\textsubscript{90}
 &  \includegraphics[width=1.0\linewidth]{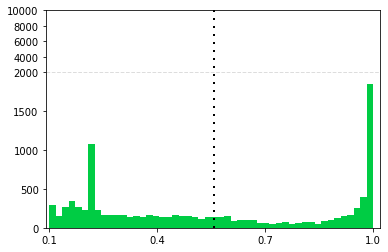}
 &  \includegraphics[width=1.0\linewidth]{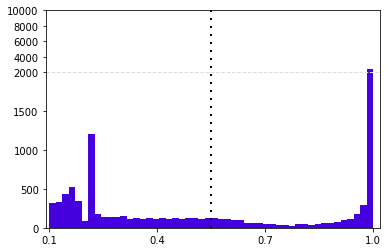}
 &  \includegraphics[width=1.0\linewidth]{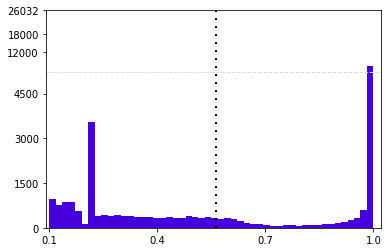}
 &  \includegraphics[width=1.0\linewidth]{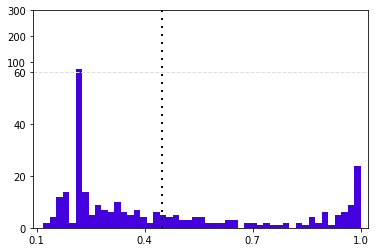}
 &  \includegraphics[width=1.0\linewidth]{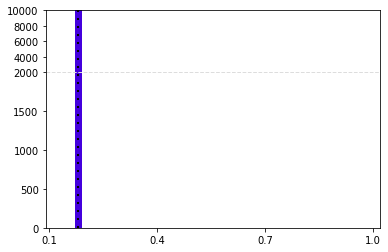}
\\
GOOD\textsubscript{100}
 &  \includegraphics[width=1.0\linewidth]{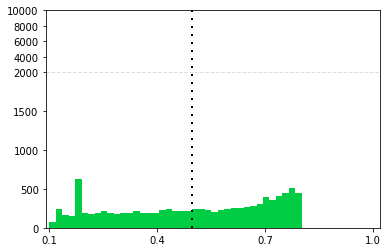}
 &  \includegraphics[width=1.0\linewidth]{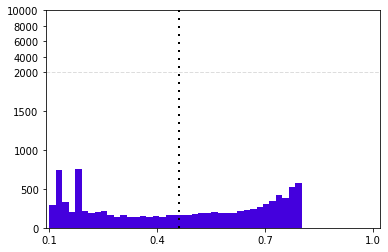}
 &  \includegraphics[width=1.0\linewidth]{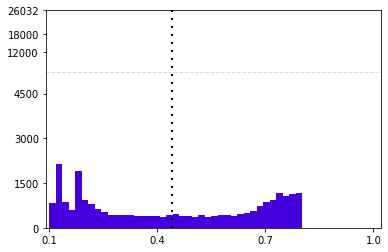}
 &  \includegraphics[width=1.0\linewidth]{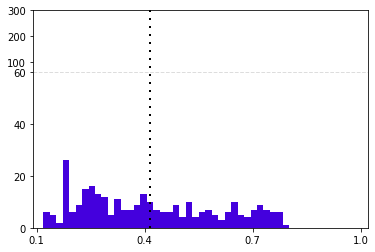}
 &  \includegraphics[width=1.0\linewidth]{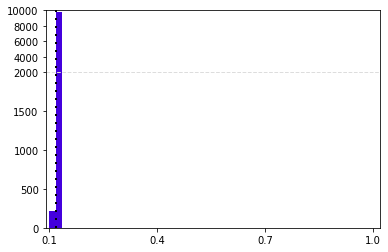}
\\
\end{tabularx}
}
\end{sc}
\end{small}
\end{center}
\vskip -0.1in
\end{table*}

%% file: sections/additional_datasets.tex
\clearpage
\newpage
\section{Evaluation on additional datasets}\label{section:additional_datasets}
\input{tables/table_additional_aucs}
\newpage
Extending the evaluation results presented in Table \ref{table:all_aucs}, we provide AUC, AAUC and GAUC values for additional out-distribution datasets in Table \ref{table:additional_aucs}.
These datasets are:
\begin{itemize}
    \item 80M Tiny Images, the out-distribution that was used during training. 
    While it is the same \textit{distribution} as seen during training, the test set consists of 1,000 samples that are not part of the training set.
    \item Omniglot (Lake, B. M., Salakhutdinov, R., and Tenenbaum, J. B. (2015). Human-level concept learning through probabilistic program induction. Science, 350(6266), 1332-1338.) is a dataset of hand drawn characters. We use the evaluation split consisting of 13180 characters from 20 different alphabets.
    \item notMNIST is a dataset of the letters A to J taken from different publicly available fonts. The dataset was retrieved from \url{https://yaroslavvb.blogspot.com/2011/09/notmnist-dataset.html}.
    We evaluate on the hand cleaned subset of 18724 images, 
    \item ImageNet-~\cite{HeiAndBit2019}, which is a subset of ImageNet~\cite{imagenet_cvpr09} without images labelled as classes equal or semantically similar to CIFAR-10 classes.
    \item Smooth Noise is generated as described by~\cite{HeiAndBit2019}. First, a uniform noise image is generated. Then, a Gaussian filter with $\sigma$ drawn uniformly at random between 1.0 and 2.5 is applied. Finally, the image is re-scaled such that the minimal pixel value is 0.0 and the maximal one is 1.0. We evaluate AUC and GAUC on 30,000 samples.
\end{itemize}
For MNIST, GOOD\textsubscript{100} has an excellent GAUC for the training out-distribution 80M Tiny images as well as for notMNIST. For Omniglot, GOOD\textsubscript{100} is again better than OE/CEDA (similar to EMNIST) in terms of clean AUC's but here ACET is slightly better.
However, again it is very difficult to provide any guarantees for this dataset even though non-trival adversarial AUC's against the employed attacks are maintained. 

For SVHN, the detection of smooth noise turns out to be the most difficult of the evaluated tasks. There, the clean AUCs of all methods except ACET and CCU are lower than the perfect scores we see on other out-distributions but still very high, and only the higher Quantile GOOD models can give some guarantees. An explanation might be that the image features of SVHN house numbers and of this kind of synthetic noise are similarly smooth.
For 80M Tiny Images and Imagenet-, on the other hand, the SVHN high quantile GOOD models, particularly GOOD\textsubscript{100}, are able to provide almost perfect guaranteed AUCs.

For CIFAR-10, on all three out-distributions we again observe the trade-off between clean and guaranteed AUC that comes with the choice of the loss quantile.
Overall, the GOOD\textsubscript{80} model again retains reasonable AUC values for the clean data while also providing useful guaranteed AUCs.

%% file: tables/table_additional_aucs.tex
\begin{table*}[!htbp]
\caption{A continuation of Table \ref{table:all_aucs} for additional out-distributions. As in Table \ref{table:all_aucs} the guaranteed AUCs (GAUC) of the highlighted GOOD models are in general better than the adversarial (AAUC) of OE (with the exception of Omniglot for MNIST).}
\label{table:additional_aucs}
\setlength\tabcolsep{.5pt} 
\vskip 0.15in
\begin{center}
\begin{small}
\begin{sc}
\makebox[\textwidth][c]{
\begin{tabularx}{1.0\textwidth}{lc|CCC|CCC|CCC}
\toprule
\multicolumn{11}{c}{in: MNIST \hspace{.8cm} $\epsilon = 0.3$} \\
\midrule
\multicolumn{1}{l}{\multirow{2}{*}{Method \ }}  & \multicolumn{1}{c|}{\multirow{2}{*}{\ Acc. \ }}
 & \multicolumn{3}{c|}{80M Tiny Images}  & \multicolumn{3}{c|}{Omniglot}  & \multicolumn{3}{c}{notMNIST} \\
& & auc & aauc & gauc & auc & aauc & gauc & auc & aauc & gauc \\ \midrule
Plain & \noindent\phantom{0}99.4 & \noindent\phantom{0}98.7 & \noindent\phantom{0}36.9 & \noindent\phantom{00}0.0 & \noindent\phantom{0}97.9 & \noindent\phantom{0}38.6 & \noindent\phantom{00}0.0 & \noindent\phantom{0}91.9 & \noindent\phantom{0}38.8 & \noindent\phantom{00}0.0 \\
CEDA & \noindent\phantom{0}99.4 & \textbf{\textbf{100.0}}             & \noindent\phantom{0}94.3 & \noindent\phantom{00}0.0 & \noindent\phantom{0}98.5 & \noindent\phantom{0}53.1 & \noindent\phantom{00}0.0 & \noindent\phantom{0}99.9 & \noindent\phantom{0}97.8 & \noindent\phantom{00}0.0 \\
OE & \noindent\phantom{0}99.4 & \textbf{100.0}             & \noindent\phantom{0}91.5 & \noindent\phantom{00}0.0 & \noindent\phantom{0}98.5 & \noindent\phantom{0}51.0 & \noindent\phantom{00}0.0 & \noindent\phantom{0}99.9 & \noindent\phantom{0}96.8 & \noindent\phantom{00}0.0 \\
ACET & \noindent\phantom{0}99.4 & \textbf{100.0}             & \noindent\phantom{0}99.2 & \noindent\phantom{00}0.0 & \noindent\phantom{0}\textbf{99.5} & \noindent\phantom{0}\textbf{76.5} & \noindent\phantom{00}0.0 & \textbf{100.0}             & \noindent\phantom{0}99.5 & \noindent\phantom{00}0.0 \\
CCU & \noindent\phantom{0}99.5 & \textbf{100.0}             & \noindent\phantom{0}75.0 & \noindent\phantom{00}0.0 & \noindent\phantom{0}98.1 & \noindent\phantom{00}3.4 & \noindent\phantom{00}0.0 & \textbf{100.0}             & \noindent\phantom{0}99.6 & \noindent\phantom{00}0.0 \\
GOOD\textsubscript{0} & \noindent\phantom{0}99.5 & \textbf{100.0}             & \noindent\phantom{0}93.8 & \noindent\phantom{00}0.0 & \noindent\phantom{0}98.6 & \noindent\phantom{0}55.7 & \noindent\phantom{00}0.0 & \noindent\phantom{0}99.9 & \noindent\phantom{0}97.7 & \noindent\phantom{00}0.0 \\
GOOD\textsubscript{20} & \noindent\phantom{0}99.0 & \textbf{100.0}             & \noindent\phantom{0}97.1 & \noindent\phantom{0}32.7 & \noindent\phantom{0}97.0 & \noindent\phantom{0}42.4 & \noindent\phantom{00}0.0 & \textbf{100.0}             & \noindent\phantom{0}99.6 & \noindent\phantom{0}19.3 \\
GOOD\textsubscript{40} & \noindent\phantom{0}99.0 & \textbf{100.0}             & \noindent\phantom{0}97.2 & \noindent\phantom{0}59.5 & \noindent\phantom{0}96.9 & \noindent\phantom{0}36.8 & \noindent\phantom{00}0.0 & \textbf{100.0}             & \noindent\phantom{0}99.7 & \noindent\phantom{0}44.7 \\
GOOD\textsubscript{60} & \noindent\phantom{0}99.0 & \textbf{100.0}             & \noindent\phantom{0}97.3 & \noindent\phantom{0}77.8 & \noindent\phantom{0}96.3 & \noindent\phantom{0}31.3 & \noindent\phantom{00}0.0 & \textbf{100.0}             & \noindent\phantom{0}99.8 & \noindent\phantom{0}76.2 \\
GOOD\textsubscript{80} & \noindent\phantom{0}99.1 & \textbf{100.0}             & \noindent\phantom{0}97.8 & \noindent\phantom{0}89.4 & \noindent\phantom{0}96.9 & \noindent\phantom{0}34.2 & \noindent\phantom{00}1.2 & \textbf{100.0}             & \noindent\phantom{0}99.9 & \noindent\phantom{0}96.7 \\
GOOD\textsubscript{90} & \noindent\phantom{0}98.8 & \textbf{100.0}             & \noindent\phantom{0}98.7 & \noindent\phantom{0}94.2 & \noindent\phantom{0}97.8 & \noindent\phantom{0}40.5 & \noindent\phantom{00}2.2 & \textbf{100.0}             & \noindent\phantom{0}99.9 & \noindent\phantom{0}99.2 \\
GOOD\textsubscript{95} & \noindent\phantom{0}98.8 & \textbf{100.0}             & \noindent\phantom{0}99.2 & \noindent\phantom{0}96.1 & \noindent\phantom{0}97.8 & \noindent\phantom{0}42.2 & \noindent\phantom{00}\textbf{2.4} & \textbf{100.0}             & \textbf{100.0}             & \noindent\phantom{0}\textbf{99.5} \\
\rowcolor{lightgrey}GOOD\textsubscript{100} & \noindent\phantom{0}98.7 & \textbf{100.0}             & \noindent\phantom{0}\textbf{99.5} & \noindent\phantom{0}\textbf{97.7} & \noindent\phantom{0}98.6 & \noindent\phantom{0}50.7 & \noindent\phantom{00}1.8 & \textbf{100.0}             & \noindent\phantom{0}99.9 & \noindent\phantom{0}99.3 \\
\bottomrule
\toprule
\multicolumn{11}{c}{in: SVHN \hspace{1cm} $\epsilon = 0.03$} \\
\midrule
\multicolumn{1}{l}{\multirow{2}{*}{Method \ }}  & \multicolumn{1}{c|}{\multirow{2}{*}{\ Acc. \ }}
 & \multicolumn{3}{c|}{80M Tiny Images}  & \multicolumn{3}{c|}{Imagenet-}  & \multicolumn{3}{c}{Smooth Noise} \\
& & auc & aauc & gauc & auc & aauc & gauc  & auc & aauc & gauc\\ \midrule
Plain & \noindent\phantom{0}95.5 & \noindent\phantom{0}94.8 & \noindent\phantom{0}11.9 & \noindent\phantom{00}0.0 & \noindent\phantom{0}95.5 & \noindent\phantom{0}13.4 & \noindent\phantom{00}0.0 & \noindent\phantom{0}96.0 & \noindent\phantom{00}5.6                 & \noindent\phantom{00}0.0 \\
CEDA & \noindent\phantom{0}95.3 & \noindent\phantom{0}99.9 & \noindent\phantom{0}64.4 & \noindent\phantom{00}0.0 & \noindent\phantom{0}99.9 & \noindent\phantom{0}75.3 & \noindent\phantom{00}0.0 & \noindent\phantom{0}96.8 & \noindent\phantom{00}5.9                 & \noindent\phantom{00}0.0 \\
OE & \noindent\phantom{0}95.5 & \noindent\phantom{0}\textbf{100.0}             & \noindent\phantom{0}61.8 & \noindent\phantom{00}0.0 & \textbf{100.0}             & \noindent\phantom{0}72.5 & \noindent\phantom{00}0.0 & \noindent\phantom{0}97.0 & \noindent\phantom{00}8.0                 & \noindent\phantom{00}0.0 \\
ACET & \noindent\phantom{0}96.0 & \textbf{100.0}             & \noindent\phantom{0}\textbf{99.3} & \noindent\phantom{00}0.0 & \textbf{100.0}             & \noindent\phantom{0}\textbf{99.6} & \noindent\phantom{00}0.0 & \noindent\phantom{0}99.9 & \noindent\phantom{0}\textbf{83.5}                 & \noindent\phantom{00}0.0 \\
CCU & \noindent\phantom{0}95.7   & \textbf{100.0}   & \noindent\phantom{0}48.8   & \noindent\phantom{00}0.0   & \textbf{100.0}       & \noindent\phantom{0}97.2    & \noindent\phantom{00}0.0  & \textbf{100.0}   & \noindent\phantom{00}5.7    & \noindent\phantom{00}0.0  \\
GOOD\textsubscript{0} & \noindent\phantom{0}97.0 & \textbf{100.0}             & \noindent\phantom{0}57.9 & \noindent\phantom{00}0.0 & \textbf{100.0}             & \noindent\phantom{0}68.3 & \noindent\phantom{00}0.0 & \noindent\phantom{0}97.8 & \noindent\phantom{0}25.0                 & \noindent\phantom{00}0.0 \\
GOOD\textsubscript{20} & \noindent\phantom{0}95.9 & \noindent\phantom{0}99.8 & \noindent\phantom{0}78.3 & \noindent\phantom{0}19.4 & \noindent\phantom{0}99.8 & \noindent\phantom{0}88.9 & \noindent\phantom{0}34.0 & \noindent\phantom{0}97.4 & \noindent\phantom{0}22.0                 & \noindent\phantom{00}0.0 \\
GOOD\textsubscript{40} & \noindent\phantom{0}96.3 & \noindent\phantom{0}99.5 & \noindent\phantom{0}81.0 & \noindent\phantom{0}44.4 & \noindent\phantom{0}99.5 & \noindent\phantom{0}90.1 & \noindent\phantom{0}62.6 & \noindent\phantom{0}97.1 & \noindent\phantom{0}21.0                 & \noindent\phantom{00}0.0 \\
GOOD\textsubscript{60} & \noindent\phantom{0}96.1 & \noindent\phantom{0}99.4 & \noindent\phantom{0}83.4 & \noindent\phantom{0}64.5  & \noindent\phantom{0}99.4 & \noindent\phantom{0}92.6 & \noindent\phantom{0}82.8 & \noindent\phantom{0}97.0 & \noindent\phantom{0}18.0                 & \noindent\phantom{00}0.0 \\
GOOD\textsubscript{80} & \noindent\phantom{0}96.3 & \textbf{100.0}             & \noindent\phantom{0}93.1 & \noindent\phantom{0}86.3 & \textbf{100.0}             & \noindent\phantom{0}97.4 & \noindent\phantom{0}95.6 & \noindent\phantom{0}96.8 & \noindent\phantom{0}29.1                 & \noindent\phantom{00}3.9 \\
GOOD\textsubscript{90} & \noindent\phantom{0}96.2 & \noindent\phantom{0}99.8 & \noindent\phantom{0}95.2 & \noindent\phantom{0}93.0 & \noindent\phantom{0}99.8 & \noindent\phantom{0}98.4 & \noindent\phantom{0}97.8 & \noindent\phantom{0}96.7 & \noindent\phantom{0}40.6                 & \noindent\phantom{0}20.6 \\
GOOD\textsubscript{95} & \noindent\phantom{0}96.4 & \noindent\phantom{0}99.7 & \noindent\phantom{0}96.4 & \noindent\phantom{0}95.2 & \noindent\phantom{0}99.8 & \noindent\phantom{0}98.8 & \noindent\phantom{0}98.4 & \noindent\phantom{0}96.8 & \noindent\phantom{0}59.1                 & \noindent\phantom{0}46.8 \\
\rowcolor{lightgrey}GOOD\textsubscript{100} & \noindent\phantom{0}96.3 & \noindent\phantom{0}99.6 & \noindent\phantom{0}97.2 &  \noindent\phantom{0}\textbf{96.8} & \noindent\phantom{0}99.8 & \noindent\phantom{0}99.1 & \noindent\phantom{0}\textbf{98.9} & \noindent\phantom{0}96.7 & \noindent\phantom{0}77.5                 & \noindent\phantom{0}\textbf{73.5} \\
\bottomrule
\toprule
\multicolumn{11}{c}{in: CIFAR-10 \hspace{.5cm} $\epsilon = 0.01$} \\
\midrule
\multicolumn{1}{l}{\multirow{2}{*}{Method \ }}  & \multicolumn{1}{c|}{\multirow{2}{*}{\ Acc. \ }}
 & \multicolumn{3}{c|}{80M Tiny Images}  & \multicolumn{3}{c|}{Imagenet-}  & \multicolumn{3}{c}{Smooth Noise} \\
& & auc & aauc & gauc & auc & aauc & gauc & auc & aauc & gauc \\ \midrule
Plain                  & \noindent\phantom{0}90.1 & \noindent\phantom{0}85.6 & \noindent\phantom{0}15.5 & \noindent\phantom{00}0.0 & \noindent\phantom{0}83.5 & \noindent\phantom{0}15.5 & \noindent\phantom{00}0.0 & \noindent\phantom{0}90.5 & \noindent\phantom{0}18.8                 & \noindent\phantom{00}0.0 \\
CEDA                   & \noindent\phantom{0}88.6 & \noindent\phantom{0}97.2 & \noindent\phantom{0}49.6 & \noindent\phantom{00}0.0 & \noindent\phantom{0}90.1 & \noindent\phantom{0}32.6 & \noindent\phantom{00}0.0 & \noindent\phantom{0}98.9 & \noindent\phantom{0}37.3                 & \noindent\phantom{00}0.0 \\
OE                     & \noindent\phantom{0}90.7 & \noindent\phantom{0}\textbf{97.3} & \noindent\phantom{0}20.5 & \noindent\phantom{00}0.0 & \noindent\phantom{0}90.3 & \noindent\phantom{0}12.1 & \noindent\phantom{00}0.0 & \noindent\phantom{0}99.5 & \noindent\phantom{0}11.3                 & \noindent\phantom{00}0.0 \\
ACET                   & \noindent\phantom{0}89.3 & \noindent\phantom{0}96.7 & \noindent\phantom{0}\textbf{88.8} & \noindent\phantom{00}0.0 & \noindent\phantom{0}89.5 & \noindent\phantom{0}\textbf{74.7} & \noindent\phantom{00}0.0 & \noindent\phantom{0}\textbf{99.9} &  \noindent\phantom{0}\textbf{98.8} &  \noindent\phantom{00}0.0 \\
CCU                    & \noindent\phantom{0}91.6 & \noindent\phantom{0}96.8 & \noindent\phantom{0}33.7 & \noindent\phantom{00}0.0 & \noindent\phantom{0}\textbf{92.0} & \noindent\phantom{0}30.0 & \noindent\phantom{00}0.0 & \noindent\phantom{0}99.5 & \noindent\phantom{0}38.0 & \noindent\phantom{00}0.0 \\
GOOD\textsubscript{0}  & \noindent\phantom{0}89.8 & \noindent\phantom{0}96.9 & \noindent\phantom{0}42.7 & \noindent\phantom{00}0.0 & \noindent\phantom{0}91.0 & \noindent\phantom{0}19.8 & \noindent\phantom{00}0.0 & \noindent\phantom{0}96.9 & \noindent\phantom{0}30.0                 & \noindent\phantom{00}0.0 \\
GOOD\textsubscript{20} & \noindent\phantom{0}88.5 & \noindent\phantom{0}96.6 & \noindent\phantom{0}48.5 & \noindent\phantom{0}16.3 & \noindent\phantom{0}88.8 & \noindent\phantom{0}30.5 & \noindent\phantom{00}6.9 & \noindent\phantom{0}96.5 & \noindent\phantom{0}64.5                 & \noindent\phantom{0}17.8 \\
GOOD\textsubscript{40} & \noindent\phantom{0}89.5 & \noindent\phantom{0}94.8 & \noindent\phantom{0}56.8 & \noindent\phantom{0}36.4 & \noindent\phantom{0}88.0 & \noindent\phantom{0}39.3 & \noindent\phantom{0}24.6 & \noindent\phantom{0}96.4 & \noindent\phantom{0}86.4                 & \noindent\phantom{0}27.5 \\
GOOD\textsubscript{60} & \noindent\phantom{0}90.2 & \noindent\phantom{0}95.2 & \noindent\phantom{0}60.7 & \noindent\phantom{0}48.7 & \noindent\phantom{0}87.4 & \noindent\phantom{0}46.1 & \noindent\phantom{0}36.7 & \noindent\phantom{0}97.5 & \noindent\phantom{0}81.4                 & \noindent\phantom{0}47.8 \\
\rowcolor{lightgrey}GOOD\textsubscript{80} & \noindent\phantom{0}90.1 & \noindent\phantom{0}93.1 & \noindent\phantom{0}62.8 & \noindent\phantom{0}55.9 & \noindent\phantom{0}84.0 & \noindent\phantom{0}50.0 & \noindent\phantom{0}42.3 & \noindent\phantom{0}95.1 & \noindent\phantom{0}74.1                 & \noindent\phantom{0}59.4 \\
GOOD\textsubscript{90} & \noindent\phantom{0}90.2 & \noindent\phantom{0}90.6 & \noindent\phantom{0}63.4 & \noindent\phantom{0}60.8 & \noindent\phantom{0}79.6 & \noindent\phantom{0}53.0 & \noindent\phantom{0}49.1 & \noindent\phantom{0}98.9 & \noindent\phantom{0}72.8                 & \noindent\phantom{0}62.3 \\
GOOD\textsubscript{95} & \noindent\phantom{0}90.4 & \noindent\phantom{0}88.9 & \noindent\phantom{0}63.4 & \noindent\phantom{0}62.0 & \noindent\phantom{0}77.6 & \noindent\phantom{0}54.3 & \noindent\phantom{0}50.3 & \noindent\phantom{0}92.0 & \noindent\phantom{0}61.8                 & \noindent\phantom{0}59.4 \\
GOOD\textsubscript{100} & \noindent\phantom{0}90.1 & \noindent\phantom{0}78.7 & \noindent\phantom{0}66.7 & \noindent\phantom{0}\textbf{66.3} & \noindent\phantom{0}69.0 & \noindent\phantom{0}56.9 & \noindent\phantom{0}\textbf{53.9} & \noindent\phantom{0}82.2 & \noindent\phantom{0}67.9                 & \noindent\phantom{0}\textbf{66.8} \\
\bottomrule
\end{tabularx}
}
\end{sc}
\end{small}
\end{center}
\vskip -0.1in
\end{table*}




%% file: sections/larger_radius.tex
\section{Generalization of provable confidence bounds to a larger radius}\label{section:bigger_bounds}
In Table~\ref{table:aucs_big_eps}, we evaluate the generalization of empirical worst case and guaranteed upper bound for the confidence within a larger $l_\infty$-ball around OOD samples than what the model was trained for.

As expected, the adversarial AUC's (AAUC) degrade for the larger radius $\epsilon$ for all methods. However, ACET and the GOOD models
with higher quantiles maintain their performance much better. Interestingly, while ACET has for the smaller radii typically better AAUCs this is reversed for the larger radii where now often the GOOD models
are better, showing that our certified methods can in this aspect sometimes outperform the ``adversarial training'' approach when it of generalization to higher radii.

On MNIST, GOOD\textsubscript{100} not only still has a perfect guaranteed 
AUC for uniform noise for an $\epsilon$ of 0.4 but even on FashionMNIST and CIFAR-10 it still has substantial guarantees.

For SVHN, the excellent guarantees of GOOD\textsubscript{100} for $\epsilon=0.03$ generalize well to the doubled radius of $\epsilon=0.06$ but the gap between GAUC and AAUC increases quite significantly, except for uniform noise where the GAUC is still high at $94.7\%$

For CIFAR-10, even when tripling the evaluation radius to $\epsilon = 0.03$, the certified the bounds of GOOD\textsubscript{80} generalize surprisingly well: for all out-distributions, we only see an at most moderate drop of the GAUC value compared to Table \ref{table:all_aucs}. 

In summary, GOOD in most cases still achieves reasonable guarantees for the larger threat model at test time. Moreover, the AAUC for the GOOD models is in most cases better than that of ACET and thus our guaranteed IBP training shows in this regard a better generalization to larger evaluation radii than adversarial training on the out-distribution (ACET).

\input{tables/table_aucs_bigger_eps}

%% file: tables/table_aucs_bigger_eps.tex
\begin{table*}[!htbp]
\caption{
Complementing Table~\ref{table:all_aucs}, an evaluation of the generalization of worst-case OOD detection, that is AAUC and GAUC, for $\epsilon$-values larger than those of the threat models used during training.
}
\label{table:aucs_big_eps}
\setlength\tabcolsep{.5pt} 
\vskip 0.15in
\begin{center}
\begin{small}
\begin{sc}
\makebox[\textwidth][c]{
\begin{tabularx}{1.0\textwidth}{lc|CCC|CCC|CCC|CCC}
\toprule
\multicolumn{14}{c}{in: MNIST \hspace{.8cm} $\epsilon = 0.4$} \\
\midrule
\multicolumn{1}{l}{\multirow{2}{*}{Method \ }}  & \multicolumn{1}{c|}{\multirow{2}{*}{\ Acc. \ }}
 & \multicolumn{3}{c|}{FashionMNIST}  & \multicolumn{3}{c|}{EMNIST Letters}  & \multicolumn{3}{c|}{CIFAR-10} & \multicolumn{3}{c}{Uniform Noise}  \\
& & auc & aauc & gauc & auc & aauc & gauc & auc & aauc & gauc & auc & aauc & gauc \\ \midrule
Plain & \noindent\phantom{0}99.4 & \noindent\phantom{0}98.0 & \noindent\phantom{0}28.6 & \noindent\phantom{00}0.0 & \noindent\phantom{0}88.0 & \noindent\phantom{0}26.9 & \noindent\phantom{00}0.0 & \noindent\phantom{0}98.8 & \noindent\phantom{0}32.4 & \noindent\phantom{00}0.0 & \noindent\phantom{0}99.2 & \noindent\phantom{0}34.3 & \noindent\phantom{00}0.0 \\
CEDA & \noindent\phantom{0}99.4 & \noindent\phantom{0}99.9 & \noindent\phantom{0}69.9 & \noindent\phantom{00}0.0 & \noindent\phantom{0}92.6 & \noindent\phantom{0}49.1 & \noindent\phantom{00}0.0 & \textbf{100.0}             & \noindent\phantom{0}82.8 & \noindent\phantom{00}0.0 & \textbf{100.0}             & \textbf{100.0}             & \noindent\phantom{00}0.0 \\
OE & \noindent\phantom{0}99.4 & \noindent\phantom{0}99.9 & \noindent\phantom{0}63.6 & \noindent\phantom{00}0.0 & \noindent\phantom{0}92.7 & \noindent\phantom{0}47.4 & \noindent\phantom{00}0.0 & \textbf{100.0}             & \noindent\phantom{0}76.0 & \noindent\phantom{00}0.0 & \textbf{100.0}             & \noindent\phantom{0}99.9 & \noindent\phantom{00}0.0 \\
ACET & \noindent\phantom{0}99.4 & \textbf{100.0}             & \noindent\phantom{0}91.3 & \noindent\phantom{00}0.0 & \noindent\phantom{0}95.9 & \noindent\phantom{0}47.8 & \noindent\phantom{00}0.0 & \textbf{100.0}             & \noindent\phantom{0}92.3 & \noindent\phantom{00}0.0 & \textbf{100.0}             & \textbf{100.0}             & \noindent\phantom{00}0.0 \\
CCU & \noindent\phantom{0}99.5 & \textbf{100.0}           & \noindent\phantom{0}62.0  & \noindent\phantom{00}0.0 & \noindent\phantom{0}92.9  & \noindent\phantom{00}2.7   & \noindent\phantom{00}0.0 & \textbf{100.0}  & \noindent\phantom{0}97.6   & \noindent\phantom{00}0.0   & \textbf{100.0}  & \textbf{100.0}  & \noindent\phantom{00}0.0    \\
GOOD\textsubscript{0} & \noindent\phantom{0}\textbf{99.5} & \noindent\phantom{0}99.9 & \noindent\phantom{0}70.8 & \noindent\phantom{00}0.0 & \noindent\phantom{0}92.9 & \noindent\phantom{0}51.8 & \noindent\phantom{00}0.0 & \textbf{100.0}             & \noindent\phantom{0}81.5 & \noindent\phantom{00}0.0 & \textbf{100.0}             & \textbf{100.0}             & \noindent\phantom{00}0.0 \\
GOOD\textsubscript{20} & \noindent\phantom{0}99.0 & \noindent\phantom{0}99.8 & \noindent\phantom{0}81.9 & \noindent\phantom{00}3.6 & \noindent\phantom{0}95.3 & \noindent\phantom{0}46.2 & \noindent\phantom{00}0.0 & \textbf{100.0}             & \noindent\phantom{0}91.4 & \noindent\phantom{00}6.4 & \textbf{100.0}             & \textbf{100.0}             & \noindent\phantom{0}99.9 \\
GOOD\textsubscript{40} & \noindent\phantom{0}99.0 & \noindent\phantom{0}99.8 & \noindent\phantom{0}81.6 & \noindent\phantom{0}18.5 & \noindent\phantom{0}95.7 & \noindent\phantom{0}46.9 & \noindent\phantom{00}0.0 & \textbf{100.0}             & \noindent\phantom{0}92.0 & \noindent\phantom{0}26.3 & \textbf{100.0}             & \textbf{100.0}             & \textbf{100.0}             \\
GOOD\textsubscript{60} & \noindent\phantom{0}99.0 & \noindent\phantom{0}99.9 & \noindent\phantom{0}82.5 & \noindent\phantom{0}30.6 & \noindent\phantom{0}96.6 & \noindent\phantom{0}47.1 & \noindent\phantom{00}0.0 & \textbf{100.0}             & \noindent\phantom{0}92.7 & \noindent\phantom{0}55.4 & \textbf{100.0}             & \textbf{100.0}             & \textbf{100.0}             \\
GOOD\textsubscript{80} & \noindent\phantom{0}99.1 & \noindent\phantom{0}99.8 & \noindent\phantom{0}84.5 & \noindent\phantom{0}41.9 & \noindent\phantom{0}97.9 & \noindent\phantom{0}\textbf{52.1} & \noindent\phantom{00}1.0 & \textbf{100.0}             & \noindent\phantom{0}93.8 & \noindent\phantom{0}77.3 & \textbf{100.0}             & \textbf{100.0}             & \textbf{100.0}             \\
GOOD\textsubscript{90} & \noindent\phantom{0}98.8 & \noindent\phantom{0}99.9 & \noindent\phantom{0}86.3 & \noindent\phantom{0}45.5 & \noindent\phantom{0}98.0 & \noindent\phantom{0}48.6 & \noindent\phantom{00}1.4 & \textbf{100.0}             & \noindent\phantom{0}95.7 & \noindent\phantom{0}77.6 & \textbf{100.0}             & \textbf{100.0}             & \textbf{100.0}             \\
GOOD\textsubscript{95} & \noindent\phantom{0}98.8 & \noindent\phantom{0}99.9 & \noindent\phantom{0}87.8 & \noindent\phantom{0}\textbf{49.0} & \noindent\phantom{0}98.7 & \noindent\phantom{0}47.0 & \noindent\phantom{00}\textbf{1.6} & \textbf{100.0}             & \noindent\phantom{0}96.8 & \noindent\phantom{0}\textbf{79.8} & \textbf{100.0}             & \textbf{100.0}             & \textbf{100.0}             \\
\rowcolor{lightgrey}GOOD\textsubscript{100} & \noindent\phantom{0}98.7 & \textbf{100.0}             & \noindent\phantom{0}\textbf{92.0} & \noindent\phantom{0}48.8 & \noindent\phantom{0}\textbf{99.0} & \noindent\phantom{0}39.1 & \noindent\phantom{00}0.8 & \textbf{100.0}             & \noindent\phantom{0}\textbf{98.2} & \noindent\phantom{0}75.9 & \textbf{100.0}             & \textbf{100.0}             & \textbf{100.0}             \\
\bottomrule
\toprule
\multicolumn{14}{c}{in: SVHN \hspace{1cm} $\epsilon = 0.06$} \\
\midrule
\multicolumn{1}{l}{\multirow{2}{*}{Method \ }}  & \multicolumn{1}{c|}{\multirow{2}{*}{\ Acc. \ }}
 & \multicolumn{3}{c|}{CIFAR-100}  & \multicolumn{3}{c|}{CIFAR-10}  & \multicolumn{3}{c|}{LSUN Classroom} & \multicolumn{3}{c}{Uniform Noise}  \\
& & auc & aauc & gauc & auc & aauc & gauc & auc & aauc & gauc & auc & aauc & gauc \\
\midrule
Plain & \noindent\phantom{0}95.5 & \noindent\phantom{0}94.9 & \noindent\phantom{00}5.6 & \noindent\phantom{00}0.0 & \noindent\phantom{0}95.2 & \noindent\phantom{00}5.6 & \noindent\phantom{00}0.0 & \noindent\phantom{0}95.7 & \noindent\phantom{00}1.3 & \noindent\phantom{00}0.0 & \noindent\phantom{0}99.4 & \noindent\phantom{0}13.2 & \noindent\phantom{00}0.0 \\
CEDA & \noindent\phantom{0}95.3 & \noindent\phantom{0}99.9 & \noindent\phantom{0}19.3 & \noindent\phantom{00}0.0 & \noindent\phantom{0}99.9 & \noindent\phantom{0}24.2 & \noindent\phantom{00}0.0 & \noindent\phantom{0}99.9 & \noindent\phantom{0}43.5 & \noindent\phantom{00}0.0 & \noindent\phantom{0}99.9 & \noindent\phantom{0}21.6 & \noindent\phantom{00}0.0 \\
OE & \noindent\phantom{0}95.5 & \textbf{100.0}             & \noindent\phantom{0}14.4 & \noindent\phantom{00}0.0 & \textbf{100.0}             & \noindent\phantom{0}15.8 & \noindent\phantom{00}0.0 & \textbf{100.0}             & \noindent\phantom{0}22.2 & \noindent\phantom{00}0.0 & \textbf{100.0}             & \noindent\phantom{0}34.0 & \noindent\phantom{00}0.0 \\
ACET & \noindent\phantom{0}96.0 & \textbf{100.0}             & \noindent\phantom{0}90.4 & \noindent\phantom{00}0.0 & \textbf{100.0}             & \noindent\phantom{0}90.6 & \noindent\phantom{00}0.0 & \textbf{100.0}             & \noindent\phantom{0}\textbf{96.8} & \noindent\phantom{00}0.0 & \noindent\phantom{0}99.9 & \noindent\phantom{0}58.6 & \noindent\phantom{00}0.0 \\
CCU & \noindent\phantom{0}95.7   & \textbf{100.0}   & \noindent\phantom{0}10.3   & \noindent\phantom{00}0.0   & \textbf{100.0}       & \noindent\phantom{00}4.7                & \noindent\phantom{00}0.0  & \textbf{100.0}   & \noindent\phantom{00}6.0    & \noindent\phantom{00}0.0   & \textbf{100.0}   & \textbf{100.0}   & \noindent\phantom{00}0.0   \\
GOOD\textsubscript{0} & \noindent\phantom{0}\textbf{97.0} & \textbf{100.0}             & \noindent\phantom{0}36.5 & \noindent\phantom{00}0.0 & \textbf{100.0}             & \noindent\phantom{0}37.1 & \noindent\phantom{00}0.0 & \textbf{100.0}             & \noindent\phantom{0}17.0 & \noindent\phantom{00}0.0 & \textbf{100.0}             & \noindent\phantom{0}44.3 & \noindent\phantom{00}0.0 \\
GOOD\textsubscript{20} & \noindent\phantom{0}95.9 & \noindent\phantom{0}99.8 & \noindent\phantom{0}54.6 & \noindent\phantom{00}7.6 & \noindent\phantom{0}99.9 & \noindent\phantom{0}55.4 & \noindent\phantom{00}2.9 & \noindent\phantom{0}99.9 & \noindent\phantom{0}80.0 & \noindent\phantom{00}2.0 & \noindent\phantom{0}99.7 & \noindent\phantom{0}99.5 & \noindent\phantom{00}0.6 \\
GOOD\textsubscript{40} & \noindent\phantom{0}96.3 & \noindent\phantom{0}99.5 & \noindent\phantom{0}59.4 & \noindent\phantom{0}12.6 & \noindent\phantom{0}99.5 & \noindent\phantom{0}64.4 & \noindent\phantom{0}11.1 & \noindent\phantom{0}99.5 & \noindent\phantom{0}86.1 & \noindent\phantom{00}7.6 & \noindent\phantom{0}99.5 & \noindent\phantom{0}99.5 & \noindent\phantom{00}0.1 \\
GOOD\textsubscript{60} & \noindent\phantom{0}96.1 & \noindent\phantom{0}99.4 & \noindent\phantom{0}64.3 & \noindent\phantom{0}27.4 & \noindent\phantom{0}99.4 & \noindent\phantom{0}68.9 & \noindent\phantom{0}26.8 & \noindent\phantom{0}99.4 & \noindent\phantom{0}87.4 & \noindent\phantom{0}25.7 & \noindent\phantom{0}99.4 & \noindent\phantom{0}99.4 & \noindent\phantom{0}18.3 \\
GOOD\textsubscript{80} & \noindent\phantom{0}96.3 & \textbf{100.0}             & \noindent\phantom{0}80.0 & \noindent\phantom{0}49.4 & \textbf{100.0}             & \noindent\phantom{0}84.0 & \noindent\phantom{0}50.7 & \textbf{100.0}             & \noindent\phantom{0}93.3 & \noindent\phantom{0}50.9 & \textbf{100.0}             & \noindent\phantom{0}99.7 & \noindent\phantom{0}28.6 \\
GOOD\textsubscript{90} & \noindent\phantom{0}96.2 & \noindent\phantom{0}99.8 & \noindent\phantom{0}86.0 & \noindent\phantom{0}57.2 & \noindent\phantom{0}99.8 & \noindent\phantom{0}89.2 & \noindent\phantom{0}60.0 & \noindent\phantom{0}99.8 & \noindent\phantom{0}95.6 & \noindent\phantom{0}61.8 & \noindent\phantom{0}99.8 & \noindent\phantom{0}99.8 & \noindent\phantom{0}91.8 \\
GOOD\textsubscript{95} & \noindent\phantom{0}96.4 & \noindent\phantom{0}99.8 & \noindent\phantom{0}89.2 & \noindent\phantom{0}73.4 & \noindent\phantom{0}99.8 & \noindent\phantom{0}91.5 & \noindent\phantom{0}76.3 & \noindent\phantom{0}99.8 & \noindent\phantom{0}96.2 & \noindent\phantom{0}78.6 & \noindent\phantom{0}99.9 & \noindent\phantom{0}99.8 & \noindent\phantom{0}\textbf{98.3} \\
\rowcolor{lightgrey}GOOD\textsubscript{100} & \noindent\phantom{0}96.3 & \noindent\phantom{0}99.6 & \noindent\phantom{0}\textbf{91.7} & \noindent\phantom{0}\textbf{80.6} & \noindent\phantom{0}99.7 & \noindent\phantom{0}\textbf{93.6} & \noindent\phantom{0}\textbf{82.9} & \noindent\phantom{0}99.9 & \noindent\phantom{0}96.4 & \noindent\phantom{0}\textbf{82.1} & \textbf{100.0}             & \noindent\phantom{0}99.8 & \noindent\phantom{0}94.7 \\
\bottomrule
\toprule
\multicolumn{14}{c}{in: CIFAR-10 \hspace{.5cm} $\epsilon = 0.03$} \\
\midrule
\multicolumn{1}{l}{\multirow{2}{*}{Method \ }}  & \multicolumn{1}{c|}{\multirow{2}{*}{\ Acc. \ }}
 & \multicolumn{3}{c|}{CIFAR-100}  & \multicolumn{3}{c|}{SVHN}  & \multicolumn{3}{c|}{LSUN Classroom} & \multicolumn{3}{c}{Uniform Noise}  \\
& & auc & aauc & gauc & auc & aauc & gauc & auc & aauc & gauc & auc & aauc & gauc \\ \midrule
Plain & \noindent\phantom{0}90.1 & \noindent\phantom{0}84.3 & \noindent\phantom{00}4.5 & \noindent\phantom{00}0.0 & \noindent\phantom{0}87.7 & \noindent\phantom{00}3.8 & \noindent\phantom{00}0.0 & \noindent\phantom{0}88.9 & \noindent\phantom{00}4.9 & \noindent\phantom{00}0.0 & \noindent\phantom{0}90.8 & \noindent\phantom{0}12.1 & \noindent\phantom{00}0.0 \\
CEDA & \noindent\phantom{0}88.6 & \noindent\phantom{0}91.8 & \noindent\phantom{00}5.2 & \noindent\phantom{00}0.0 & \noindent\phantom{0}\textbf{97.9} & \noindent\phantom{00}3.8 & \noindent\phantom{00}0.0 & \noindent\phantom{0}98.9 & \noindent\phantom{00}7.5 & \noindent\phantom{00}0.0 & \noindent\phantom{0}97.4 & \noindent\phantom{0}18.6 & \noindent\phantom{00}0.0 \\
OE & \noindent\phantom{0}90.7 & \noindent\phantom{0}92.4 & \noindent\phantom{00}0.4 & \noindent\phantom{00}0.0 & \noindent\phantom{0}97.6 & \noindent\phantom{00}0.1 & \noindent\phantom{00}0.0 & \noindent\phantom{0}98.9 & \noindent\phantom{00}0.2 & \noindent\phantom{00}0.0 & \noindent\phantom{0}98.7 & \noindent\phantom{00}1.7 & \noindent\phantom{00}0.0 \\
ACET & \noindent\phantom{0}89.3 & \noindent\phantom{0}90.7 & \noindent\phantom{0}34.5 & \noindent\phantom{00}0.0 & \noindent\phantom{0}96.6 & \noindent\phantom{0}49.1 & \noindent\phantom{00}0.0 & \noindent\phantom{0}98.3 & \noindent\phantom{0}47.7 & \noindent\phantom{00}0.0 & \noindent\phantom{0}99.7 & \noindent\phantom{0}86.3 & \noindent\phantom{00}0.0 \\
CCU & \noindent\phantom{0}\textbf{91.6} & \noindent\phantom{0}\textbf{93.0} & \noindent\phantom{00}1.6 & \noindent\phantom{00}0.0 & \noindent\phantom{0}97.1 & \noindent\phantom{0}11.2 & \noindent\phantom{00}0.0 & \noindent\phantom{0}\textbf{99.3} & \noindent\phantom{00}0.4 & \noindent\phantom{00}0.0 & \textbf{100.0} & \textbf{100.0} & \noindent\phantom{00}0.0 \\
GOOD\textsubscript{0} & \noindent\phantom{0}89.8 & \noindent\phantom{0}92.9 & \noindent\phantom{00}4.1 & \noindent\phantom{00}0.0 & \noindent\phantom{0}97.0 & \noindent\phantom{00}3.3 & \noindent\phantom{00}0.0 & \noindent\phantom{0}98.3 & \noindent\phantom{00}3.7 & \noindent\phantom{00}0.0 & \noindent\phantom{0}96.4 & \noindent\phantom{0}66.9 & \noindent\phantom{00}0.0 \\
GOOD\textsubscript{20} & \noindent\phantom{0}88.5 & \noindent\phantom{0}90.3 & \noindent\phantom{0}12.6 & \noindent\phantom{00}3.6 & \noindent\phantom{0}95.9 & \noindent\phantom{0}17.0 & \noindent\phantom{00}3.9 & \noindent\phantom{0}98.2 & \noindent\phantom{00}4.1 & \noindent\phantom{00}0.0 & \noindent\phantom{0}99.4 & \noindent\phantom{0}83.1 & \noindent\phantom{00}0.0 \\
GOOD\textsubscript{40} & \noindent\phantom{0}89.5 & \noindent\phantom{0}89.6 & \noindent\phantom{0}19.5 & \noindent\phantom{0}15.8 & \noindent\phantom{0}95.4 & \noindent\phantom{0}20.9 & \noindent\phantom{0}18.5 & \noindent\phantom{0}96.0 & \noindent\phantom{0}17.6 & \noindent\phantom{0}11.8 & \noindent\phantom{0}92.1 & \noindent\phantom{0}89.8 & \noindent\phantom{0}89.8 \\
GOOD\textsubscript{60} & \noindent\phantom{0}90.2 & \noindent\phantom{0}88.6 & \noindent\phantom{0}27.3 & \noindent\phantom{0}25.2 & \noindent\phantom{0}95.6 & \noindent\phantom{0}33.0 & \noindent\phantom{0}30.5 & \noindent\phantom{0}97.0 & \noindent\phantom{0}35.2 & \noindent\phantom{0}31.9 & \noindent\phantom{0}91.8 & \noindent\phantom{0}91.1 & \noindent\phantom{0}91.0 \\
\rowcolor{lightgrey}GOOD\textsubscript{80} & \noindent\phantom{0}90.1 & \noindent\phantom{0}85.6 & \noindent\phantom{0}36.0 & \noindent\phantom{0}32.6 & \noindent\phantom{0}94.0 & \noindent\phantom{0}33.1 & \noindent\phantom{0}31.7 & \noindent\phantom{0}93.3 & \noindent\phantom{0}45.4 & \noindent\phantom{0}41.0 & \noindent\phantom{0}95.8 & \noindent\phantom{0}95.2 & \noindent\phantom{0}95.1 \\
GOOD\textsubscript{90} & \noindent\phantom{0}90.2 & \noindent\phantom{0}81.7 & \noindent\phantom{0}44.0 & \noindent\phantom{0}43.0 & \noindent\phantom{0}91.4 & \noindent\phantom{0}42.8 & \noindent\phantom{0}41.2 & \noindent\phantom{0}90.2 & \noindent\phantom{0}50.4 & \noindent\phantom{0}49.0 & \noindent\phantom{0}89.3 & \noindent\phantom{0}87.7 & \noindent\phantom{0}87.6 \\
GOOD\textsubscript{95} & \noindent\phantom{0}90.4 & \noindent\phantom{0}80.3 & \noindent\phantom{0}45.1 & \noindent\phantom{0}44.7 & \noindent\phantom{0}90.2 & \noindent\phantom{0}39.2 & \noindent\phantom{0}38.2 & \noindent\phantom{0}88.3 & \noindent\phantom{0}52.6 & \noindent\phantom{0}51.3 & \noindent\phantom{0}96.6 & \noindent\phantom{0}95.8 & \noindent\phantom{0}95.7 \\
GOOD\textsubscript{100} & \noindent\phantom{0}90.1 & \noindent\phantom{0}70.0 & \noindent\phantom{0}\textbf{47.9} & \noindent\phantom{0}\textbf{46.5} & \noindent\phantom{0}75.5 & \noindent\phantom{0}\textbf{52.1} & \noindent\phantom{0}\textbf{50.0} & \noindent\phantom{0}75.2 & \noindent\phantom{0}\textbf{53.8} & \noindent\phantom{0}\textbf{52.2} & \noindent\phantom{0}99.5 & \noindent\phantom{0}98.7 & \noindent\phantom{0}\textbf{97.6} \\
\bottomrule
\end{tabularx}
}
\end{sc}
\end{small}
\end{center}
\vskip -0.1in
\end{table*}


